\RequirePackage{fix-cm}
\documentclass[]{article}
\usepackage{arxiv}
\usepackage[english]{babel}
\usepackage{lipsum}
\usepackage{graphicx} 
\usepackage{subcaption}
\usepackage{placeins}
\usepackage{bm}
\usepackage{booktabs} 
\usepackage{csquotes}
\usepackage{multirow}
\usepackage{listings}
\usepackage{xcolor}
\usepackage{float}
\usepackage{color, colortbl}
\definecolor{ao(english)}{rgb}{0.0, 0.5, 0.0}
\usepackage{xr-hyper}
\usepackage{hyperref}
\hypersetup{colorlinks,linkcolor={ao(english)},citecolor={ao(english)},urlcolor={red}} 
\usepackage{stackrel,amssymb}
\usepackage{amsfonts,amsmath,amstext,amsbsy,amsthm}
\theoremstyle{definition}

\usepackage{cleveref}
\usepackage{amsmath}
\definecolor{LightCyan}{rgb}{0.88,1,1}
\definecolor{LightRed}{rgb}{1,0.7,0.7}

\title{Discrete-Time Nonlinear Feedback Linearization \\
via Physics-Informed Machine Learning}

\author{
  Hector Vargas Alvarez\\
  Scuola Superiore Meridionale,\\
  Naples, Italy\\
  \texttt{hector.vargasalvarez-ssm@unina.it} \\
  \And
Gianluca Fabiani\\
Scuola Superiore Meridionale,\\
Naples, Italy \&\\
Dept. Chemical \& Biomolecular Engineering,\\
Johns Hopkins University,\\
Baltimore, USA\\
  \texttt{gianluca.fabiani@unina.it} \\
   \And
Nikolaos Kazantzis\\
Dept. Chemical Engineering,\\
Worcester Polytechnic Institute,\\
Worcester, USA\\
\texttt{nikolas@wpi.edu}
  \And
  Constantinos Siettos\thanks{Corresponding author}\\
Dept. Mathematics \& Applications,\\
Universit\`a degli Studi di Napoli ``Federico II",\\
Naples, Italy\\
\texttt{constantinos.siettos@unina.it}
\And
  Ioannis G. Kevrekidis\thanks{Corresponding author}\\
  Dept. Chemical \& Biomolecular Engineering,\\
  Dept. Applied Mathematics,\\
  Medical School,\\
  Johns Hopkins University,\\
  Baltimore, USA\\
  \texttt{yannisk@jhu.edu}
  }

\usepackage{amsopn}

\begin{document}

\maketitle

\begin{abstract}
We present a physics-informed machine learning (PIML) scheme for the feedback linearization of nonlinear discrete-time dynamical systems. The PIML finds the nonlinear transformation law, thus ensuring stability via pole placement, in one step. In order to facilitate convergence in the presence of steep gradients in the nonlinear transformation law, we address a greedy-wise training procedure. We assess the performance of the proposed PIML approach via a benchmark nonlinear discrete map for which the feedback linearization transformation law can be derived analytically; the example is characterized by steep gradients, due to the presence of singularities, in the domain of interest.
We show that the proposed PIML outperforms, in terms of numerical approximation accuracy, the ``traditional'' numerical implementation, which involves the construction --and the solution in terms of the coefficients of a power-series expansion--of a system of homological equations as well as the implementation of the PIML in the entire domain, thus highlighting the importance of continuation techniques in the training procedure of PIML.\end{abstract}

\newpage
\section{Introduction\label{sec:intro}}
A fundamental controller synthesis and design approach for nonlinear discrete-time systems relies on the use of feedback to explicitly modify the system dynamics and induce desirable dynamic characteristics that conform to a prespecified set of performance requirements and design objectives \cite{Isidori1995,krstic1995nonlinear,Sepulchre2011,Chen2013}. In particular, introducing feedback action to explicitly assign desirable dynamic modes to the controlled (closed-loop) system, by placing its poles at specific locations on the complex plane, represents an important and powerful technique in modern nonlinear control theory and practice \cite{Isidori1995,krstic1995nonlinear,Sepulchre2011,kravaris1990a,kravaris1990b}. Two dominant pole-placing nonlinear feedback control approaches can be discerned in the pertinent literature, with historical roots in geometric control theory \cite{kravaris1990a,kravaris1990b,Isidori1995,Sepulchre2011}. The first, known as the exact input/output feedback linearization approach, uses appropriately derived state feedback control laws to induce linear input/output behavior by forcing the system’s output variable to track a pre-specified “target” linear and stable trajectory. This approach offers a nonlinear analogue to the classic linear pole-placement method, where the closed-loop poles are placed
at prespecified values; yet it is limited to the special class of minimum-phase systems. However, in broad classes of nonlinear system regulation and/or stabilization problems, the primary objective is not only to force the system output variable to track a prespecified set-point profile, but rather to force {\em all system states} to return to desirable design steady states (equilibria) in a fast and smooth manner whenever the system experiences inevitable dynamic excursions due to the effect of disturbances \cite{Isidori1995,Sepulchre2011,kazantzis2000}. Within the context of geometric exact feedback linearization, this second approach was first introduced in the seminal and insightful works presented in \cite{Monaco1983,Lee1986,Grizzle1986,Jakubczyk1987,Nam1989,Lin1995,ArandaBricaire1996,kumar1996state,Krener1999} and is realized through a two-step controller synthesis/design procedure. In the first step, a nonlinear coordinate transformation and a state feedback control law are derived capable of transforming the original system into a linear and controllable one, under an external reference input and in an affine state space representation. The natural second step involves the employment of well-established pole-placement methods applied to the transformed linear system. It should be pointed out, however, that the exact feedback linearization approach impinges on a set of rather restrictive conditions that can hardly be met by physical and engineering systems. 

In a conceptually different problem reformulation, Guardabassi and Savaresi \cite{Guardabassi1997} addressed the feedback linearization problem using the so-called  {\em virtual input direct design} approach, whose principal characteristic is that it reduces the control problem into a standard non-linear mapping approximation problem, without resorting to a preliminary construction of an ODE-based model.
It is worth noting that in the formulation of a nonlinear feedback regulation/stabilization problem as described earlier, the presence of an external reference input variable, introduced in the first step of the classic exact feedback linearization approach, becomes irrelevant and redundant \cite{Isidori1995,Sepulchre2011,kazantzis2000}.  In the light of the above realization, and conceptually inspired by Luenberger’s early ideas on a single-step approach to feedback-induced pole-placement in continuous-time linear systems theory \cite{Luenberger1963}, Kazantzis \cite{Kazantzis2001} developed a nonlinear discrete-time analogue by formulating the problem within the context of nonlinear functional equations theory. This approach allows the assignment of the controlled (closed-loop) system’s dynamic modes by meeting {\em both the feedback linearization and the pole-placement objectives} in a single-step, while effectively overcoming the restrictive conditions associated with the traditional two-step exact feedback linearization approach. Within a similar conceptual and methodological context, further interesting investigations on deriving approximants of the feedback-linearizing and pole-placing control laws \cite{deutscher06}, rigorously establishing links to key system-theoretic concepts such as immersion and invariance properties \cite{karagiannis05} as well as implementing the controller synthesis method to partially distributed systems \cite{xu19} are noteworthy.

Between the mid 90s and early 2000s, emerging research activity focused on the development of various feedback-linearization methods with integrated machine learning capabilities \cite{Yesildirek1995,taprantzis1997fuzzy,he1998neural,siettos1999advanced,ge1999nonlinear,Siettos2002}. For example, Yeşildirek and Lewis \cite{Yesildirek1995} addressed the control of a class of single input single output (SISO) nonlinear systems using a multilayer artificial neural network (ANN)-based controller that performs feedback linearization while ensuring Lyapunov stability. He et al. \cite{he1998neural}, proposed a scheme based on the concept of feedback linearization and ANNs to simultaneously approximate the  nonlinear  transformation and the controller dynamics itself. Siettos et al. \cite{siettos1999advanced} proposed a fuzzy controller for the stabilization of equilibria of fluidized bed dryers and compared its performance with input-output linearization. Ge et al. \cite{ge1999nonlinear} used multilayer ANNs to reconstruct an implicit feedback linearization scheme for adaptive tracking control purposes. Siettos and Bafas \cite{Siettos2002} combined fuzzy logic and feedback linearization in order to achieve ``semiglobal'' stabilization of nonlinear singularly perturbed systems; a fuzzy scheme was used to decompose the full system into fast and slow dynamics, and then feedback linearization was employed under a set of properly derived sufficient conditions for Lyapunov stability. Deng et al. \cite{deng2008feedback} proposed a feedback linearization scheme implemented by ANNs for the adaptive control of non-affine nonlinear discrete-time systems.\par

More recently, theoretical and technological advances have renewed the interest of the control community towards the development of new schemes, based on machine learning, by  revisiting well established methods as well as introducing new ones. For example, Yang et al. \cite{Yang2014} proposed a direct adaptive control scheme based on reinforcement learning to improve the tracking performance for multi input, multi output unknown non-affine nonlinear discrete time systems. Umlauft et al. \cite{Umlauft2017} developed a model using Gaussian process regression in order to apply feedback linearization based on sets of training data, whereas Wu et al. \cite{wu2019real} proposed a machine learning-based predictive control scheme based on recurrent neural networks (RNNs) to approximate nonlinear dynamics and guarantee Lyapunov stability in the presence of model uncertainty. Tang and Daoutidis \cite{tang2019dissipativity} proposed a data-driven dissipative-based control scheme based on input–output data, where the learning and the controller design are carried out successively. Moreover, Westenbroek et al. \cite{Westenbroek2020} used reinforcement learning to build a linearizing controller for a given system using numerical approximation architectures. For a comprehensive review of machine-learning based model predictive control schemes, the interested reader is referred to the review paper by Ren et al.\cite{ren2022tutorial}. Recently, Patsatzis et al.\cite{patsatzis2023data} proposed an equation/variable free data-driven control approach for agent-based models based on machine/manifold learning which does not require knowledge of the ``correct'' macroscopic observables nor of any physical insights into the ``correct'' type of ODEs or PDEs, thus obviating the need to construct explicitly surrogate, reduced-order machine-learning models (such as ANNs, Deep Learning and Gaussian Processes), that {\em de facto} introduce biases.\par

Here, based on the concept of physics-informed machine learning (PIML) \cite{Raissi2019,karniadakis2021physics,cai2021physics}, we propose a scheme for learning a feedback linearizing control law for nonlinear discrete-time systems. In contrast to other previous works that used machine learning and in particular Artificial Neural Networks (ANNs) to first approximate the feedback linearizing transformation, and only then to apply a control law, our approach achieves this objective in a single step, based on a problem reformulation aligned with the methodological framework of simultaneously attaining  feedback linearization and pole placement \cite{kazantzis2000,Kazantzis2001,siettos2006}. The theoretical background (conceptual, methodological and analytical foundations) of the ``traditional'' scheme is discussed in \cite{kazantzis2000,Kazantzis2001}, and its Equation-free version for the control of microscopic simulators is presented in \cite{siettos2006} (see also  \cite{Siettos2002,siettos2004coarse,armaou2004time,siettos2012equation,patsatzis2023data} for the Equation-free control approach). In these research studies, the numerical approximation of the feedback-linearizing transformation and control is performed by (i) first approximating the transformation map with a power-series expansion, and (ii) then, by using a symbolic software package, recursively solving a standard Lyapunov matrix equation and a system of linear algebraic equations for the unknown coefficients of the aforementioned power-series expansion. However, such a power-series expansion procedure becomes intractable even for medium-scale dimensions and, cannot guarantee the desired numerical approximation accuracy {\em in the entire domain}, especially in regimes  that contain very-steep gradients that resemble singularities. Thus, in order to facilitate the numerical approximation ability of the proposed PIML scheme in the presence of steep-gradients, we follow a step/greedy-wise  training procedure (see also \cite{larochelle2009exploring}): we start learning the nonlinear transformation {\em on a subset of the entire domain}, where the transformation is sought, and gradually augmenting its size, thus ``warm-restarting'' the training procedure using as initial guesses for the unknown weights of the ANN, the ones found from the previous step, in the spirit of continuation/homotopy computations. 

This PIML scheme can be  easily implemented. In fact, for our illustrations, we developed a ``home-made'' code  in Matlab 2022; for training, we wrapped around it the Levenberg-Marquard optimization algorithm as implemented by the \texttt{nonlinsq} function. For comparison purposes, an implementation in Python using the Keras API of TensorFlow library was also performed. To demonstrate the efficiency of the proposed continuation/greedy-wise PIML scheme, we used a benchmark two-dimensional discrete-time model whose feedback-linearizing control law can be derived analytically \cite{Kazantzis2001}. The particular transformation map (and thus the attendant feedback-linearizing control law) exhibits a singularity at the domain boundary, thus making it difficult to approximate it numerically close to that point, especially through a power-series expansion. For illustration purposes, we also compared the numerical approximation accuracy of the proposed PIML greedy scheme against  the standard power-series expansion, as well as Matlab and Python's TensorFlow-based implementations with automatic differentiation that was used to learn the transformation in the entire domain. Furthermore, we considered two different scenarios, namely: (a) one where  we assumed that the equations of the model are explicitly known, and (b) one where we assumed that only a black-box simulator is available, i.e., pertinent equations are not available explicitly in closed form.\par

The paper is organized as follows: in Section 2, we provide a brief review and preliminaries related to the methodological framework associated with the single-step feedback linearization method for nonlinear discrete-time systems, and then we describe the proposed PIML scheme.  The benchmark problem is presented in Section 3. Section 4 encompasses the numerical results obtained under the various approaches, as well as a comparative performance assessment. Finally, concluding remarks are offered in Section 5.

\section{Methodological Framework: Using Physics-Informed Machine Learning for Feedback Linearization and Pole-Placement in a Single Step.}
Nonlinear discrete-time input-driven dynamical systems are considered with the following non-affine state-space realization:
\begin{equation}
x(t+1)=f(x(t),u(t)),
\label{eq:discrete_system}
\end{equation}
where $t=0,1,...$ is the discrete-time index, $x(t) \in R^{n}$ is the vector of state variables, $u(t) \in
R$ is the input variable and $f(x,u)$ is a real analytic vector function
defined on $R^{n} \times R$.

Without loss of generality, let us assume that the origin $x^{0}=0$ is an equilibrium point of \eqref{eq:discrete_system} that corresponds to $u^{0}=0$: $f(0,0)=0$. If a non-zero equilibrium state $(x^{0}, u^{0}) \neq (0,0)$ is considered, then a simple transformation of variables: $\hat{x}=x-x^{0}$, $\hat{u}=u-u^{0}$ will map it onto the origin in the new coordinates. Moreover, let
$J$ be the Jacobian matrix of $f(x,u)$ evaluated at the equilibrium point $(x^{0},u^{0})=(0,0)$: $\displaystyle{J=\frac{\partial f}{\partial
x}(0,0)}$, and $G$ a non-zero vector: $\displaystyle{G=\frac{\partial f}{\partial
u}(0,0)} \neq 0$. 

We now seek the attainment of the feedback linearization and pole-placement objectives in a single-step. In particular, a transformation map: $z=T(x)$, $T: R^{n} \longrightarrow R^{n}$ and a 
state feedback control law: $u=-cz=-cT(x)$, with $c$ an n-dimensional constant row vector are sought, that induce linear dynamics with prescribed modes/poles in the new coordinates:
\begin{equation}
z(t+1)=Az(t),
\label{eq:Az_linearized_sys}
\end{equation}
where the matrix $A$ represents a ``design adjustable parameter" whose eigenvalues are placed at the desirable set of dynamic modes/poles. The existence of such a nonlinear feedback linearizing control law is guaranteed by the following Theorem \cite{Kazantzis2001}:

{\bf Theorem 2.1:} {\it Consider the nonlinear discrete-time system \eqref{eq:discrete_system} and the associated system of nonlinear functional equations (NFEs) (3):
\begin{eqnarray}
T(f(x,-cT(x)))&=&AT(x) \nonumber \\
T(0)&=&0
\label{eq:NFESYS}
\end{eqnarray}
The following assumptions are made:

{\bf Assumption 1:} The ($n \times n$) matrix ${\mathcal{C}}$:
\begin{equation}
{\mathcal{C}}=\left[\begin{matrix}{G|JG|...|J^{n-1}G}\end{matrix}\right]
\end{equation}
has rank $n$: rank$({\mathcal{C}})=n$ (local controllability rank condition).

{\bf Assumption 2:} The eigenspectrum $\sigma(A)$ of matrix $A$ comprises eigenvalues: $k_{i} \in \sigma(A)$, $i=1,...n$  that all lie inside the unit disc on the complex plane (Poincar\'{e} domain).

{\bf Assumption 3:} The eigenspectra $\sigma(A),\sigma(J)$ of matrices $A$ and $J$ respectively are disjoint: $\sigma(A) \cap \sigma(J) = \emptyset$.

{\bf Assumption 4:} The eigenvalues $k_{i}$ of $A$ are not related to the eigenvalues $\lambda_{j}$ of the Jacobian matrix $J$ through any equations of the type:
\begin{equation}
\prod_{i=1}^{n}k_{i}^{m_{i}}=\lambda_{j}
\label{eq:cond1}
\end{equation}
$(j=1,...,n)$, where all the $m_{i}$'s are non-negative integers that satisfy the condition:
\begin{equation}
\sum_{i=1}^{n}m_{i}>0
\label{eq:cond2}
\end{equation}

{\bf Assumption 5:} The pair of matrices $(c,A)$ is
chosen such that the following matrix $O$:
\begin{equation}
O=\begin{bmatrix}
    c\\
    cA\\
    .\\
    .\\
  cA^{n-1}  
\end{bmatrix}
\end{equation}
has rank $n$: rank$(O)=n$ (observability rank condition on the $(c,A)$ pair).

Then, the associated system of NFEs \eqref{eq:NFESYS} with initial condition $T(0)=0$, admits a unique and locally invertible analytic solution $T(x)$ in a neighborhood of the origin $x=0$. Furthermore, the simultaneous implementation of the nonlinear coordinate transformation: $z=T(x)$ and the state feedback control law: $u=-cz=-cT(x)$ induces the linear closed-loop dynamics:
\begin{equation}
z(t+1)=Az(t),
\end{equation}
whose poles coincide with the eigenvalues of the matrix $A$.}

Please notice, that the initial condition $T(0)=0$ that accompanies the above system of nonlinear functional equations, reflects the fact that under the proposed coordinate transformation, equilibrium properties are preserved. It is worth noting that in the original coordinates, the state feedback law: $u=-cT(x)$ regulates the states of system \eqref{eq:discrete_system} at their nominal equilibrium values due to the invertibility of the map $T(x)$ and the fact that the entire eigenspectrum of the matrix $A$ lies entirely within the unit disc on the complex domain due to Assumption 2 (thus ensuring local asymptotic stability in the Lyapunov sense). Furthermore, the choice of the eigenspectrum and eigenspace of matrix $A$ induces the desirable dynamic modes and characteristics for the controlled system under the above state feedback law. Finally, due to the fact that matrix $A$ is ``adjustable", the set of assumptions of  Theorem 2.1 does not introduce any essential restrictions in the implementation of the proposed method.

The linearization of \eqref{eq:discrete_system} around the equilibrium $(0,0)$ gives:
\begin{equation}
dx(t+1)=\frac{\partial f}{\partial
x}(0,0) dx(t) + \frac{\partial f}{\partial u}(0,0) du(t),
\label{eq:linearizedsystem}
\end{equation}
while the feedback control law in (\ref{eq:linearizedsystem}) around the equilibrium $(0,0)$  is given by:
\begin{equation}
  du(t)=-c\frac{\partial T}{\partial x}(0)dx(t).
\end{equation}
Then, (\ref{eq:linearizedsystem}) can be written as follows:
\begin{equation}
   dx(t+1)=\frac{\partial f}{\partial
x}(0,0) dx(t) - \frac{\partial f}{\partial u}(0,0)c\frac{\partial T}{\partial x}(0)dx(t)=\bigl(\frac{\partial f}{\partial
x}(0,0) -\frac{\partial f}{\partial u}(0,0)c \frac{\partial T}{\partial x}(0)\bigr) dx(t).
   \label{eq:linearizedsystemcontrolled}
\end{equation}
Thus, the linearization of the transformed system (\ref{eq:Az_linearized_sys}) around the equilibrium  reads:
\begin{equation}
    \frac{\partial T}{\partial x}(0)dx(t+1)=A \frac{\partial T}{\partial x}(0)dx(t).
    \label{eq:lin_transformed_around_equilibrium}
\end{equation}
Multiplying both sides of Eq.(\ref{eq:linearizedsystemcontrolled}) by $\frac{\partial T}{\partial x}(0)$, one obtains:
\begin{equation}
  \frac{\partial T}{\partial x}(0)dx(t+1)=\bigg(\frac{\partial T}{\partial x}(0)\frac{\partial f}{\partial x}(0,0) -\frac{\partial T}{\partial x}(0)\frac{\partial f}{\partial u}(0,0)c \frac{\partial T}{\partial x}(0)\bigg)dx(t).
   \label{eq:linearizedsystemcontrolled2}
\end{equation}
Hence, from (\ref{eq:lin_transformed_around_equilibrium}),(\ref{eq:linearizedsystemcontrolled2}), it is inferred that the nonlinear transformation around the equilibrium $(x_0, u_0)$ has to satisfy the following (phase) condition:
\begin{equation}
\frac{\partial T}{\partial x}(0) \frac{\partial f}{\partial x}(0,0)-A\frac{\partial T}{\partial x}(0) =\frac{\partial T}{\partial x}(0)\frac{\partial f}{\partial u}(0,0)c\frac{\partial T}{\partial x}(0).
\label{eq:pinning}
\end{equation}
If $f$ is explicitly known, the elements of the  Jacobian matrix $\frac{\partial T}{\partial x}(0)$ can be calculated analytically.\par 
Here, for learning an approximation of $T(x)$, say $\hat{T}(x)$,  we have used an ANN with two hidden layers and $N_1$, $N_2$ neurons for the first and second hidden layers respectively, as well as a linear output layer, leading to the following equation:
\begin{equation}
    \hat{T}_j(x)=\sum_{i=1}^{N_2} W^{o}_{ij} \phi^{(2)}_i\biggl(\sum_{s=1}^{N_1}W^{(2)}_{si}\phi_s^{(1)}\biggl(\sum_{k=1}^n W_{ks}^{(1)}x_k+\beta^{(1)}_s\biggr)+\beta^{(2)}_i\biggr) +\beta^{(o)}_j,
\end{equation}
or equivalently, in matrix form:
\begin{equation}
   \hat{T}(x) = {W^{(o)}}^T \Phi_2({W^{(2)}}^T \Phi_1({W^{(1)}}^T x+\beta^{(1)})+\beta^{(2)})+\beta^{(0)}.
   \label{eq:FNN}
\end{equation}
$W^{(o)}$ is the $N_2\times n$ matrix containing the weights $W^{(o)}_{ji}$ connecting the second hidden layer to the linear output layer, $\Phi_1:\mathbb{R}^n \rightarrow \mathbb{R}^{N_1}, \Phi_2:\mathbb{R}^{N_1} \rightarrow \mathbb{R}^{N_2}$ denote multivariate vector-valued functions (maps) with components corresponding to activation functions $\phi^{(1)}_s$ and $\phi^{(2)}_i$ of the first and second hidden layers respectively,
$W^{(1)}$ is the $ n \times N_1$ matrix containing the weights $W^{(1)}_{sk}$ from the input to the first hidden layer, $W^{(2)}$ is the $N_1\times N_2$ matrix with the weights $W^{(2)}_{js}$ connecting the first hidden to the second hidden layer,
$\beta^{(1)}\in \mathbb{R}^{N_1}$, $\beta^{(2)}\in \mathbb{R}^{N_2}$ are the column vectors containing the biases $\beta^{(1)}_s$ and $\beta^{(2)}_i$ of the nodes in the first and second layers, respectively, and $\beta^{(o)} \in \mathbb{R}$ is the column vector containing the biases $\beta^{(o)}_j$ of the output nodes.
As has been demonstrated by Chen and Chen \cite{chen1995universal}, such a structure (with sufficient neurons) can approximate, to any accuracy, non-linear laws for the time evolution of dynamical systems.\par 

\begin{figure}[ht!]
    \centering
    \includegraphics[width=1\textwidth]{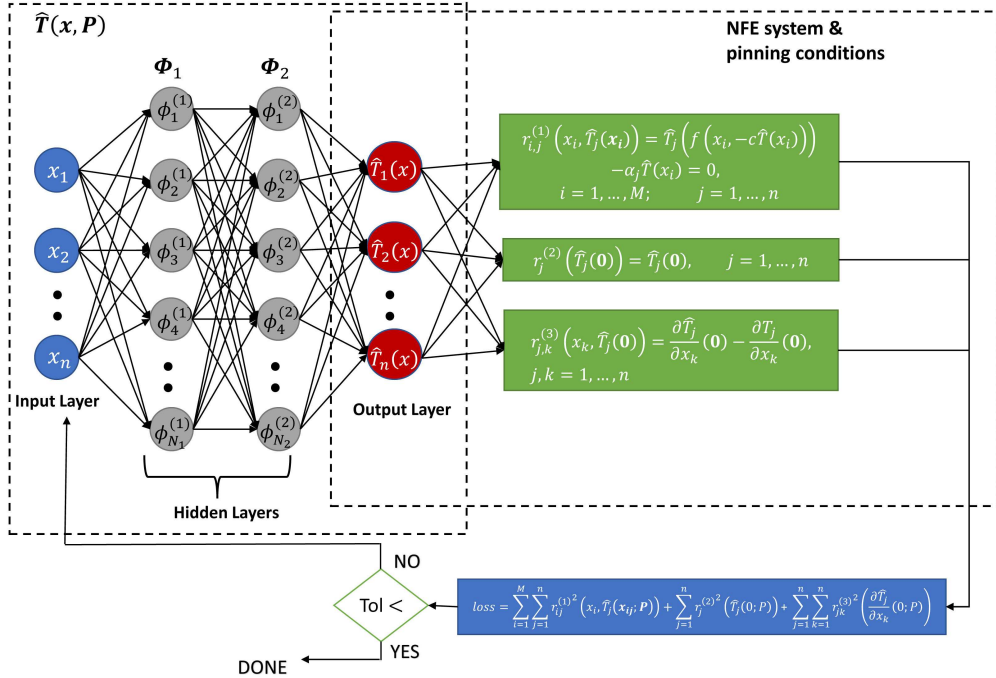}
    \caption{A schematic of the PIML scheme for the feedback linearization of discrete-time systems.}
    \label{fig:PINN diagram}
\end{figure}

Here, for learning the transformation $T(x)$ (for a schematic see also Figure (\ref{fig:PINN diagram}), we considered a certain domain $D\subset\mathbb{R}^n$ around the equilibrium point $(0,0)$ discretized in a grid of $M$ points $x_i$ with $i=1,\dots,M$. Thus, finding  $T(x)$ reduces to the task of minimizing the  loss function:
\begin{gather}
\mathcal{L}(P)=\sum_{i=1}^M \sum _{j=1}^n {r^{(1)}_{ij}}^2(x_{ij},\hat{T}(x_{ij};P))+
\sum _{j=1}^n {r^{(2)}_j}^2( \hat{T}_j(0;P))+
\sum_{j=1}^n \sum _{k=1}^n {r^{(3)}_{jk}}^2(\frac{\partial\hat{T}_{j}}{\partial x_k}(0;P)),
\label{eq:loss1}
\end{gather}
with respect to the unknown parameters $P =({W}^{(0)},{W}^{(2)}, {W}^{(1)}, {\beta}^{(0)},{\beta}^{(2)},{\beta}^{(1)})$ of the FNN  given by (\ref{eq:FNN}). In the above:
\begin{gather}
r^{(1)}_{ij}(x_i, \hat{T}(x_i))= \hat{T}_j\bigg(f(x_i,-c \hat{T}(x_i))\bigg)-{\alpha_j} \hat{T}(x_i), \qquad i=1,\dots,M, \quad j=1,2,\dots,n,
\label{eq:NFE_discretized}
\end{gather}
where ${\alpha_j}$ is the $j$-th row of the matrix $A$ and $\hat{T}_j$ is $j$-th output component of $\hat{T}$, and:
\begin{gather}
r^{(2)}_j(\hat{T}_{j}(0))= \hat{T}_{j}(0),
\quad r^{(3)}_{jk}(x_{k}, \hat{T}_{j}(0))= \frac{\partial \hat{T}_j}{\partial x_k}(0)-\frac{\partial {T}_j}{\partial x_k}(0), \quad j,k=1,2,\dots,n,
\end{gather}
where $\frac{\partial T_j}{\partial x_k}(0)$ is the $(j,k)$-th element of the Jacobian matrix of $T(x)$ computed at the equilibrium, obtained by solving the system of equations in (\ref{eq:pinning}).Notice that for our illustrations, in the loss function, we consider all three terms equally weighted.

Assuming that the objective function in \eqref{eq:loss1} is smooth enough, we may apply an optimization method to solve the least squares problem using (at least) first-order derivatives. To this aim, here we also provide analytically the derivatives with respect to ${x}$ and the parameters ${P}$ of the ANN, i.e., $\frac{\partial \hat{T}_j}{\partial x_k}$, $\frac{\partial \hat{T}_j}{\partial W^{(0)}_{jl}}, j,l=1,2,\dots,N_2$, $\frac{\partial \hat{T}_j}{\partial W^{(2)}_{jl}}, j,l=1,2,\dots,N_1$, $\frac{\partial \hat{T}_j}{\partial W^{(1)}_{jl}}, j,l=1,2,\dots,N_2$.
Note, that these quantities in TensorFlow but also in Matlab can be computed using automatic differentiation, or numerically, using, e.g., centered finite differences, when only a black-box simulator is available.\par
In particular, the analytical derivative w.r.t the $k$-th component of $x=(x_1,x_2,\dots,x_k,\dots,x_n)$ of the first layer of activation function is given by:
\begin{gather}
    \frac{\partial}{\partial x_k} \phi^{(1)}_s \biggl(\sum_{h=1}^n W^{(1)}_{hs}x_h+\beta_s\biggr)=W^{(1)}_{ks}\phi^{(1)'}_s\biggl(\sum_{h=1}^n W^{(1)}_{hs}x_h+\beta_s\biggr),
    \label{eq:firstderivative_firstlayer}
\end{gather}
where $\phi^{(1)'}_s$ is the first derivative of the activation function. Thus, Eq.\eqref{eq:firstderivative_firstlayer} in matrix form reads
\begin{gather}
    \frac{\partial \Phi_1({W}^{(1)T} x+{\beta}^{(1)})}{\partial x_k} = {W_k}^{(1)T} \odot  {\Phi}'_1({W}^{(1)T} x+{\beta}^{(1)}),
\end{gather}
where $\Phi_1'$ is the column vector with the values of the corresponding derivative of the activation functions $\phi_j^{(1)'}$ of the first layer and $W_k^{(1)}$ is the $k$-th row of $W^{(1)}$.
Then, the derivative of the composition of two consecutive layers activation functions is given by:
\begin{gather}
\frac{\partial}{\partial x_k}\phi^{(2)}_i\biggl(\sum_{s=1}^{N_1}W^{(2)}_{si}\phi_s^{(1)}\biggl(\sum_{h=1}^n W_{hs}^{(1)}x_h+\beta^{(1)}_s\biggr)+\beta^{(2)}_i\biggr)=\\ \phi^{(2)'}_i\biggl(\sum_{s=1}^{N_1}W^{(2)}_{si}\phi_s^{(1)}\biggl(\sum_{h=1}^n W_{hs}^{(1)}x_h+\beta^{(1)}_s\biggr)+\beta^{(2)}_i\biggr)\biggl(\sum_{s=1}^{N_1}W^{(2)}_{si}W_{ks}^{(1)}\phi_s^{(1)'}\biggl(\sum_{h=1}^n W_{hs}^{(1)}x_h+\beta^{(1)}_s\biggr)\biggr)\nonumber,
\label{eq:firstderivative_secondlayer}
\end{gather}
where $\phi^{(2)'}_i$ denotes the first derivative of activation functions of the second hidden layer. Thus, Eq. (22) 
in matrix form reads:
\begin{gather}
    \frac{\partial {\Phi}_2({W}^{(2)T}{\Phi}_1({W}^{(1)T}{x}+{\beta}^{(1)})+{\beta}^{(2)})}{\partial x_k}=\\
    ={\Phi}_2'({W}^{(2)T}{\Phi}_1({W}^{(1)T}{x}+{\beta}^{(1)})+{\beta}^{(2)}) \odot \biggl({W}^{(2)T} \cdot \big({W}_{k}^{(1)T} \odot {\Phi}_1'({W}^{(1)T}{x}+{\beta}^{(1)})\big) \biggr)\nonumber,
\end{gather}
where $\Phi_2'$ is the column vector with the values of the corresponding derivative of the activation functions $\phi_j^{(2)'}$ of the second hidden layer, $W_k^{(1)}$ is the $k$-th row of $W^{(1)}$ and the symbol $\odot$ denotes the Hadamard product (element-wise product).
Finally, in order to compute the Jacobian matrix $\frac{\partial \hat{T}}{\partial x}$, let's consider the $j$-th component, say $\hat{T}_j$, of the transformation $\tilde{{T}}=(\hat{T}_1,\hat{T}_2,\dots,\hat{T}_j,\dots,\hat{T}_n)$, so that the element $(j,k)$ of the Jacobian matrix is given by:
\begin{gather}
    \frac{\partial}{\partial x_k}\hat{T}_j(x)=\sum_{i=1}^{N_2} W^{o}_{ij} \phi^{(2)'}_i\biggl(\sum_{s=1}^{N_1}W^{(2)}_{si}\phi_s^{(1)}\biggl(\sum_{h=1}^n W_{hs}^{(1)}x_h+\beta^{(1)}_s\biggr)+\beta^{(2)}_i\biggr)\biggl(\sum_{s=1}^{N_1}W^{(2)}_{si}W_{ks}^{(1)}\phi_s^{(1)'}\biggl(\sum_{h=1}^n W_{hs}^{(1)}x_h+\beta^{(1)}_s\biggr)\biggr),
\end{gather}
and equivalently, in matrix form is expressed as follows:
\begin{gather}
\frac{\partial \hat{T}_j}{\partial x_k}={W}^{j(o)T} \cdot \Biggl({\Phi}_2'({W}^{(2)T}{\Phi}_1({W}^{(1)T}{x}+{\beta}^{(1)})+{\beta}^{(2)}) \odot \biggl({W}^{(2)T} \cdot \big({W}_{k}^{(1)T} \odot {\Phi}_1'({W}^{(1)T}{x}+{\beta}^{(1)})\big) \biggr)\Biggr),
\label{eq:der_network}
\end{gather}
where $W^{j(o)}$ is the $j$-th column of the matrix $W^{(o)}$. 

The derivative of the loss function w.r.t. an unknown parameter, say, $p \in \bm{P}$ is calculated as follows:
\begin{gather}
\frac{\partial \mathcal{L}(\bm{P})}{\partial p}=\sum_{i=1}^M \sum _{j=1}^n {r^{(1)}_{ij}}\frac{\partial r^{(1)}_{ij}}{\partial p}+
\sum _{j=1}^n {r^{(2)}_j}\frac{\partial r^{(2)}_{j}}{\partial p}+
\sum_{j=1}^n \sum _{k=1}^n {r^{(3)}_{jk}}\frac{\partial r^{(3)}_{jk}}{\partial p},
\label{eq:loss}
\end{gather} 
Hence, for the three residuals $r^{(i)},i=1,2,3$, we have:
\begin{gather}
    \frac{\partial r^{(1)}_{ij}}{\partial p}= \frac{\partial \hat{T}_j\bigg(f(x_i,-c^T \hat{T}(x_i))\bigg)}{\partial p}\frac{\partial f(x_i,-c^T \hat{T}(x_i))}{\partial u}\cdot(-c^T\frac{\partial\hat{T}(x_i))}{\partial p})-{\alpha_j}^T \frac{\partial \hat{T}(x_i)}{\partial p}\\
    \frac{\partial r^{(2)}_{j}}{\partial p}= \frac{\partial \hat{T}_j(x_0)}{\partial p},\\
    \frac{\partial r^{(3)}_{jk}(x_{k}, \hat{T}_{j}(\boldsymbol{x}_0))}{\partial p} = \frac{\partial^2 \hat{T}_j}{\partial p \partial x_k}(x_0)\\
i=1,\dots,M, \quad j,k=1,\dots,n, \quad p \in \bm{P}
\end{gather}
In what follows, we compute the derivatives w.r.t. to the weights and biases of the ANN. In particular, one obtains:
\begin{itemize}
    \item for $p=W^{(o)}_{hq}$:
\begin{gather}
    \frac{\partial \hat{T}_j(x)}{\partial W^{(o)}_{hq}}=\begin{Bmatrix} \phi^{(2)}_h\biggl(\sum_{s=1}^{N_1}W^{(2)}_{sh}\phi_s^{(1)}\biggl(\sum_{k=1}^nW_{ks}^{(1)}x_k+\beta^{(1)}_s\biggr)+\beta^{(2)}_h\biggr)& if \qquad q=j\\
        0 & if \qquad q\neq j
    \end{Bmatrix} 
\end{gather}
    \item for $p=W^{(2)}_{hq}$
    \begin{gather}
        \frac{\partial \hat{T}_j(x)}{\partial W^{(2)}_{hq}}=W^{o}_{hj} \phi^{(2)'}_h\biggl(\sum_{s=1}^{N_1}W^{(2)}_{sh}\phi_s^{(1)}\biggl(\sum_{k=1}^n W_{ks}^{(1)}x_k+\beta^{(1)}_s\biggr)+\beta^{(2)}_h\biggr)\phi_q^{(1)}\biggl(\sum_{k=1}^n W_{ks}^{(1)}x_k+\beta^{(1)}_s\biggr)
    \end{gather}
    \item for $p=W^{(1)}_{hq}$
    \begin{gather}
        \frac{\partial \hat{T}_j(x)}{\partial W^{(1)}_{hq}}=\sum_{i=1}^{N_2} W^{o}_{ij} \phi^{(2)'}_i\biggl(\sum_{s=1}^{N_1}W^{(2)}_{si}\phi_s^{(1)}\biggl(\sum_{k=1}^n W_{ks}^{(1)}x_k+\beta^{(1)}_s\biggr)+\beta^{(2)}_i\biggr)\biggl(W^{(2)}_{hi}x_q\phi_h^{(1)'}\biggl(\sum_{k=1}^nW_{kh}^{(1)}x_k+\beta^{(1)}_h\biggr)\biggr)
    \end{gather}
    \item for $p=\beta^{(o)}_h$
    \begin{gather}
        \frac{\partial \hat{T}_j(x)}{\partial \beta^{(o)}_h}=\begin{Bmatrix}
            1 & if \quad h=j\\
            0 & if \quad h\neq j\\
        \end{Bmatrix}
    \end{gather}
    \item for $p=\beta^{(2)}_h$
    \begin{gather}
        \frac{\partial \hat{T}_j(x)}{\partial \beta^{(2)}_h}=W^{o}_{hj} \phi^{(2)}_h\biggl(\sum_{s=1}^{N_1}W^{(2)}_{sh}\phi_s^{(1)}\biggl(\sum_{k=1}^n W_{ks}^{(1)}x_k+\beta^{(1)}_s\biggr)+\beta^{(2)}_h\biggr)
    \end{gather}
    \item for $p=\beta^{(1)}_h$
    \begin{gather}
        \frac{\partial \hat{T}_j(x)}{\partial \beta^{(1)}_h}=\sum_{i=1}^{N_2} W^{o}_{ij} \phi^{(2)'}_i\biggl(\sum_{s=1}^{N_1}W^{(2)}_{si}\phi_s^{(1)}\biggl(\sum_{k=1}^n W_{ks}^{(1)}x_k+\beta^{(1)}_s\biggr)+\beta^{(2)}_i\biggr)\biggl(W^{(2)}_{hi}\phi_h^{(1)}\biggl(\sum_{k=1}^n W_{kh}^{(1)}x_k+\beta^{(1)}_h\biggr)+\beta^{(2)}_i\biggr).
    \end{gather}
\end{itemize}

\section{The Benchmark Problem}
To evaluate the performance of the methods proposed in this work, we have considered the following nonlinear discrete-time system \cite{Kazantzis2001}:
\begin{gather}
x_1(t+1)=exp(0.3x_2(t))\sqrt{(1+x_1(t)+x_2(t))}-1-0.4x_2(t)+0.5u(t) \nonumber \\
x_2(t+1)=0.5\ln(1+x_1(t)+x_2(t))+0.4x_2(t) \label{Discretesystem}
\end{gather}
The Jacobian matrix of the above system at the equilibrium $(0,0)$ is $\frac{\partial f}{\partial x}(0,0)=\begin{bmatrix}
0.5\quad0.4 \\ 
0.5\quad0.9
\end{bmatrix}$, and its eigenvalues $\lambda_1=0.2101$ and $\lambda_2=1.1899$. The matrix $A$ is chosen to be $A=\begin{bmatrix}
0.5 \quad0.3 \\ 
0.5 \quad 0.4
\end{bmatrix}$, with eigenvalues $k_1=0.8405$ and $k_2=0.0595$. Due to the choice of matrix $A$, its eigenvalues are not related to the eigenvalues of the Jacobian matrix $\frac{\partial f}{\partial x}(0,0)$ through any equations of the type \eqref{eq:cond1}, \eqref{eq:cond2}.
Moreover, the following row vector $c$ was chosen: 
\begin{equation}
c=\begin{bmatrix}
1 \quad0 
\end{bmatrix}.
\end{equation}
Please notice, that all conditions of Theorem 2.1. are met by the system (\ref{Discretesystem}), and therefore the associated system of NFEs (3):
\begin{gather}
T_1(exp(0.3x_2(t))\sqrt{(1+x_1(t)+x_2(t))}-1-0.4x_2(t)-0.5T_1,\nonumber\\ 0.5\ln(1+x_1(t)+x_2(t))+0.4x_2(t))=0.5T_1 + 0.3T_2 \nonumber \\
T_2(exp(0.3x_2(t))\sqrt{(1+x_1(t)+x_2(t))}-1-0.4x_2(t)-0.5T_1,\nonumber\\0.5\ln(1+x_1(t)+x_2(t))+0.4x_2(t))=0.5T_1 + 0.4T_2 \nonumber \\
    T_1(0,0) = 0 \nonumber\\
T_2(0,0) = 0,
    \label{NFEsys}
\end{gather}
admits a unique locally analytic and invertible solution around the equilibrium point $(x_1,x_2)=(0,0)$.
Indeed, please notice that
\begin{gather}
\frac{\partial T}{\partial x}(0,0)=\begin{bmatrix}
1\quad1 \\ 
0\quad1
\end{bmatrix},   
\end{gather}
with a $det[T]\neq 0$, results in a locally invertible around the equilibrium point $(x_1,x_2) = (0,0)$ solution, that can be also calculated analytically in closed-form \cite{Kazantzis2001}:
\begin{equation}
T_1(x_1,x_2)=\ln(1+x_1+x_2), \quad T_2(x_1,x_2)=x_2. \label{NonlinTransformation}
\end{equation}
In agreement with Theorem 2.1, the proposed feedback-linearizing and pole-placing nonlinear feedback control law can be explicitly written as follows:
\begin{gather}
 u = -cT(x)= -\ln(1 + x_1 + x_2).   \label{Controllaw}
\end{gather}
Our benchmark problem encompasses a set of singularities of the nonlinear transformation when $x_1+x_2=-1$. Here, we sought to learn the feedback-linearizing control law in the domain $[x_L,0] \times [x_L,0] =[-0.495,0]\times[-0.495,0]$.\par
\begin{figure}[htbp]
 \centering
 \begin{subfigure}[h]{0.45\textwidth}
     \centering
     \includegraphics[width=\textwidth]{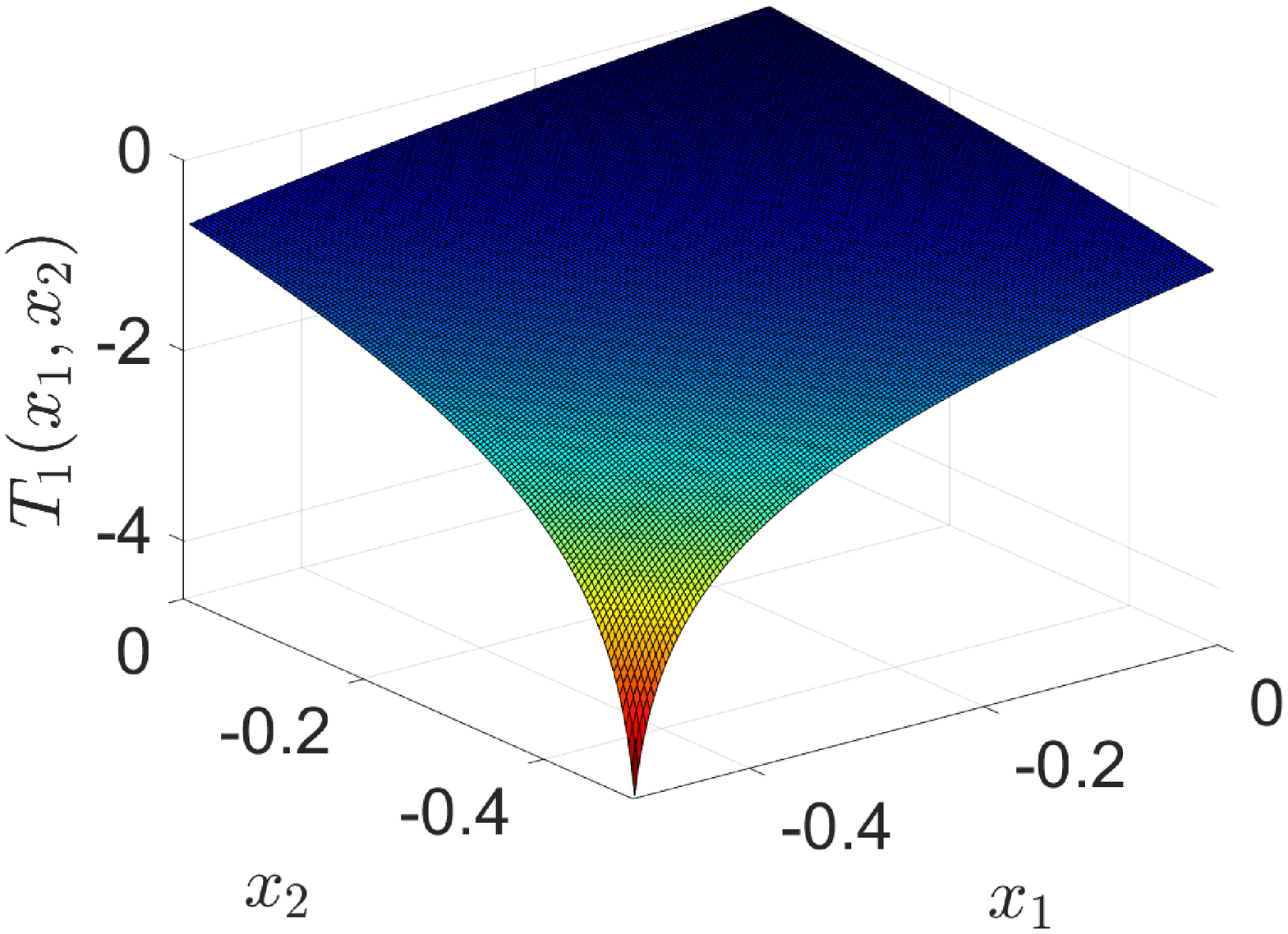}
     \caption{}
 \end{subfigure}
 \hfill
 \begin{subfigure}[h]{0.45\textwidth}
     \centering
     \includegraphics[width=\textwidth]{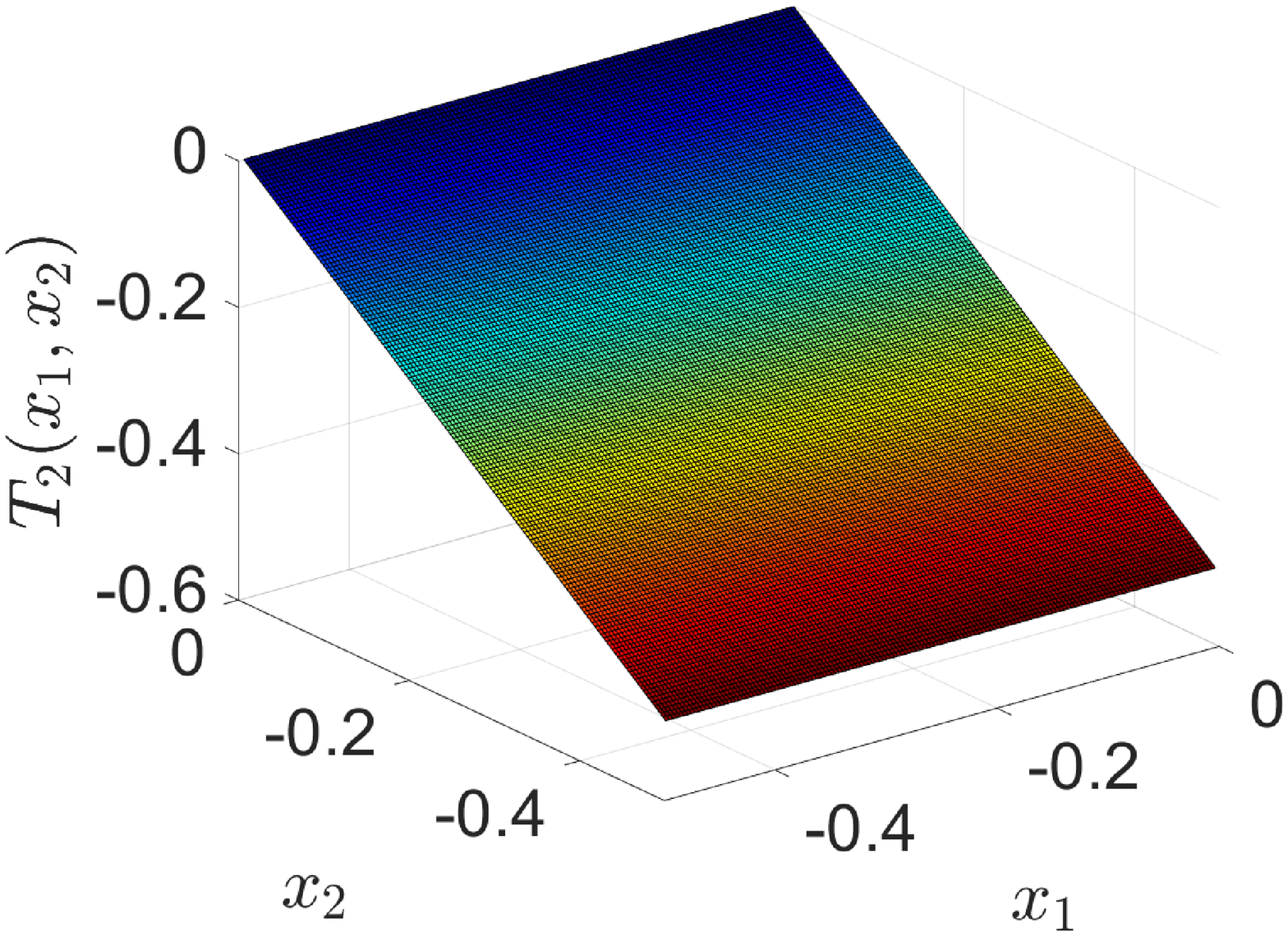}
     \caption{}
 \end{subfigure}
    \caption{Analytical solution of the NFEs (\ref{NFEsys}) in $[-0.495,0]\times [-0.495,0] $. (a) $T_1(x_1,x_2)=\ln(1+x_1+x_2)$. A steep-gradient at $(-0.495,-0.495)$ is due to the presence of a singular point at $(x_1,x_2)=(-0.5,-0.5)$. (b) $T_2(x_1,x_2)=x_2$.}.
  \label{fig:sX}
\end{figure}

 
Figure (\ref{fig:sX}) depicts the analytical solutions $T_1(x_1,x_2), T_2(x_1,x_2)$, of the associated system of NFEs. 
We note that it encompasses the solution domain, which has been deliberately chosen to be $x_1,x_2\in [-0.495,0]$ since $T_1(x_1,x_2)$ exhibits a singular point at $(x_1,x_2)=(-0.5,-0.5)$. The first reasonable step is, therefore, to
verify by comparing with the analytical solution, the numerical approximation accuracy of the proposed PIML scheme in this region and assess its impact on the performance profile of the resulting feedback-linearizing control law.

\section{Numerical results}
We present the numerical results in two subsections. In subsection (4.1), we provide numerical results for the case when the discrete nonlinear model \eqref{eq:discrete_system} is assumed to be explicitly available. In this case, one can analytically as above (or exploiting the automatic differentiation toolkit) compute the necessary derivatives required in the optimization algorithm. Within the greedy PIML framework, we implemented two schemes: (a) a home-made code in Matlab, using the Levenberg-Marquardt algorithm for training; and, (b) the Keras API of TensorFlow library running in Python using the BFGS optimization algorithm; this scheme uses by default automatic differentiation to compute the necessary derivatives. To assess the performance of the greedy approach, we also show the results obtained with Matlab and TensorFlow, when the PIML was trained in the entire domain at once.
In subsection (4.2), we present the results assuming that the discrete nonlinear model \eqref{eq:discrete_system} is not explicitly available, but we have access to a black-box simulator. In this case, the necessary derivatives are estimated numerically using centered finite differences.\par
For training purposes using the greedy approach, we considered $20$ equispaced distributed collocation points for each one of the two inputs $x_1$ and $x_2$, i.e. we used a grid of $20 \times 20$ points equispaced distributed for each step of the greedy approach. Different (denser) sizes of the grid (e.g. using a grid of $100\times 100$ points) with more neurons for each layer did not affect qualitatively the results and corresponding performances, as we also show. Thus for the greedy-approach, we started by considering a grid in $[-0.2,0]\times[-0.2,0]$. Then with a step of $-0.05$, we used as initial guesses of the unknown weights and biases of the PIML the ones obtained from the training procedure from the previous grid, and repeated training until the interval $[-0.45,0]\times [-0.45,0]$ was reached. From this interval and on, we optimized iteratively the PIML using progressively bigger intervals with a step of $-0.01$, for each of the two inputs $x_1$ and $x_2$, up to the interval $[-0.49,0]\times[-0.49,0]$. After this interval we augmented progressively the grid with steps of $-0.001$ until the entire domain of interest, $[-0.495,0]\times[-0.495,0]$, was reached. As mentioned before, additionally, we have also used the TensorFlow library to learn the transformation law both with the greedy approach and in the entire domain $[-0.495,0]\times[-0.495,0]$. The performances of the PIML schemes were also compared with a $6th$ order power-series expansion of  $T_1(x_1,x_2), T_2(x_1,x_2)$, thus resulting in equivalent, to the PIML, number of unknowns.\par\par 
For testing purposes, we used the roots of the Chebyshev polynomial of the first kind of degree $k$. In particular, in the interval $[a, b]$ with $a,b\in \mathbb{R}$, the grid was created using the following Chebyshev collocation points:
\begin{gather}
x_n=\frac{1}{2}(a+b)-\frac{1}{2}(a-b)cos\left ( \frac{2n-1}{2n}\pi \right )\quad n=1,\cdots,k \label{Cheby}
\end{gather}
with $a=-0.495$ and $b=0$, and therefore equation (\ref{Cheby}) can now be expressed as follows:
\begin{gather}
x_n=\frac{1}{2}(-0.495)-\frac{1}{2}(-0.495)cos\left ( \frac{2n-1}{2n}\pi \right ))\quad n=1,\cdots,k\label{Cheby300}
\end{gather}
For our illustrations, we selected a grid of $50 \times 50$ Chebyshev points for each of the intervals in $[-0.495,0]\times[-0.495,0]$.

\subsection{The case of the explicitly available model}

\paragraph{Power-series solution.}First, we have expanded both the right-hand-side of the model \eqref{eq:discrete_system} and the nonlinear transformation $T(x_1,x_2)$ to a $6th$ order power-series, and equated same order terms up to the $6th$ order of $T(f(x,-cT(x)))$ and $AT(x)$ on both sides of the associated NFEs (see Appendix \ref{sec:appendix_A}), thus building a system of algebraic equations that can be solved for the coefficients of the $6th$ order polynomial approximation. For this task, we used Matlab's symbolic toolbox. 
\paragraph{PIML-based solution.} For our PIML scheme, we constructed an ANN with two hidden layers, fully connected, with five neurons in each layer. More specifically, for both the home-made Matlab code and the TensorFlow (TF) implementation, the activation function was chosen to be the $sigmoid(x)$ function. For the training, in the Matlab implementation, we used the function \textit{lsqnonlin}, which implements the Levenberg-Marquart (LM) algorithm. In the LM scheme, we have set as stopping criterion a threshold of  $\text{FuncTol}=10^{-12}$. Moreover, we have set a maximum number of $100,000$ iterations and a maximum number of function iterations equal to $12,000$. Finally, to initialize the weights and biases, we have used uniformly distributed random numbers in the interval $[0,1]$ by employing the \textit{rand} function of Matlab.\par
For our computations, we have also used the Keras API of TensorFlow. Thus, we have used automatic differentiation (AD) \cite{AutomaticDT} to find the required derivatives for the optimization process. With the TensorFlow, we have tried different optimizers; namely the Stochastic gradient descent (SGD), the ADAM optimizer and the BFGS optimizer. For the SGD and ADAM optimizers, we have used the default values, except for the learning rate which has been selected using a piecewise constant decay varying from $10^{-2}$ to $10^{-4}$ as the number of epochs increased. We have chosen $100,000$ epochs since using more epochs did not seem to considerably improve the results derived, whereas for the BFGS  we have used the library \textit{tensorflow probability}. Furthermore, we have used a maximum number of iterations equal to $100,000$, and a  stopping condition of $\text{FuncTol}=10^{-16}$. The best results obtained with the TF were derived with the aid of BFGS, and these are the results presented in our comparative assessment, shown below.\par
\begin{figure}[htbp]
 \centering
 \begin{subfigure}[h]{0.45\textwidth}
     \centering
     \includegraphics[width=5.8cm]{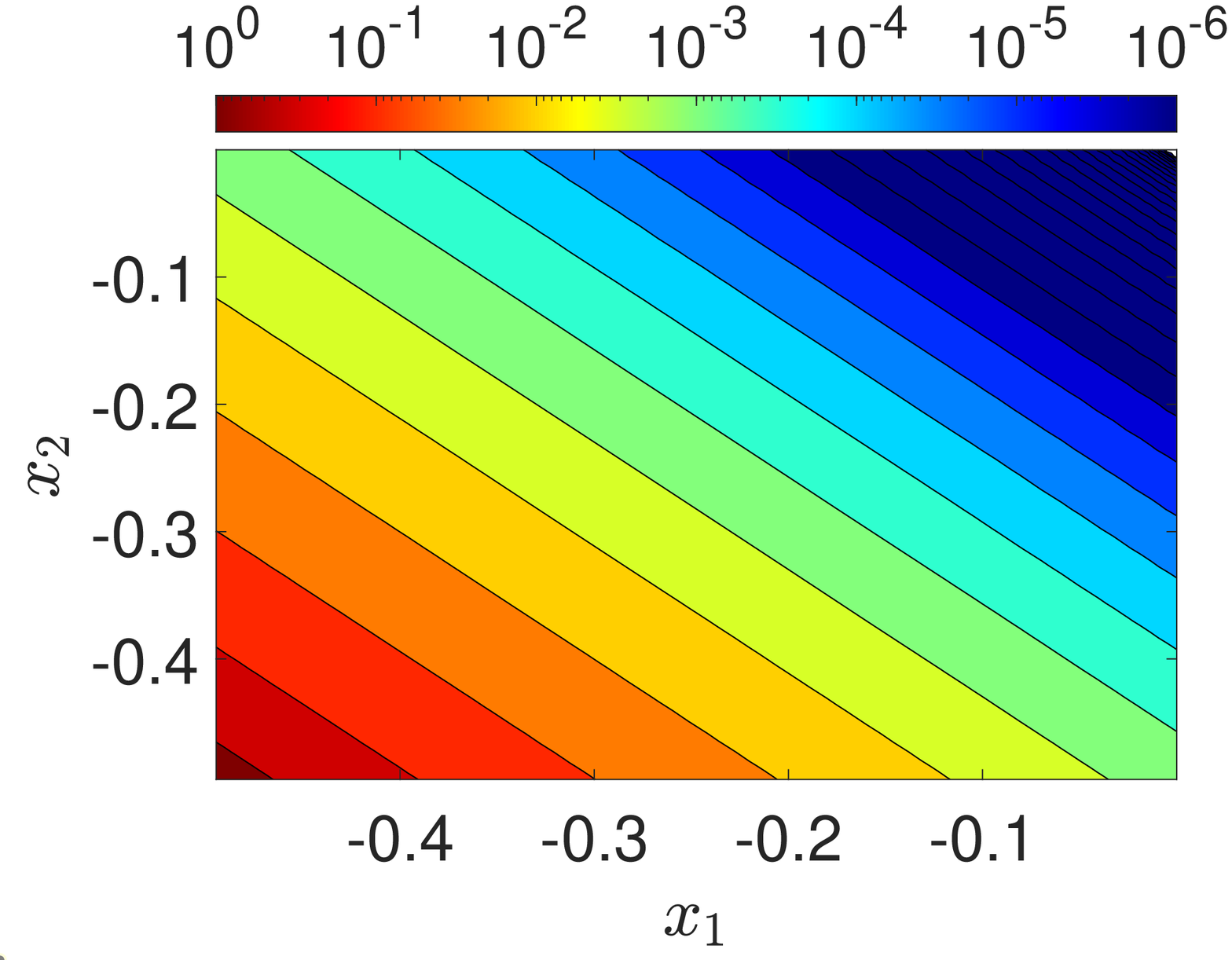}
     \caption{}
 \end{subfigure}
 \hfill
 \begin{subfigure}[h]{0.45\textwidth}
     \centering
     \includegraphics[width=5.8cm]{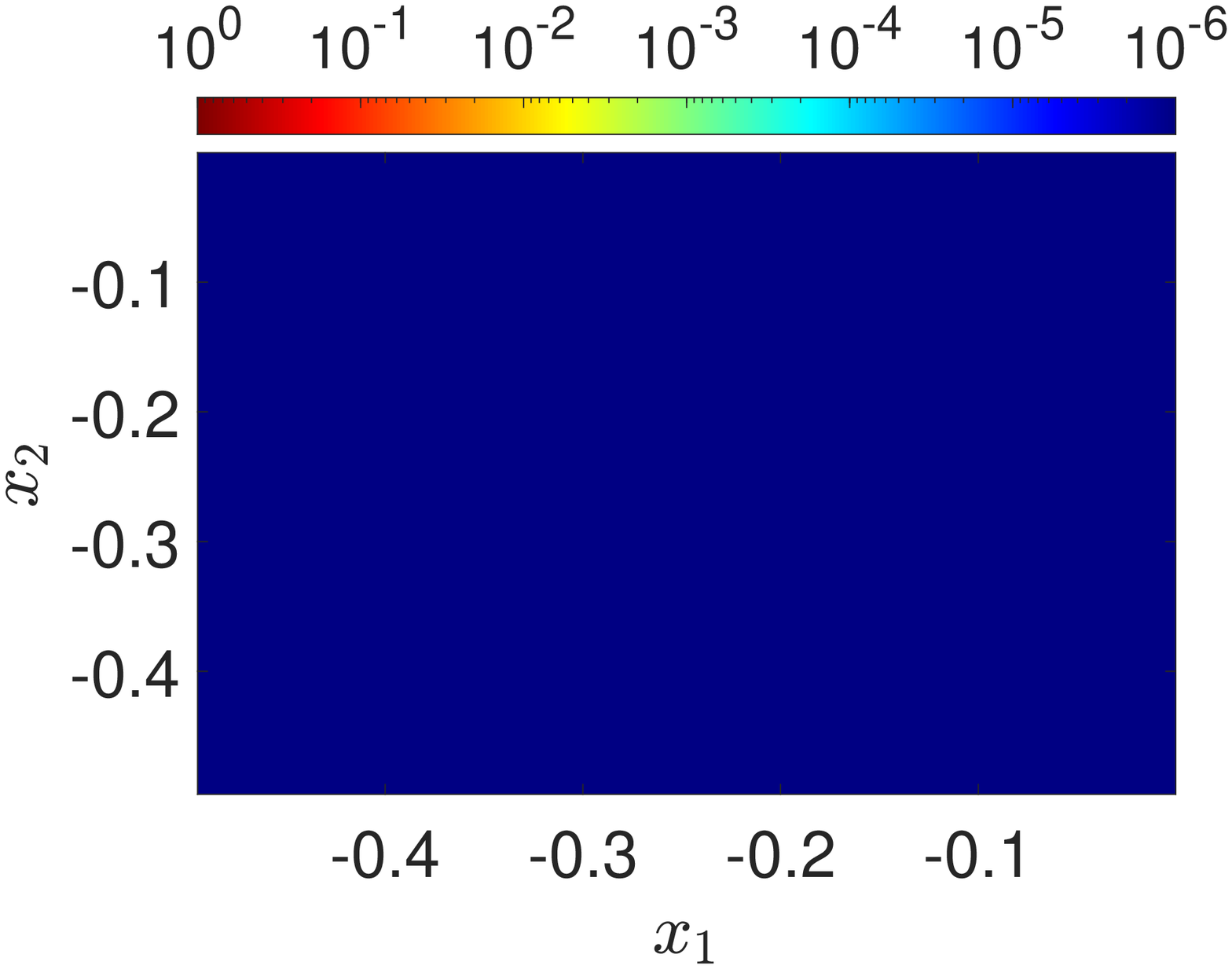}
     \caption{}
 \end{subfigure}
 \begin{subfigure}[h]{0.45\textwidth}
     \centering
     \includegraphics[width=5.8cm]{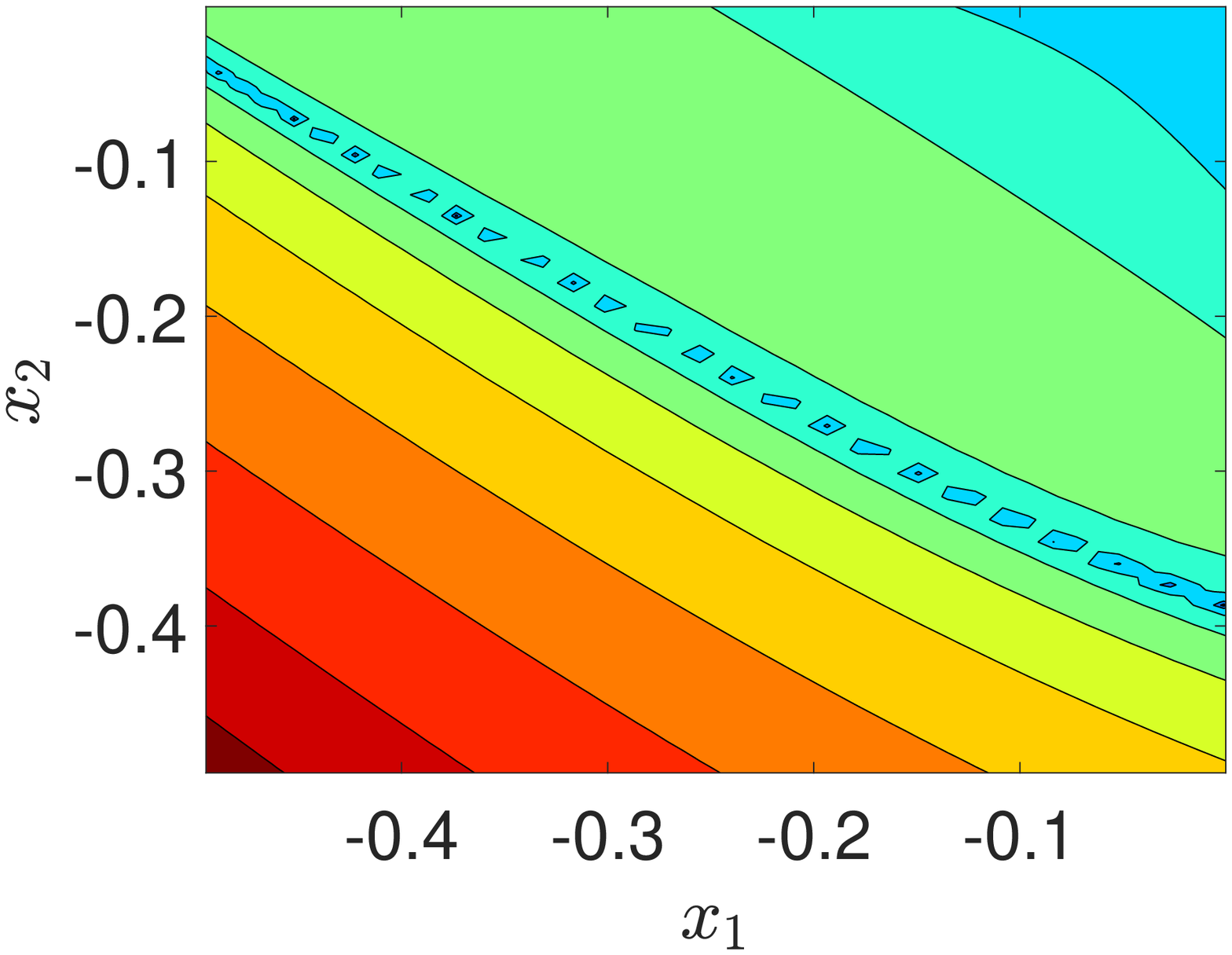}
     \caption{}
 \end{subfigure}
 \hfill
 \begin{subfigure}[h]{0.45\textwidth}
     \centering
     \includegraphics[width=5.8cm]{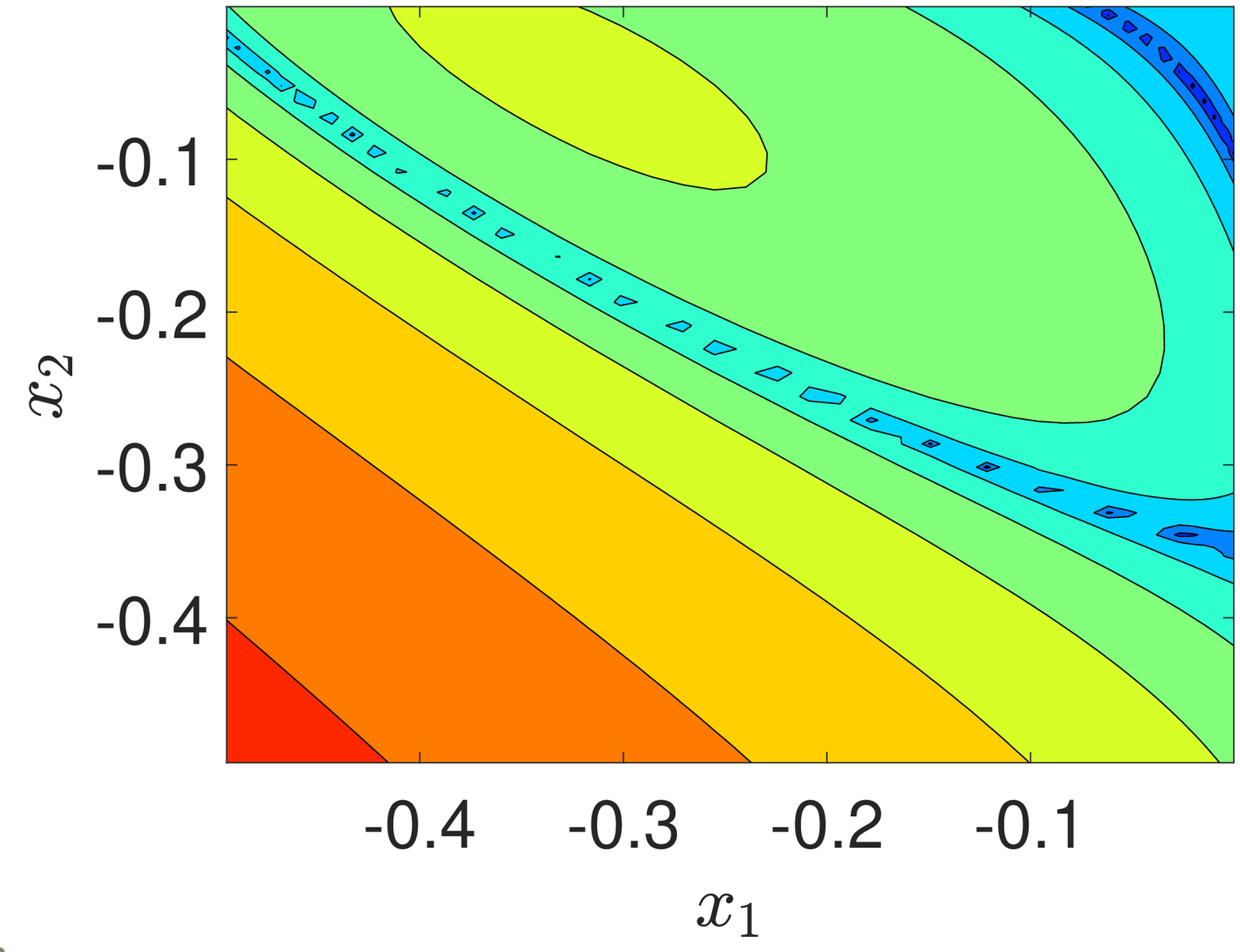}
     \caption{}
 \end{subfigure}
 \centering
 \begin{subfigure}[h]{0.45\textwidth}
     \centering
     \includegraphics[width=5.8cm]{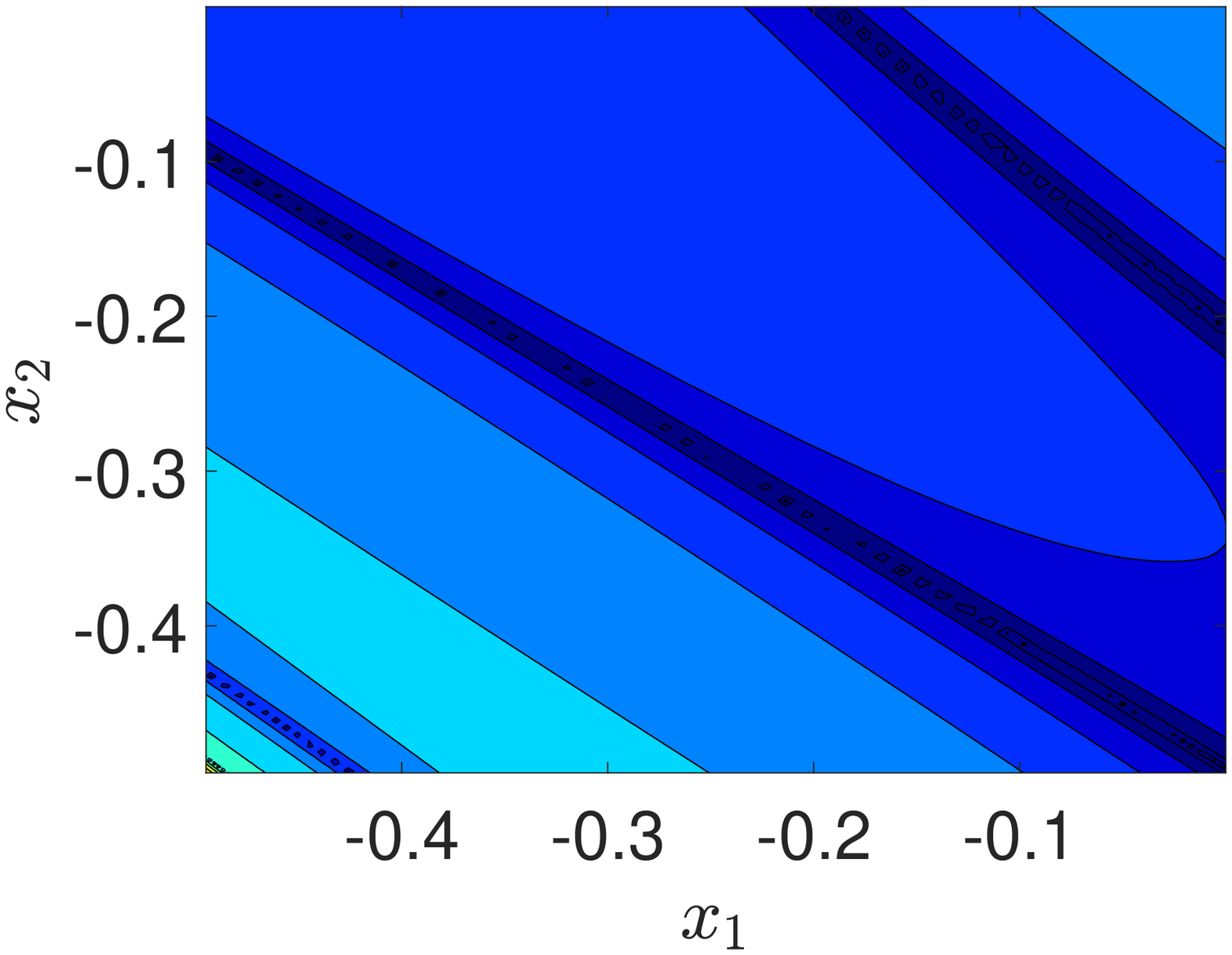}
     \caption{}
 \end{subfigure}
 \hfill
 \begin{subfigure}[h]{0.45\textwidth}
     \centering
     \includegraphics[width=5.8cm]{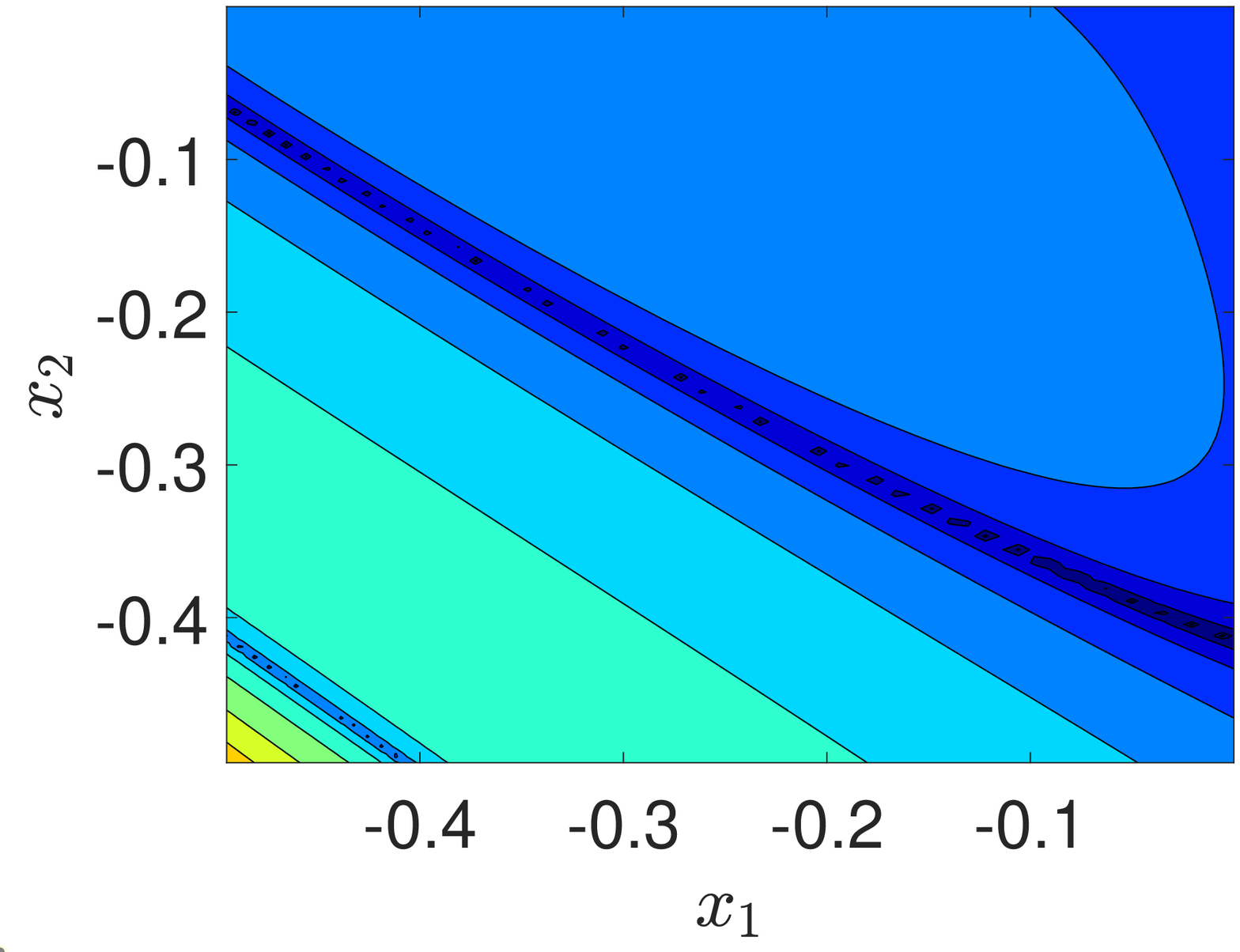}
     \caption{}
 \end{subfigure}
   \centering
 \begin{subfigure}[h]{0.45\textwidth}
     \centering    \includegraphics[width=5.5cm]{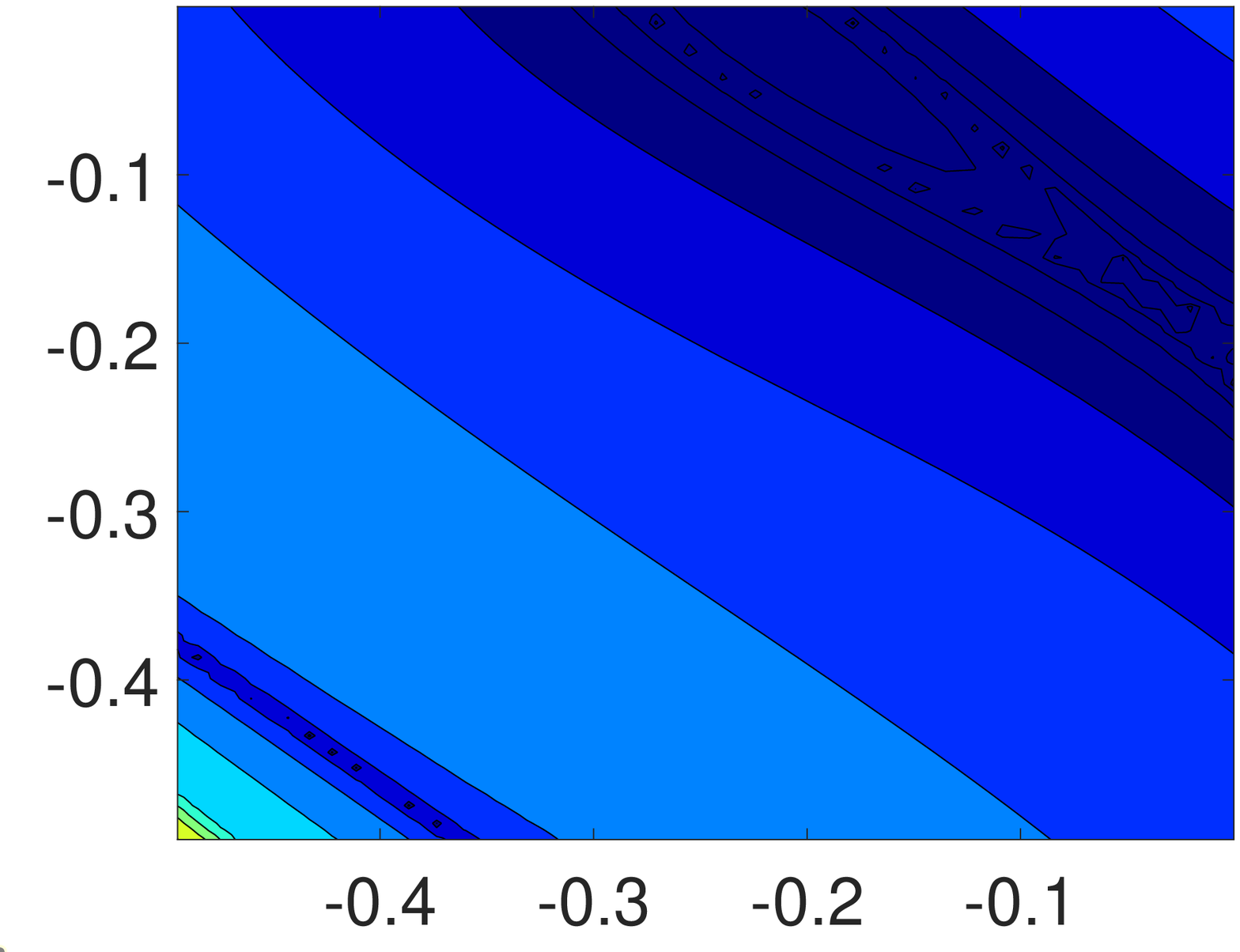}
     \caption{}
 \end{subfigure}
 \hfill
 \begin{subfigure}[h]{0.45\textwidth}
     \centering
     \includegraphics[width=5.5cm]{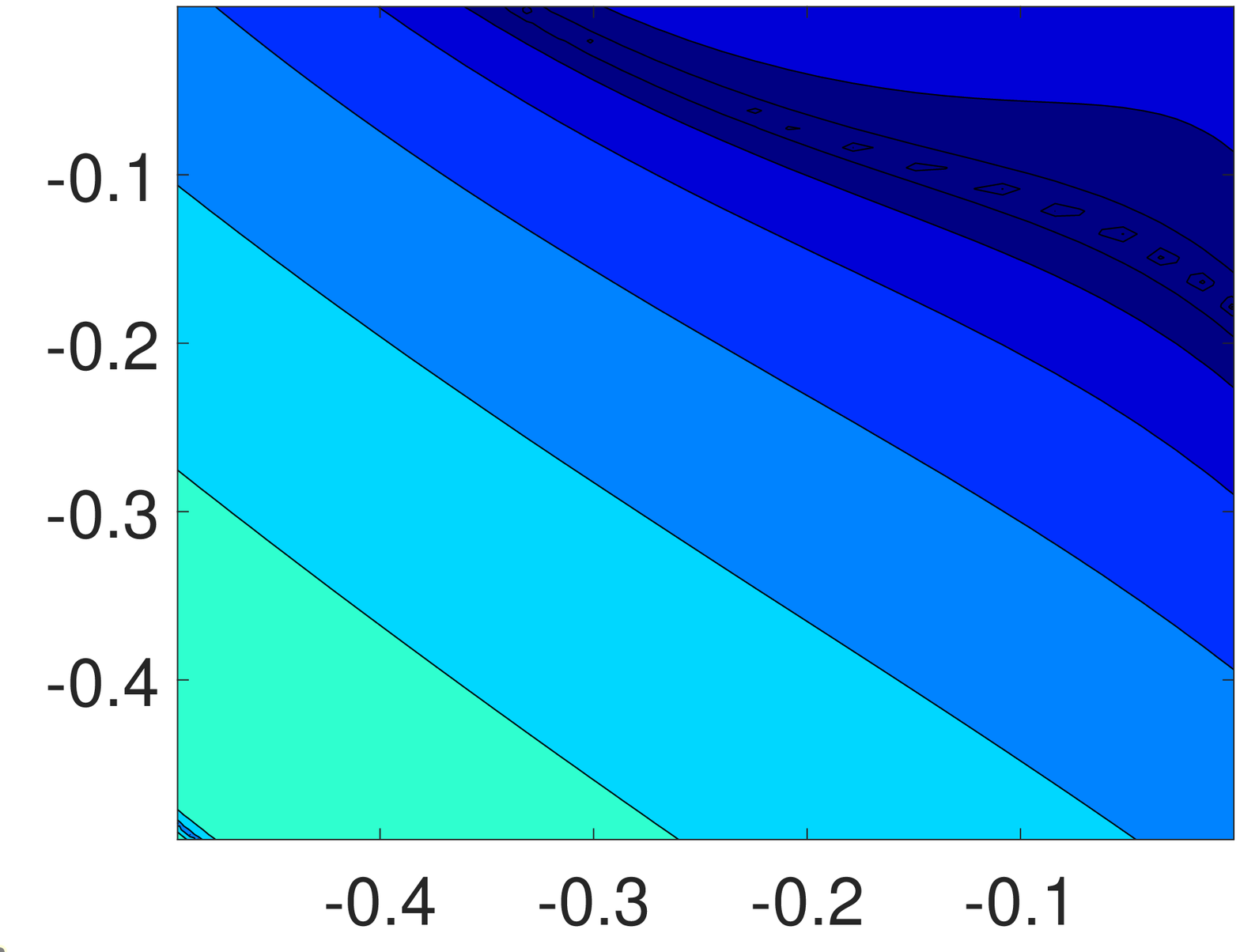}
     \caption{}
 \end{subfigure}
\caption{Model explicitly available. Test sets (grids of $50 \times 50$ Chebyshev-distributed points). Numerical approximation accuracy (difference between the computed and analytical solution) of $T_1(x_1,x_2)$ (left column) and $T_2(x_1,x_2)$ (right column) using the various schemes. (a),(b) $6th$ order power-series expansion of $T_1(x_1,x_2)$ and $T_2(x_1,x_2)$ and the right-hand side of the model (\ref{Discretesystem}) in $[-0.495,0]\times[-0.495,0]$. (c),(d) PIML in Tensorflow trained in the entire domain $[-0.495,0]\times[-0.495,0]$. (e),(f) PIML in Tensorflow trained via the greedy-wise approach. (g),(h) PIML in Matlab trained via the greedy-wise approach.}
\label{fig:Modelavailable_TS}
\end{figure}

%
Figure (\ref{fig:Modelavailable_TrS}) shows the numerical approximation accuracy (difference between computed and analytical solutions) obtained by the various schemes for the training set. Figure (\ref{fig:Modelavailable_TrS})(a),(b) show the results obtained with a $6th$ order power-series expansion of the nonlinear transformation and the right-hand-side of the discrete model. As expected, the power-series expansion results in zero error for the $T_2(x_1,x_2)$ component, and in a good approximation accuracy for the $T_1(x_1,x_2)$ component, but only in the region close to the linearization-relevant point, which in this case is $[0,0]$, while away from it, the numerical approximation is poor (of the order of $10^0$ at the edge of the grid) close to the singular point. Figures (\ref{fig:Modelavailable_TrS})(c),(d) depict the results obtained with the PIML implemented in Tensorflow using the BFGS optimizer for learning the non-linear transformation law in the entire domain. The approximation accuracy of the scheme is rather poor, especially near the singularity  (of the order of $10^{0}$). Figures (\ref{fig:Modelavailable_TrS})(e),(f) depict the results obtained from the PIML implemented with the Tensorflow and the BFGS optimizer using the greedy-wise training procedure.
Figures (\ref{fig:Modelavailable_TrS})(g),(h) depict the results obtained with the PIML implemented in Matlab using the LM optimizer and the greedy-wise training procedure. It is evident that the greedy-wise training procedure results in significantly enhanced numerical approximation accuracy compared to the PIML schemes trained in the entire domain at once. In particular, the PIML scheme implemented in TensorFlow resulted in an numerical approximation error of the order of $10^{-2}$ and the home-made Matlab resulted in a numerical approximation error of the order of $10^{-3}$, in the region close to the edge point $(-0.495,-0.495)$. In addition, Figure (\ref{fig:Modelavailable_TS}) depicts the performance of the schemes tested on a Chebyshev grid. The results are similar for the training  and the test set. Table \ref{tab:NormsS1} and Table \ref{tab:NormsS2} detail the numerical approximation accuracy of the various schemes for the training and the test sets, respectively.
\begin{figure}[htbp]
 \centering
 \begin{subfigure}[h]{0.45\textwidth}
     \centering
     \includegraphics[width=5.5cm]{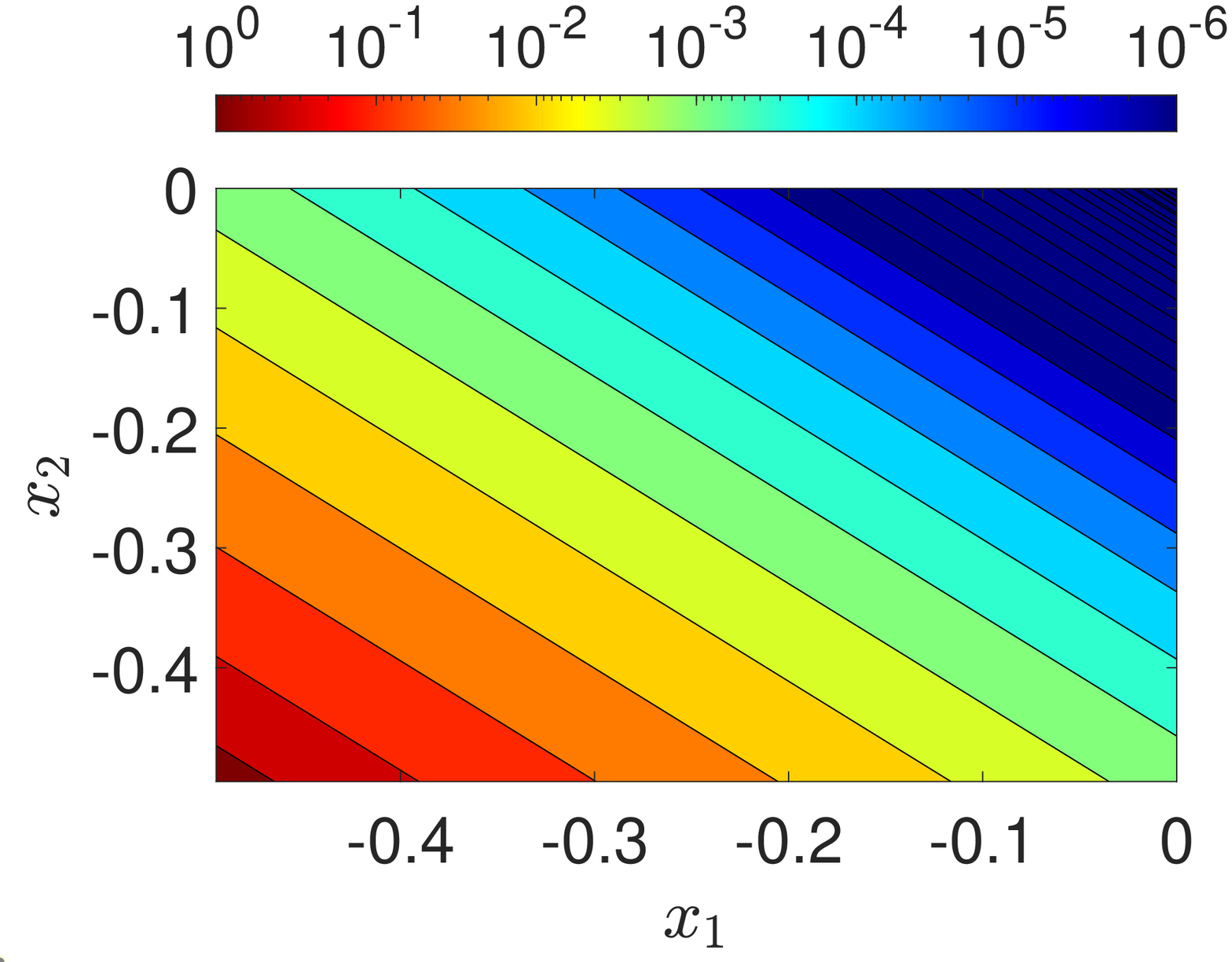}
     \caption{}
 \end{subfigure}
 \hfill
 \begin{subfigure}[h]{0.45\textwidth}
     \centering
     \includegraphics[width=5.5cm]{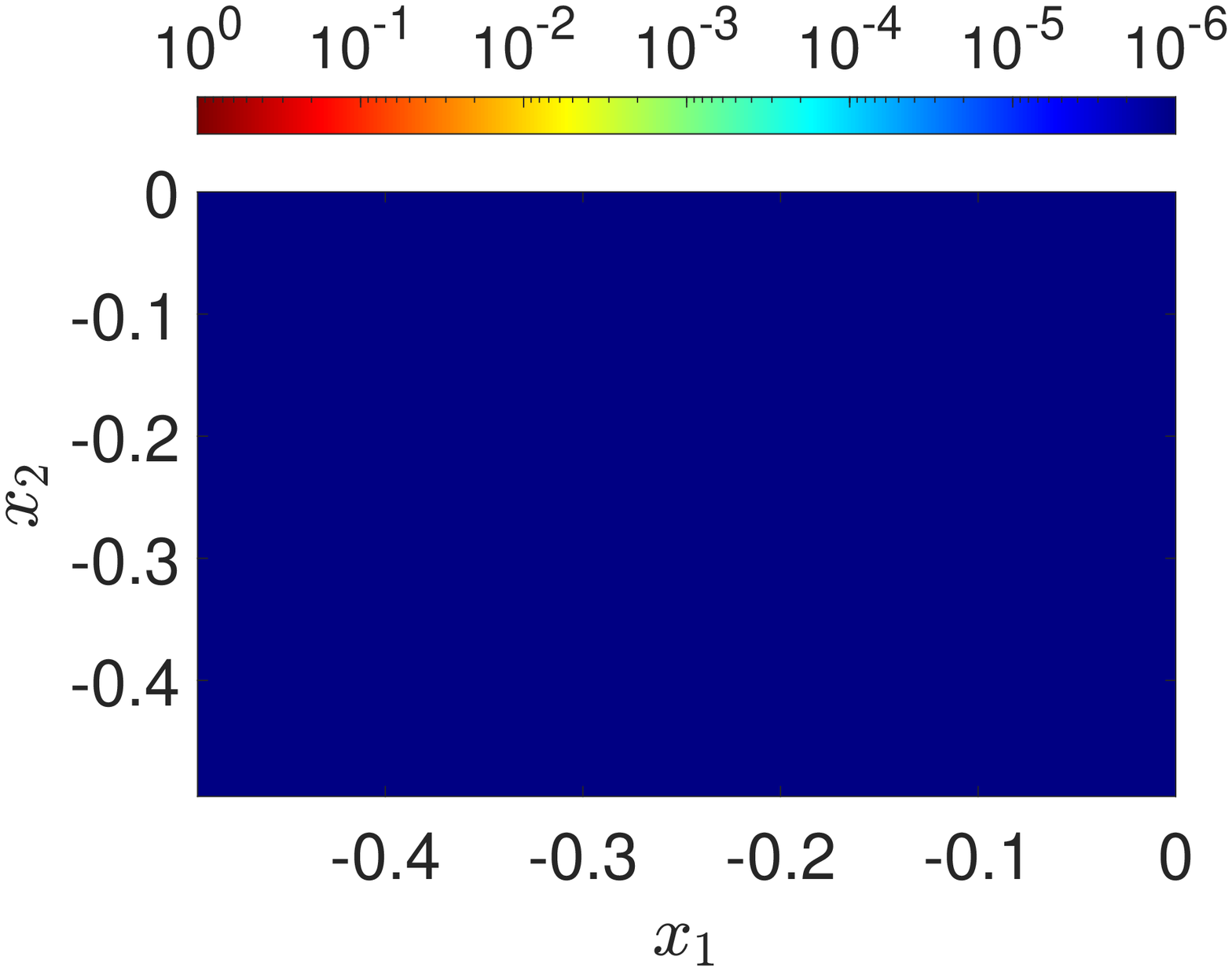}
     \caption{}
 \end{subfigure}
 \begin{subfigure}[h]{0.45\textwidth}
     \centering
     \includegraphics[width=5.4cm]{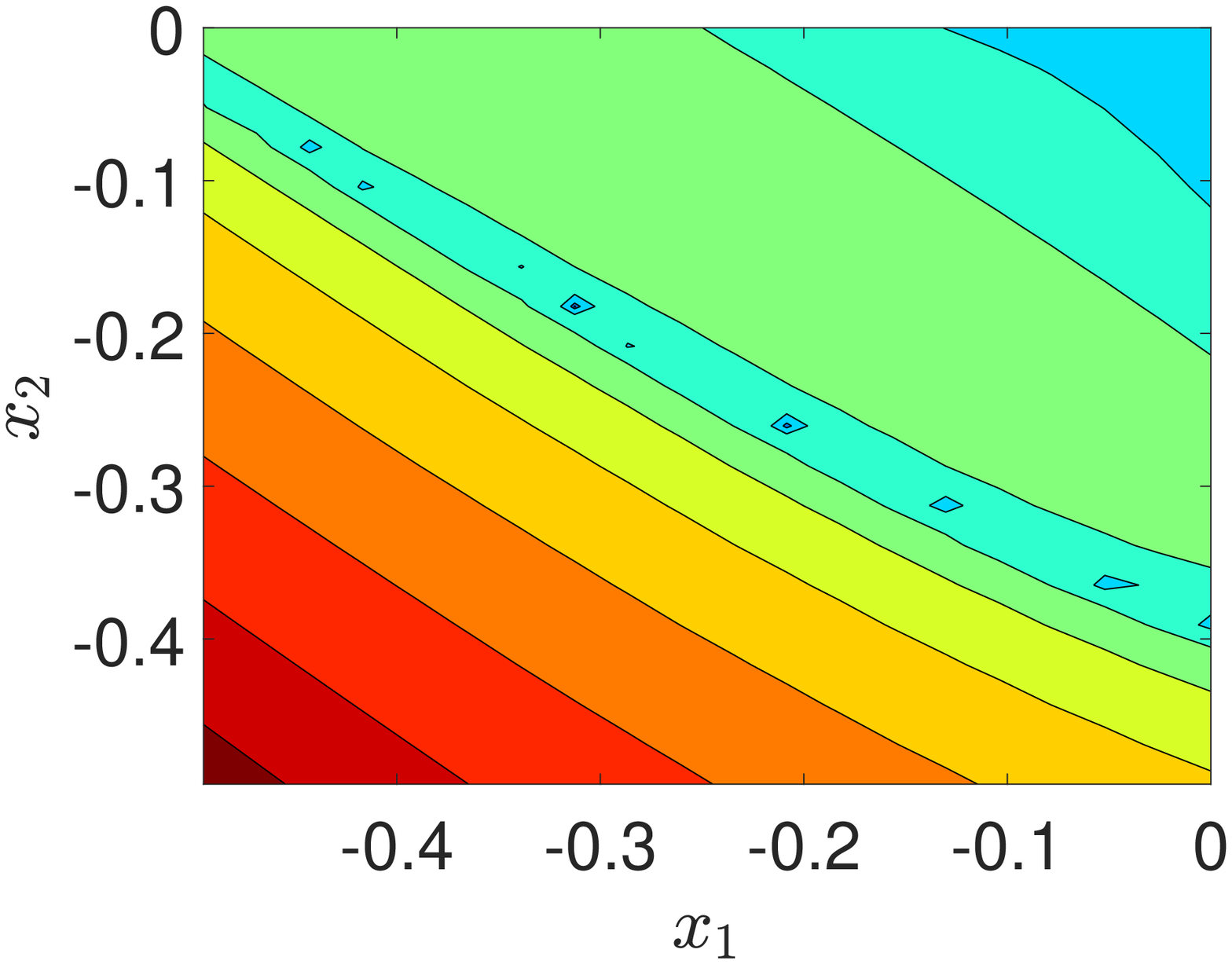}
     \caption{}
 \end{subfigure}
 \hfill
 \begin{subfigure}[h]{0.45\textwidth}
     \centering
     \includegraphics[width=5.4cm]{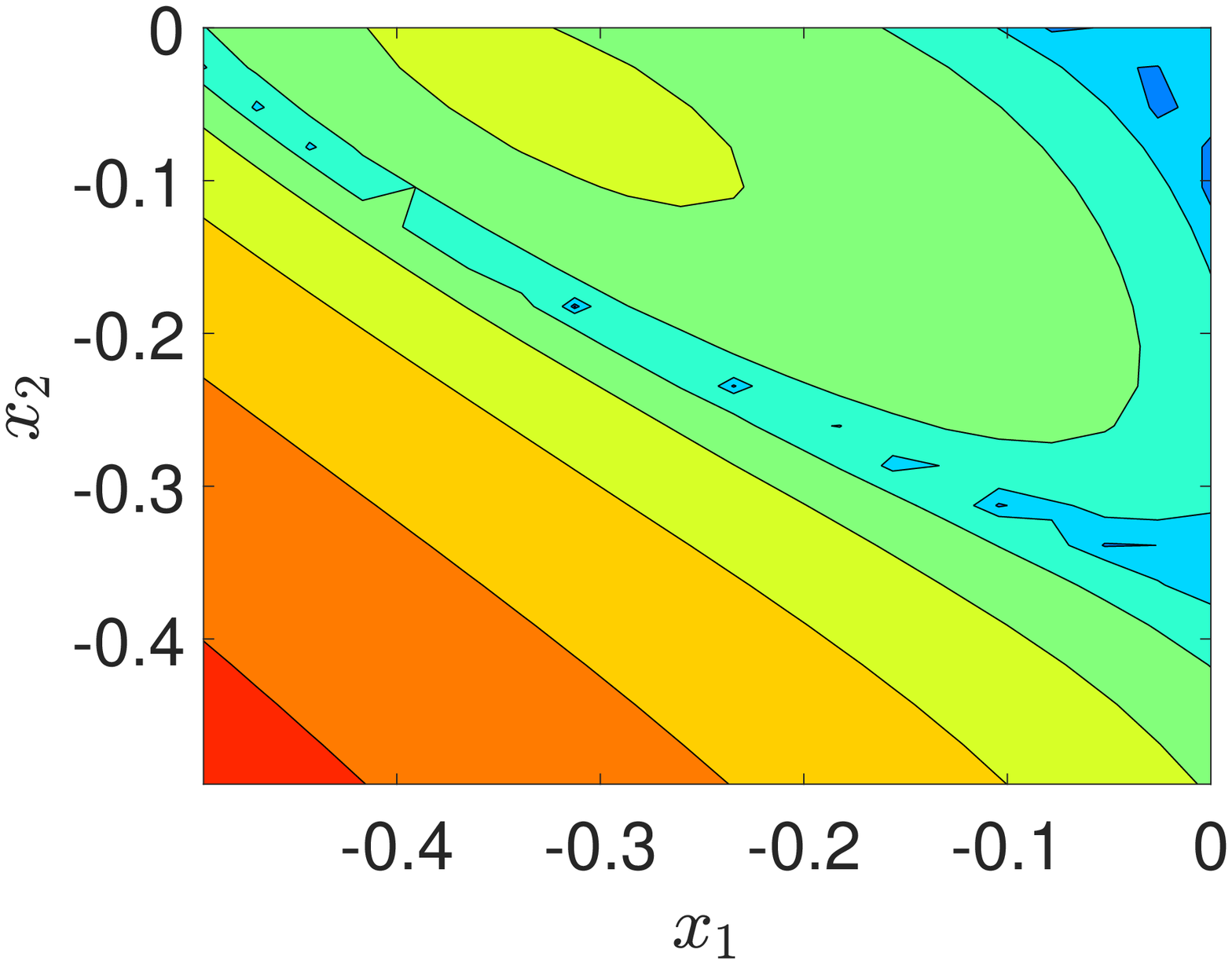}
     \caption{}
 \end{subfigure}
   \centering
 \begin{subfigure}[h]{0.45\textwidth}
     \centering
     \includegraphics[width=5.4cm]{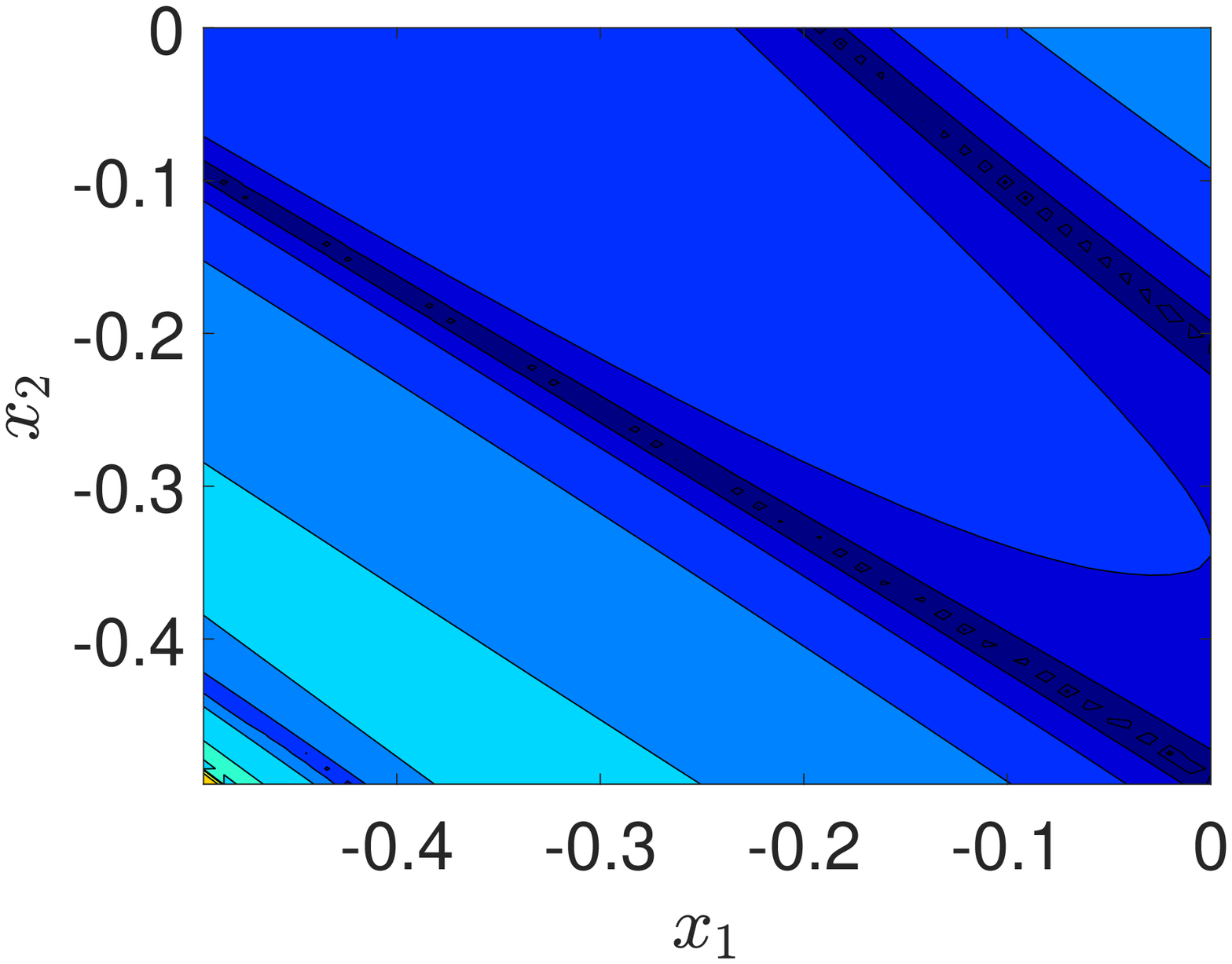}
     \caption{}
 \end{subfigure}
 \hfill
 \begin{subfigure}[h]{0.45\textwidth}
     \centering
     \includegraphics[width=5.4cm]{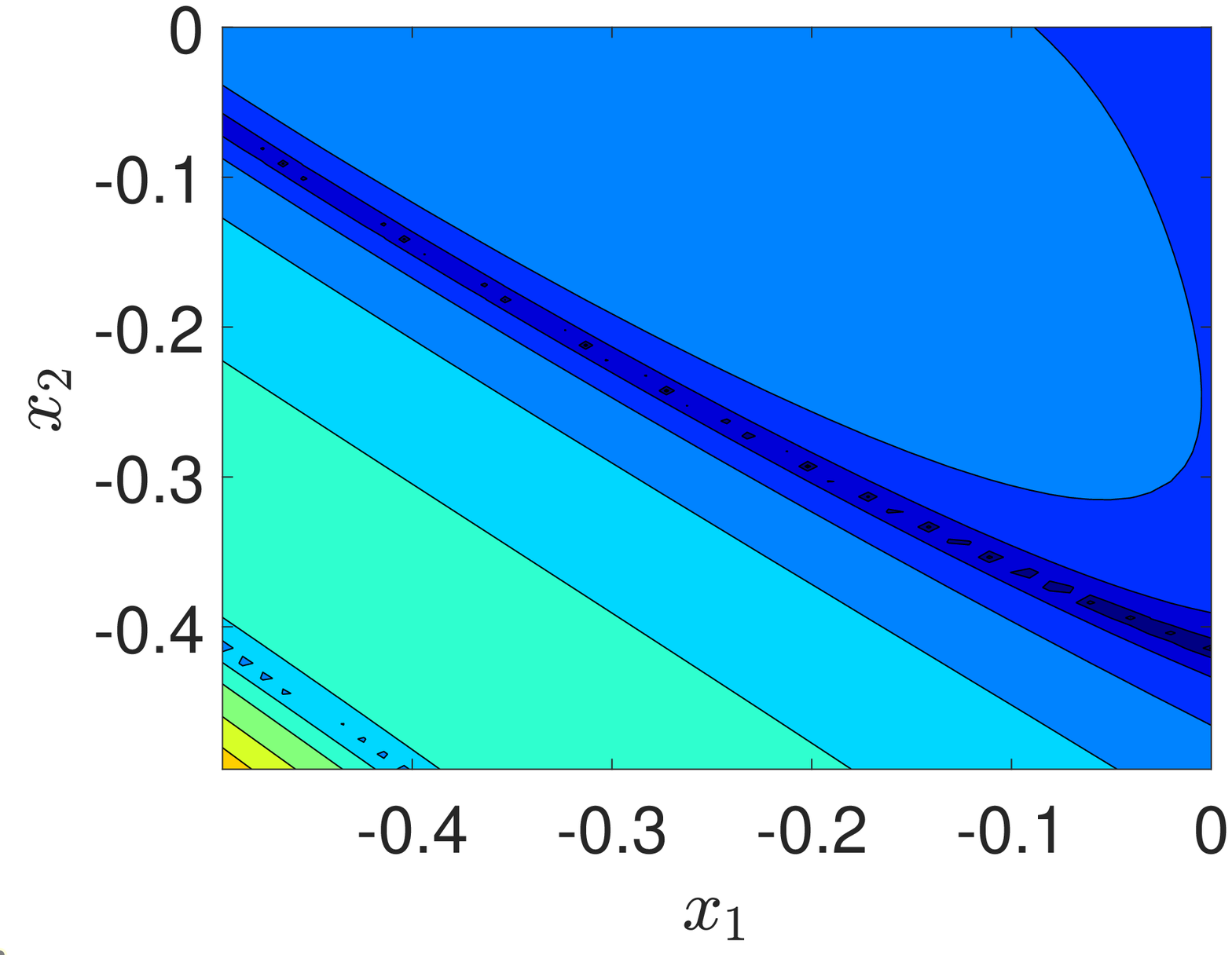}
     \caption{}
 \end{subfigure}
 \begin{subfigure}[h]{0.45\textwidth}
     \centering
     \includegraphics[width=5.4cm]{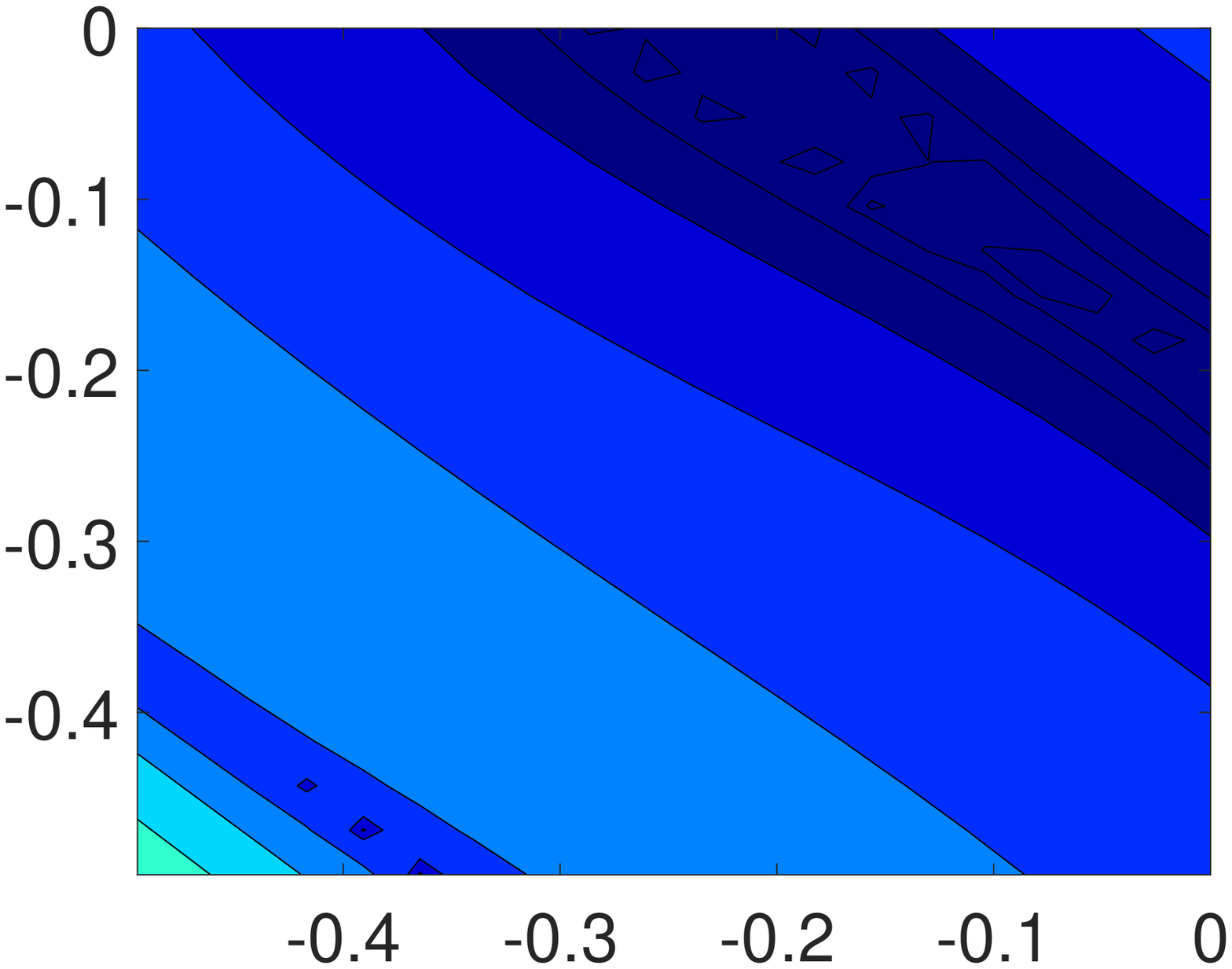}
     \caption{}
 \end{subfigure}
 \hfill
 \begin{subfigure}[h]{0.45\textwidth}
     \centering
     \includegraphics[width=5.4cm]{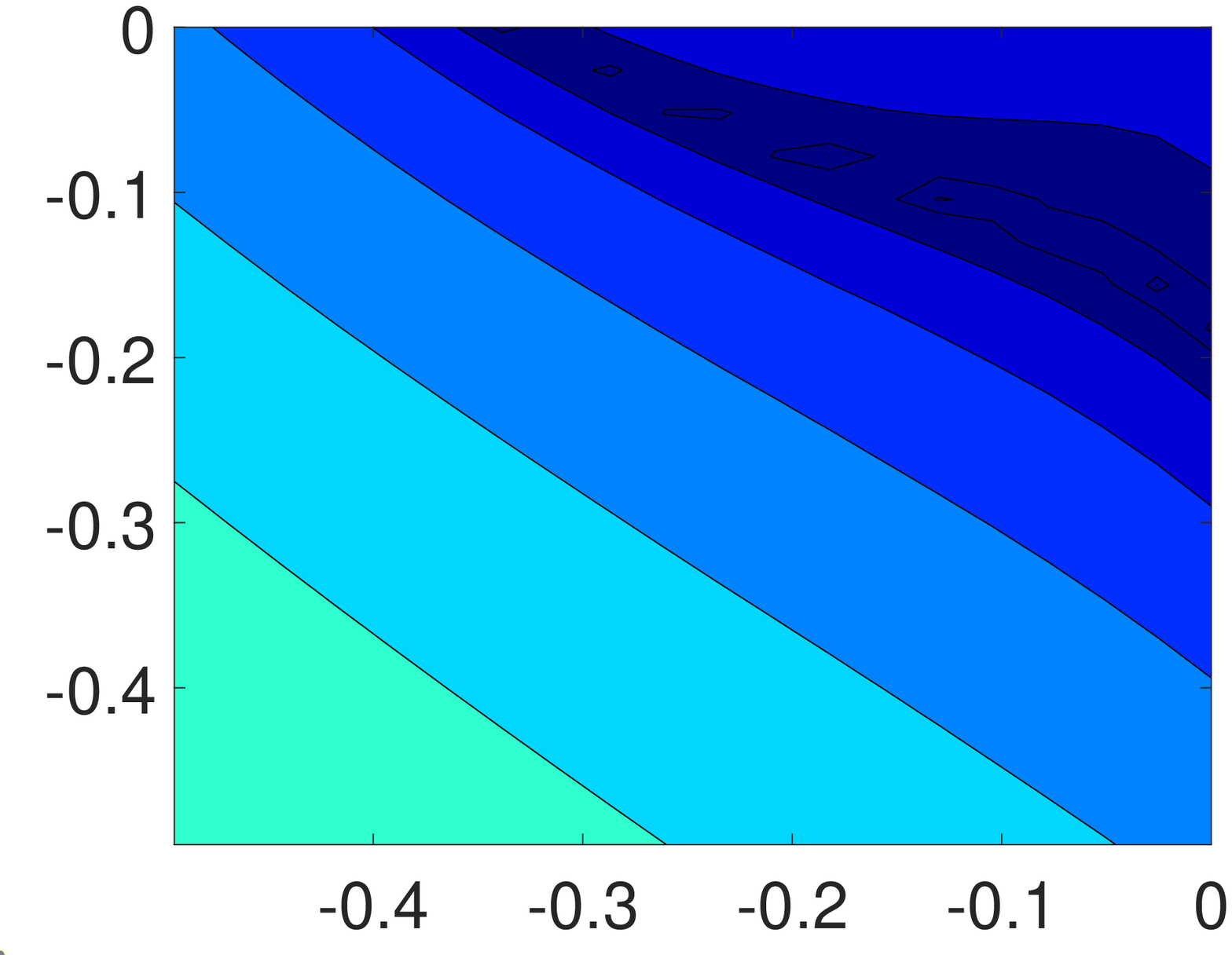}
     \caption{}
 \end{subfigure}
\caption{Model explicitly available. Training sets (grids of $20 \times 20$ equispaced distributed points). Numerical approximation accuracy (difference between the computed and analytical solution) of $T_1(x_1,x_2)$ (left column) and $T_2(x_1,x_2)$ (right column) using the various schemes. (a),(b) $6th$ order power-series expansion of $T_1(x_1,x_2)$ and $T_2(x_1,x_2)$ and the right-hand side of the model (\ref{Discretesystem}) in $[-0.495,0]\times[-0.495,0]$. (c),(d) PIML in Tensorflow trained in the entire domain $[-0.495,0]\times[-0.495,0]$. (e),(f) PIML in Tensorflow trained via the greedy-wise approach. (g),(h) PIML in Matlab trained via the greedy-wise approach.}
\label{fig:Modelavailable_TrS}
\end{figure}

\begin{figure}[htbp]
 \centering
 \begin{subfigure}[h]{0.45\textwidth}
     \centering
     \includegraphics[width=5.8cm]{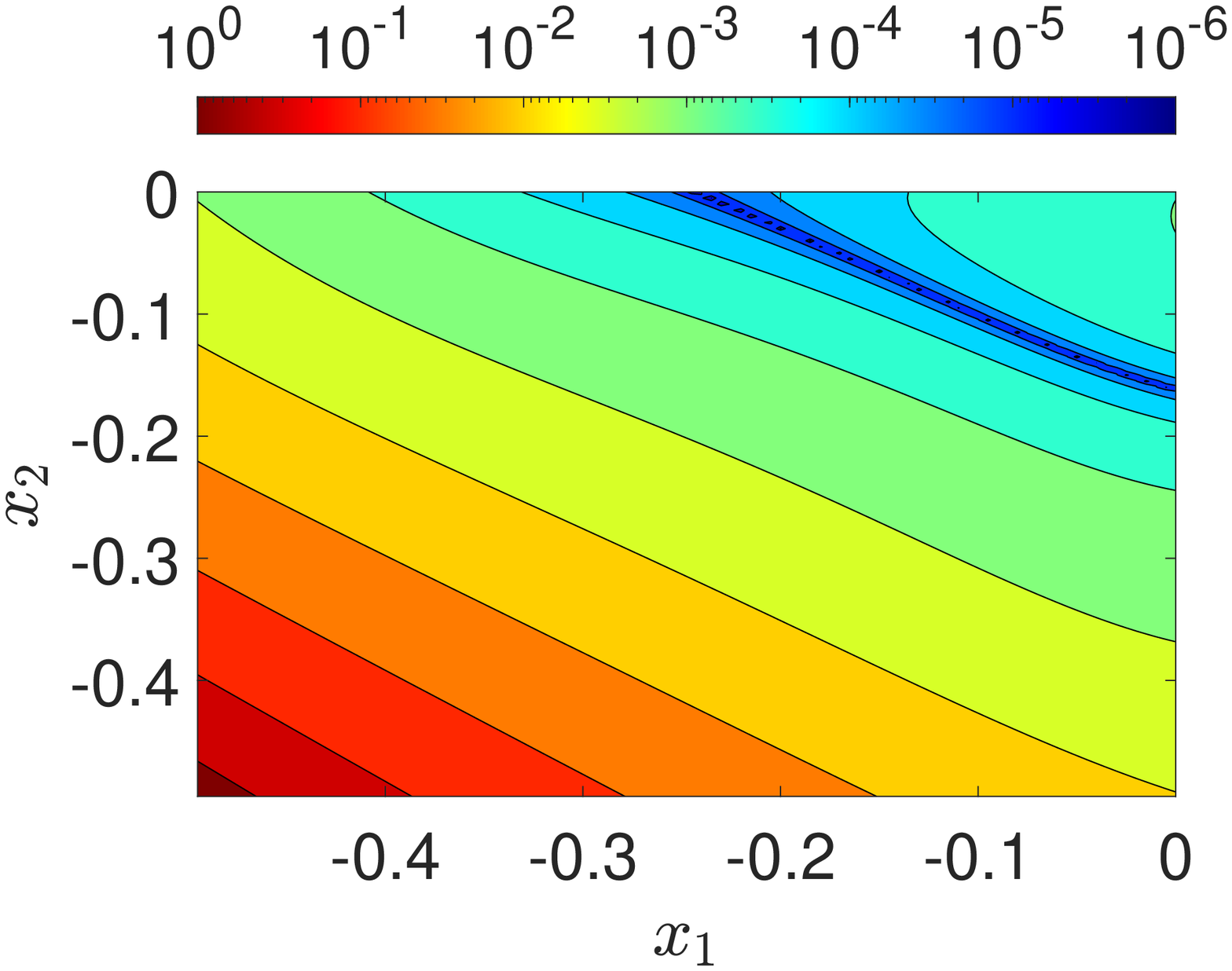}
     \caption{}
 \end{subfigure}
 \hfill
 \begin{subfigure}[h]{0.45\textwidth}
     \centering
     \includegraphics[width=5.8cm]{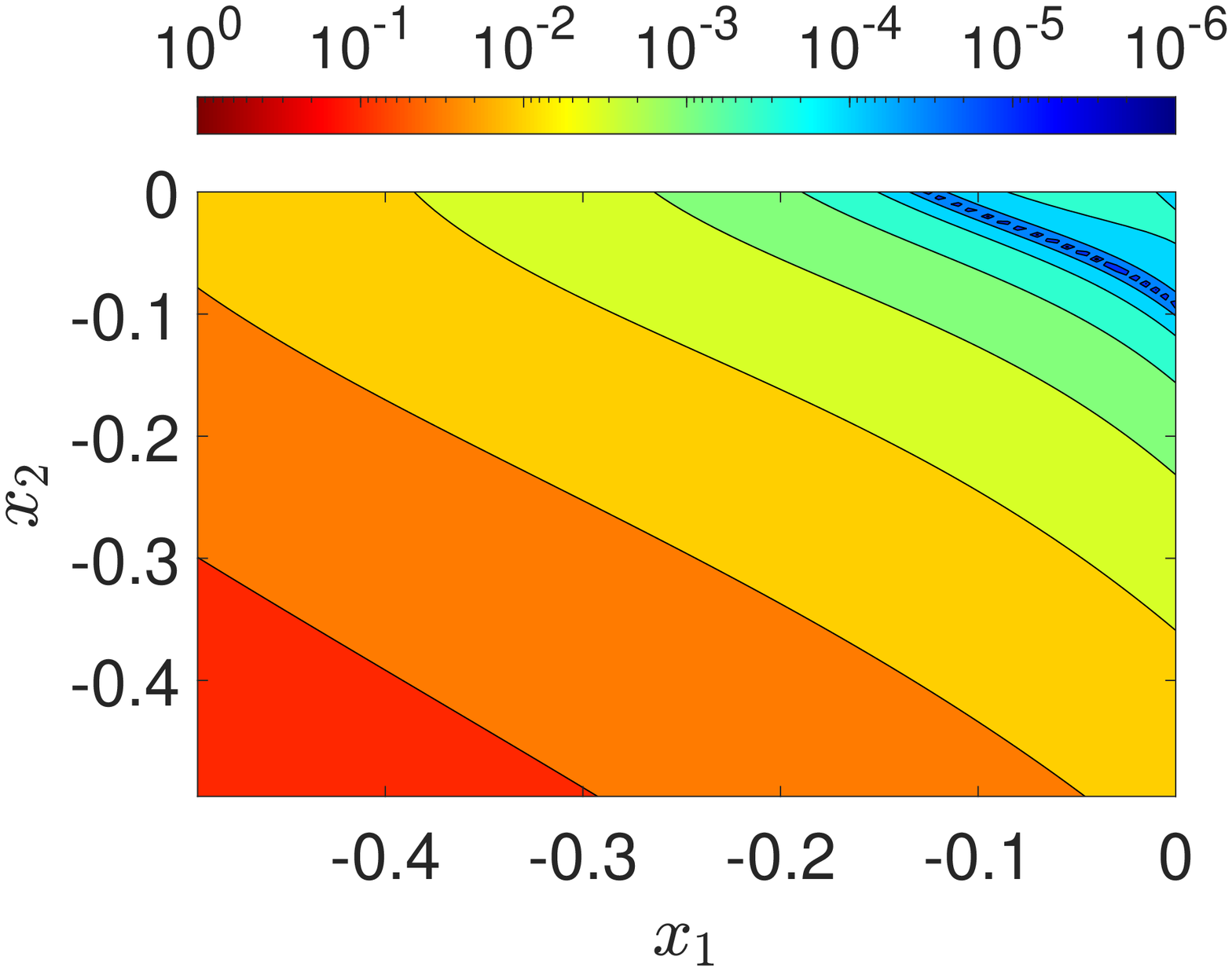}
     \caption{}
 \end{subfigure}
\centering
 \begin{subfigure}[h]{0.45\textwidth}
     \centering
     \includegraphics[width=5.7cm]{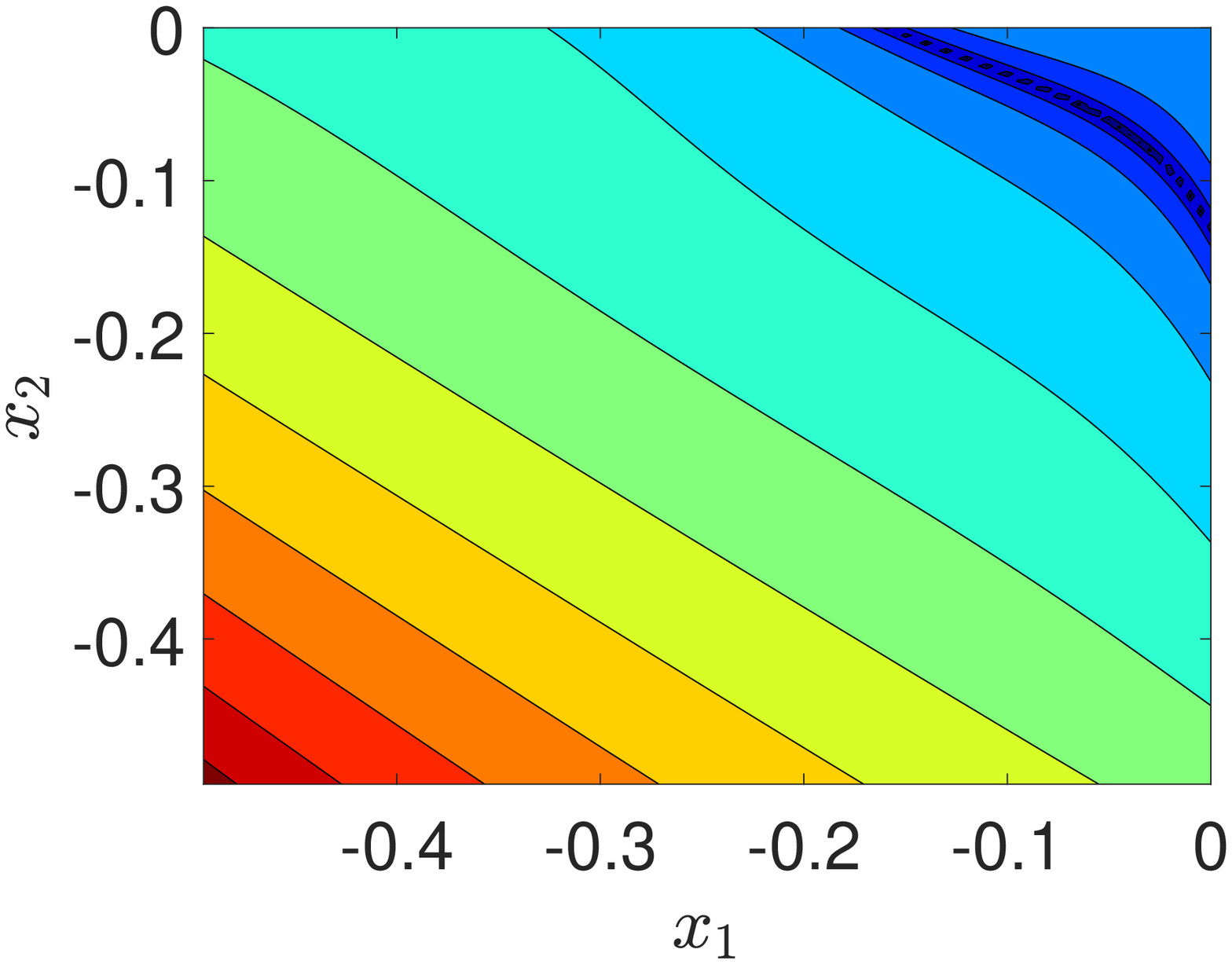}
     \caption{}
 \end{subfigure}
 \hfill
 \begin{subfigure}[h]{0.45\textwidth}
     \centering
     \includegraphics[width=5.7cm]{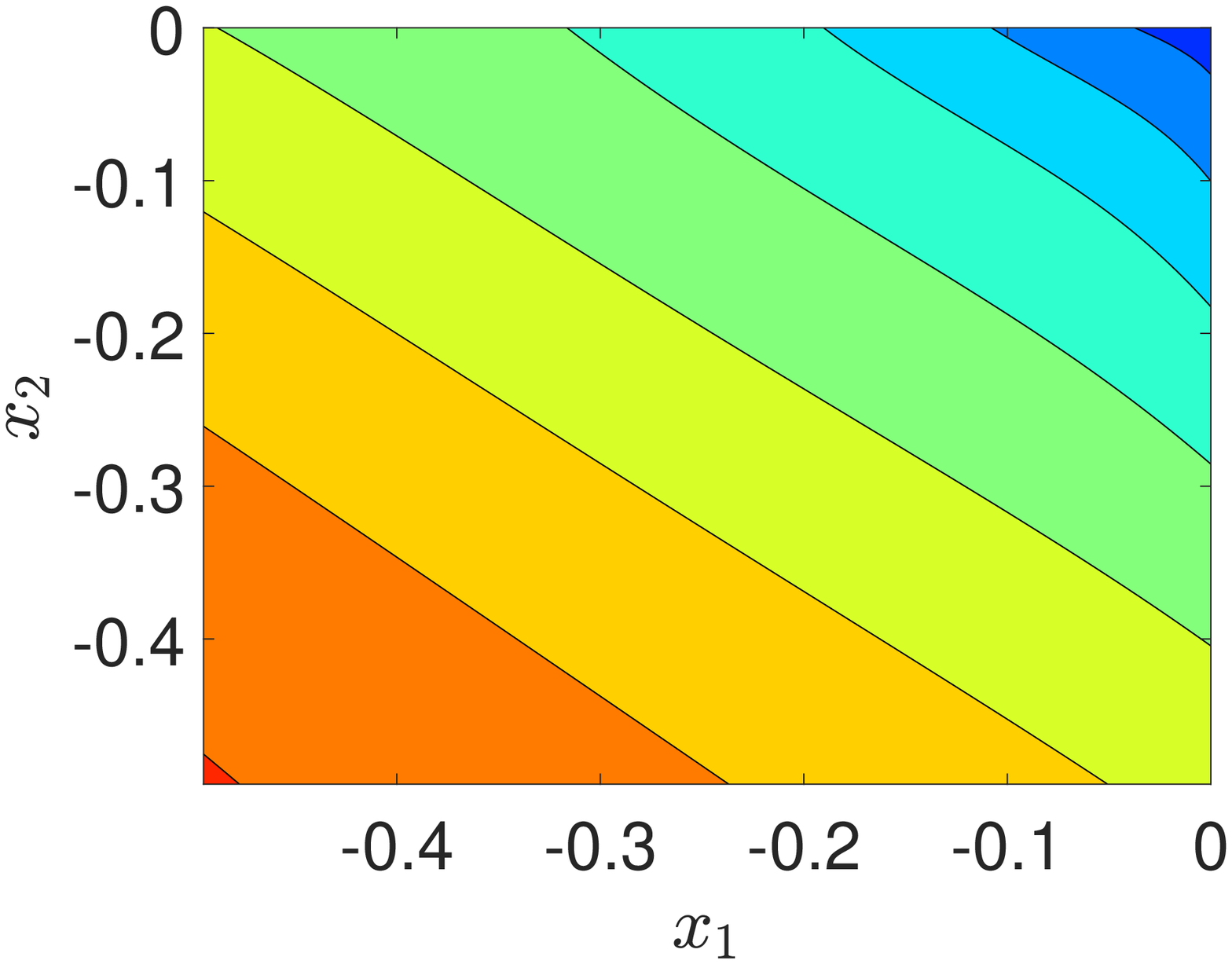}
     \caption{}
 \end{subfigure}
 \centering
 \begin{subfigure}[h]{0.45\textwidth}
     \centering
     \includegraphics[width=5.7cm]{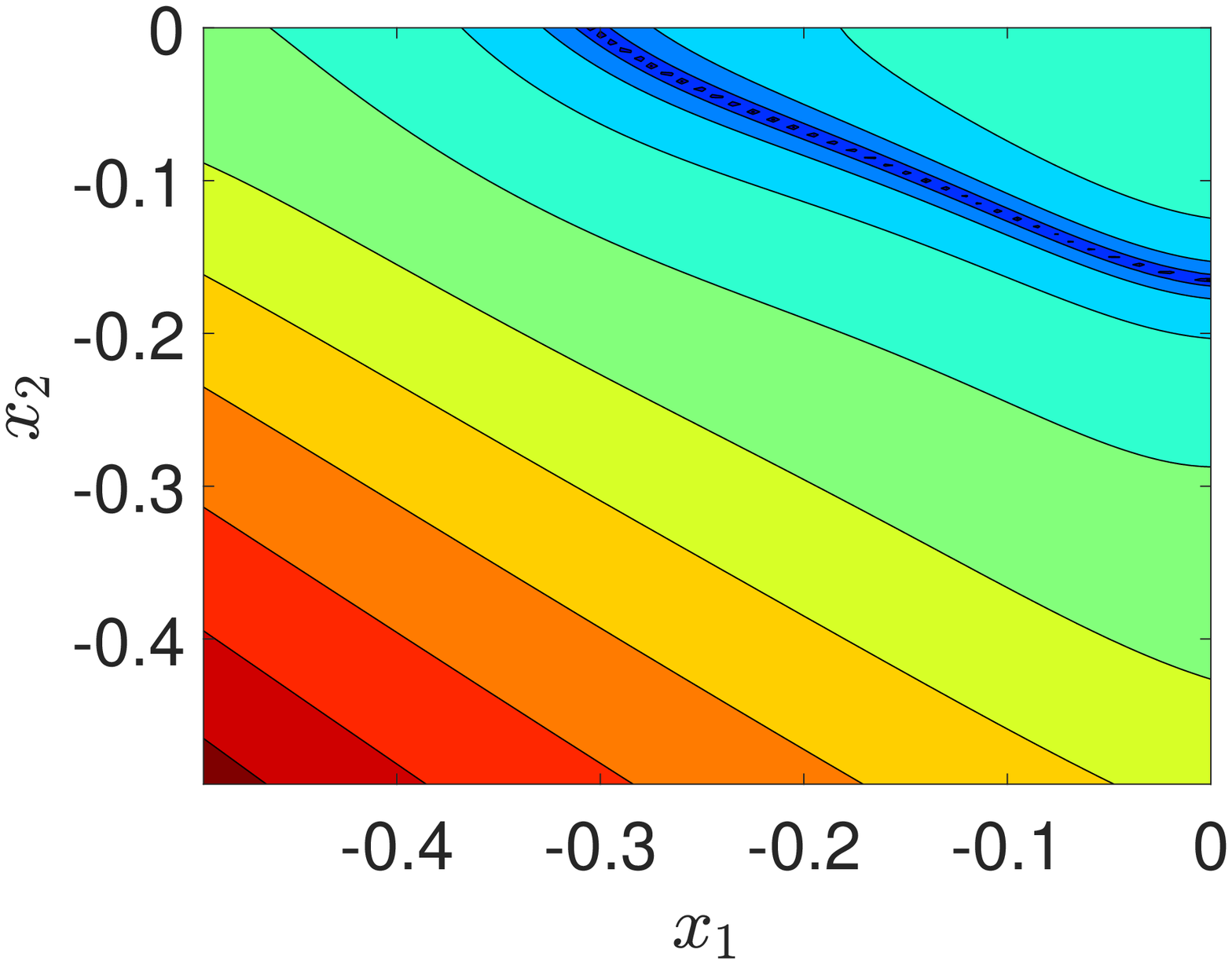}
     \caption{}
 \end{subfigure}
 \hfill
 \begin{subfigure}[h]{0.45\textwidth}
     \centering
     \includegraphics[width=5.7cm]{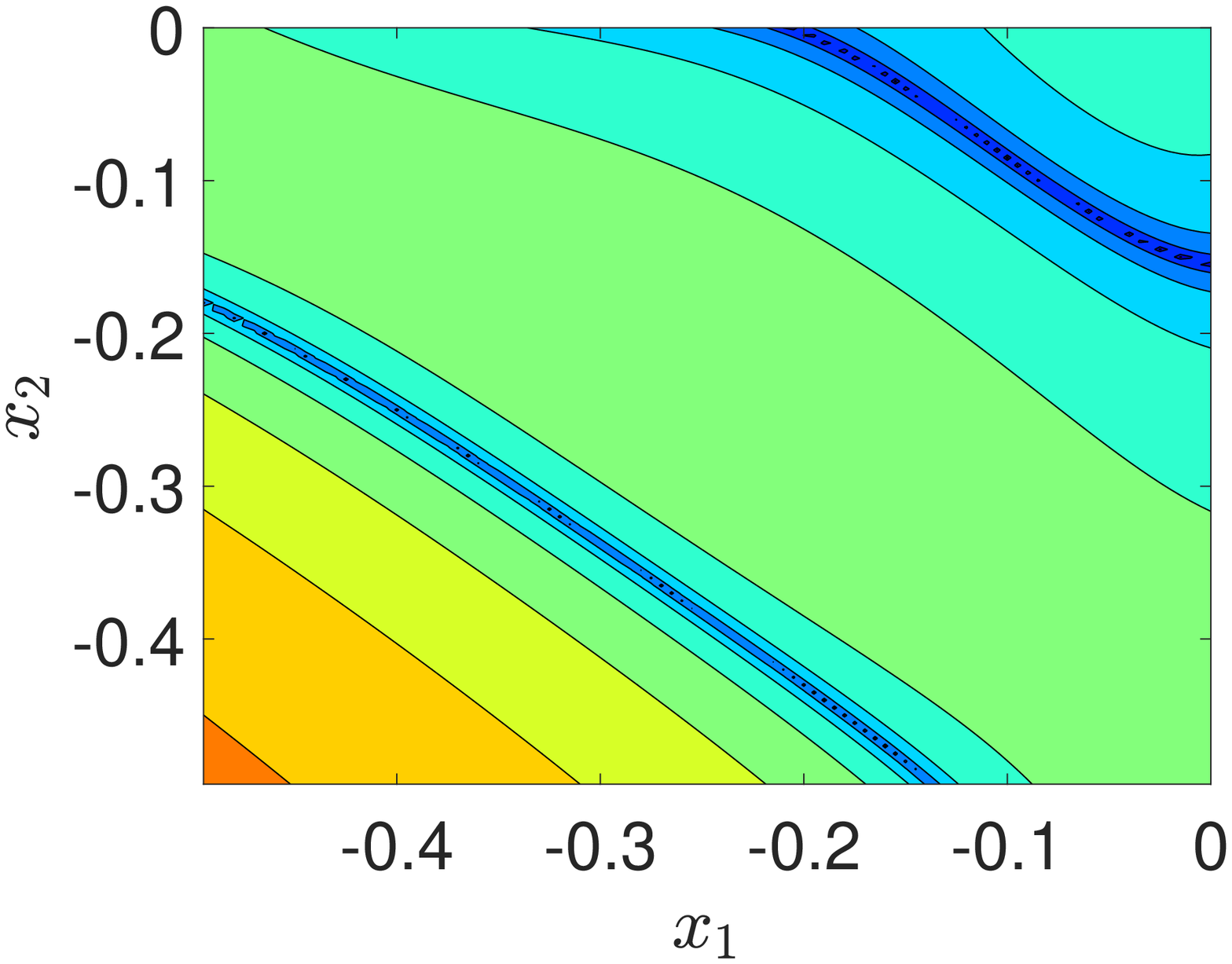}
     \caption{}
 \end{subfigure}
\caption{Model explicitly available. Training set (grid of $100 \times 100$ equispaced distributed points in $[-0.495,0]\times[-0.495,0]$). Numerical approximation accuracy (difference between the computed and analytical solution) of $T_1(x_1,x_2)$ (left column) and $T_2(x_1,x_2)$ (right column) using the Keras API in TensorFlow; training was performed in the entire domain. (a),(b) PIML, two hidden layers, five neurons in each layer. (c),(d) PIML, two hidden layers, ten neurons in each layer. (e),(f) PIML, two hidden layers, fifteen neurons in each layer.}
\label{fig:100x100_Diffneurons}
\end{figure}

\begin{figure}[htbp]
 \centering
 \begin{subfigure}[h]{0.45\textwidth}
     \centering
     \includegraphics[width=5.8cm]{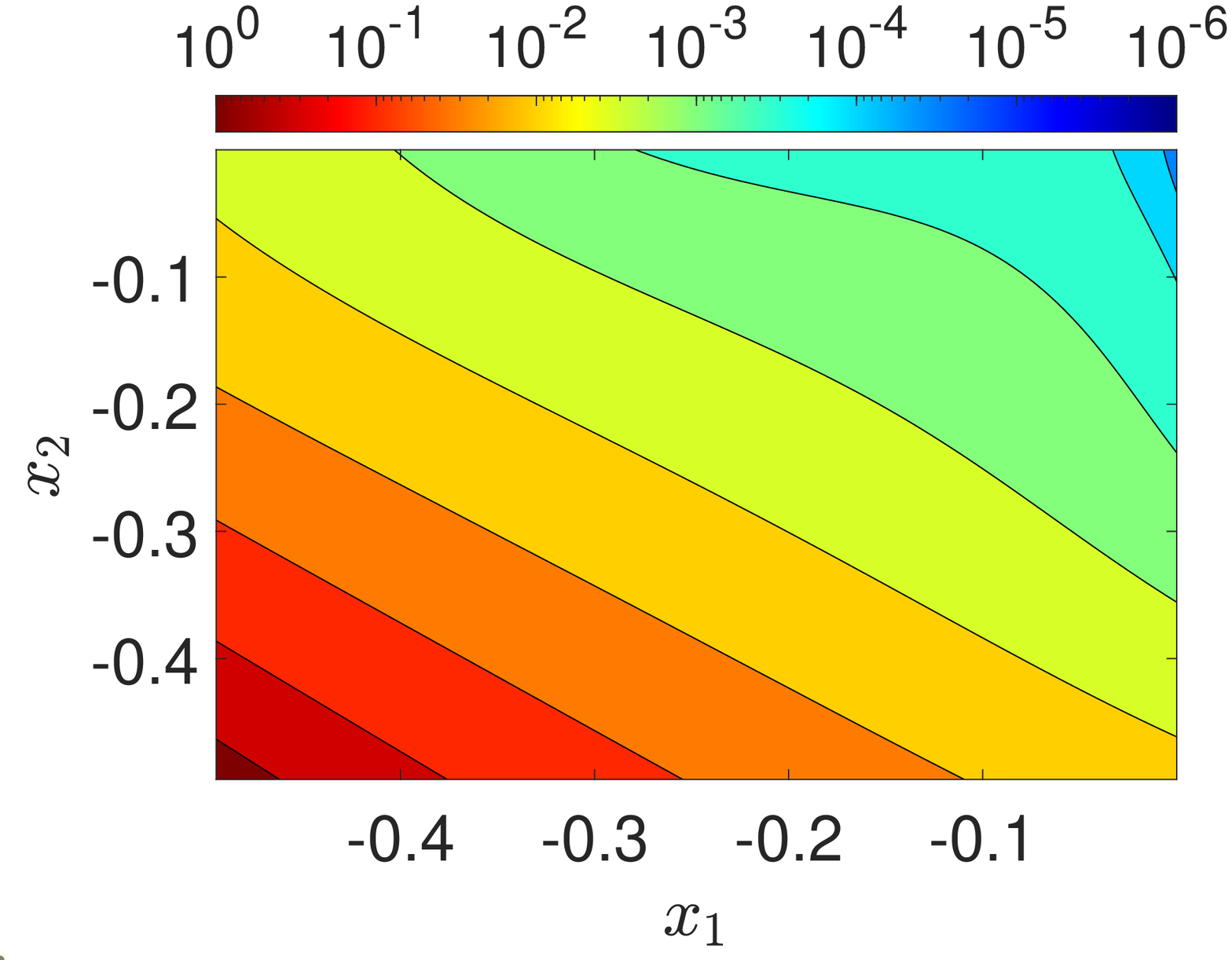}
     \caption{}
 \end{subfigure}
 \hfill
 \begin{subfigure}[h]{0.45\textwidth}
     \centering
     \includegraphics[width=5.8cm]{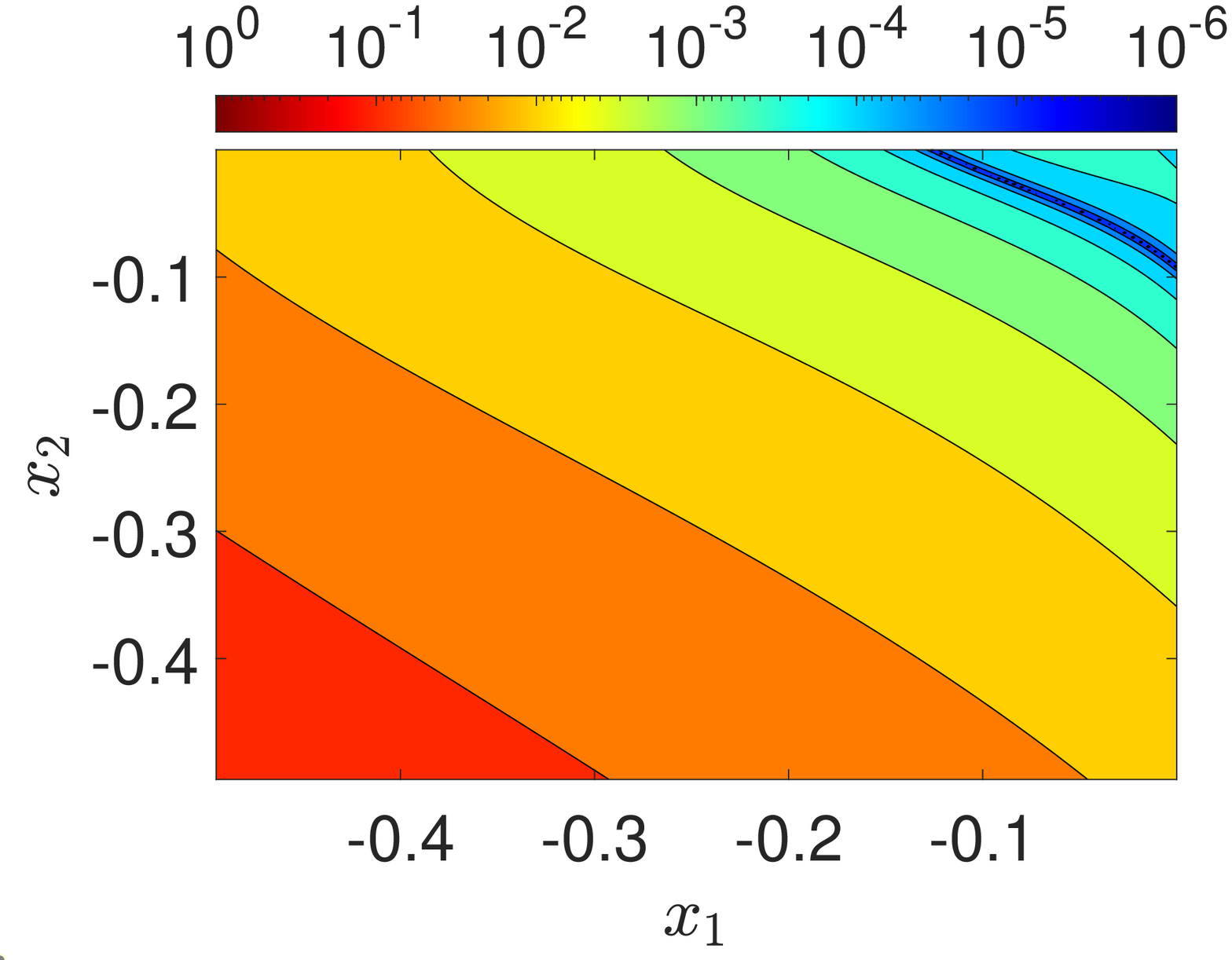}
     \caption{}
 \end{subfigure}
\centering
 \begin{subfigure}[h]{0.45\textwidth}
     \centering
     \includegraphics[width=5.7cm]{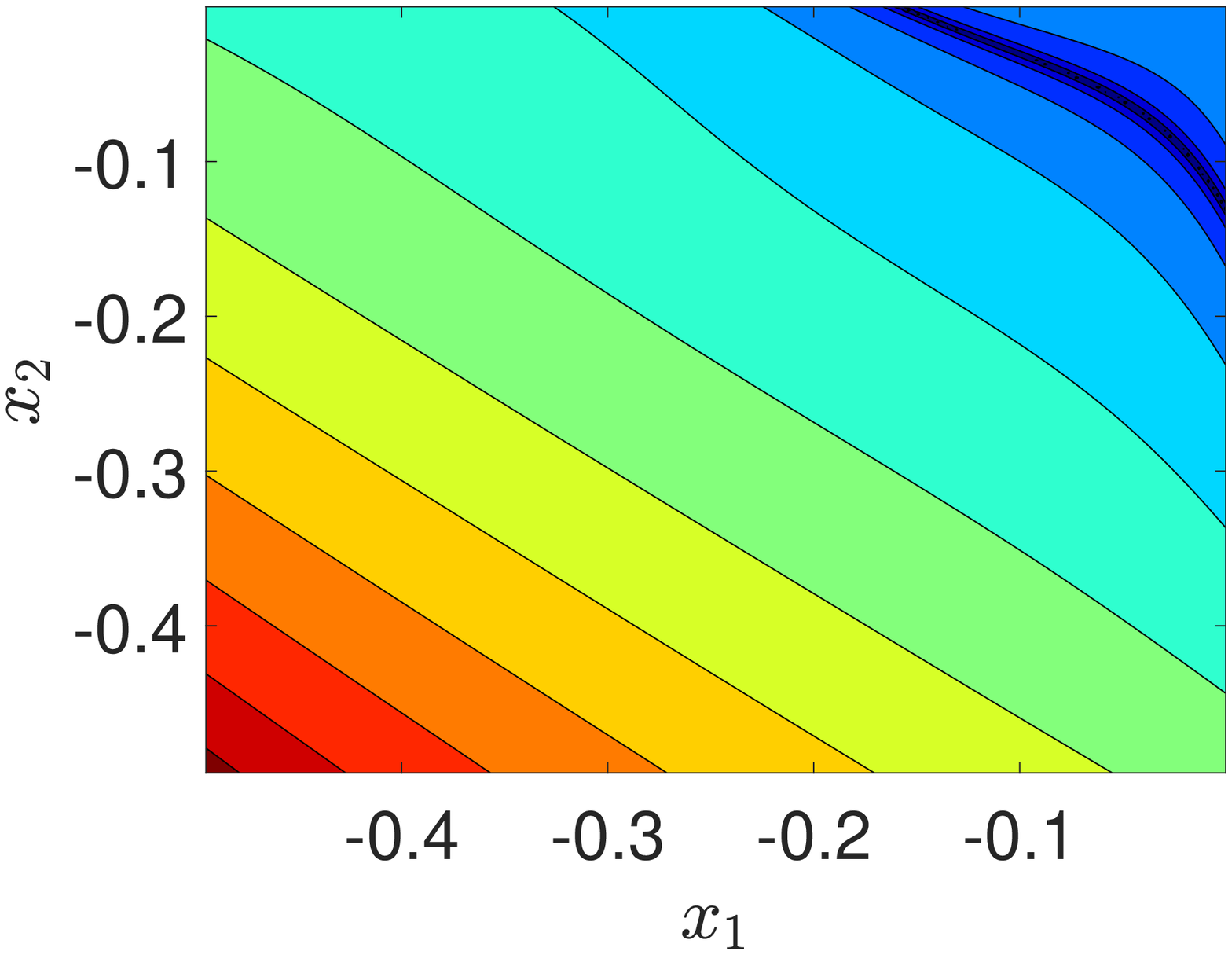}
     \caption{}
 \end{subfigure}
 \hfill
 \begin{subfigure}[h]{0.45\textwidth}
     \centering
     \includegraphics[width=5.7cm]{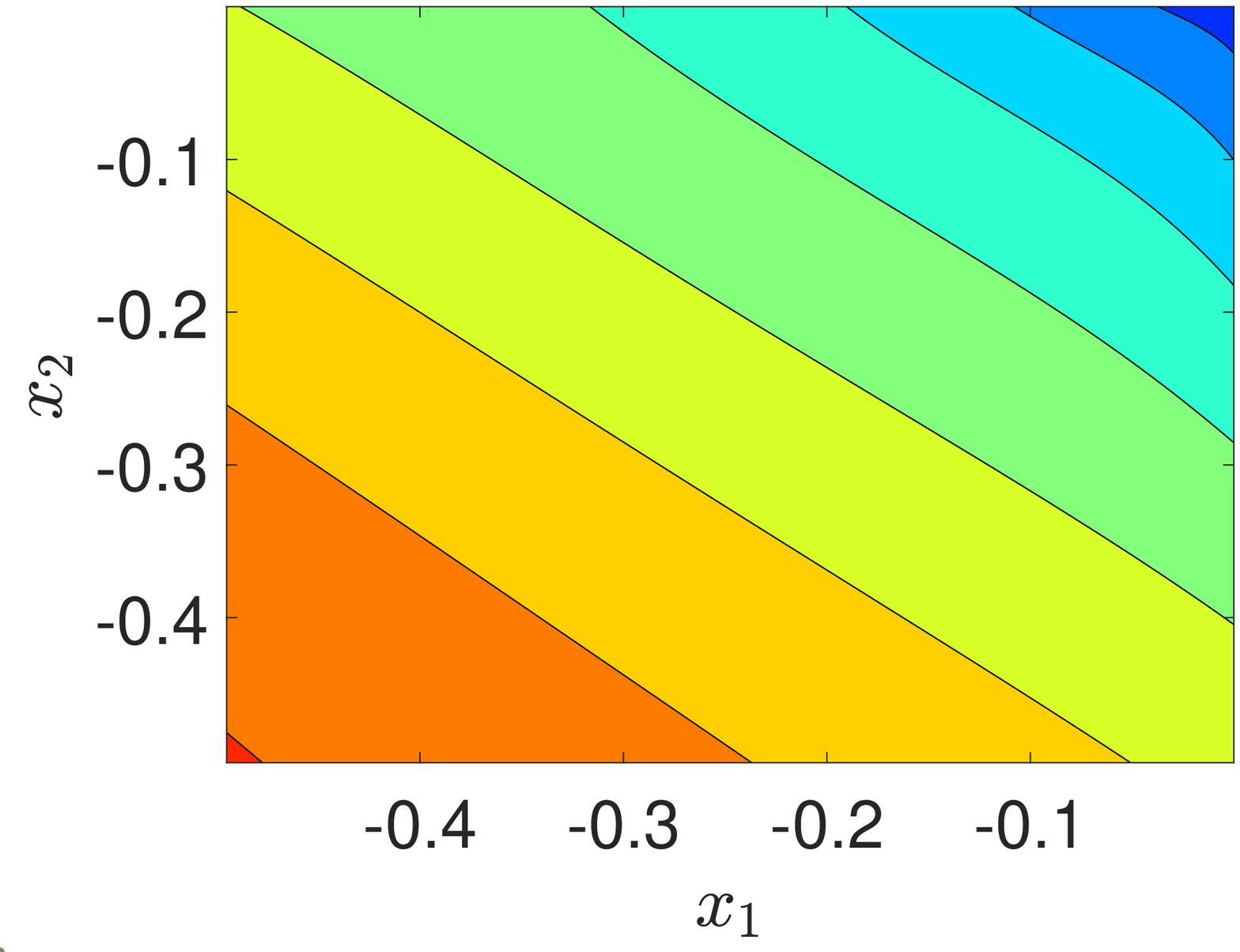}
     \caption{}
 \end{subfigure}
 \centering
 \begin{subfigure}[h]{0.45\textwidth}
     \centering
     \includegraphics[width=5.7cm]{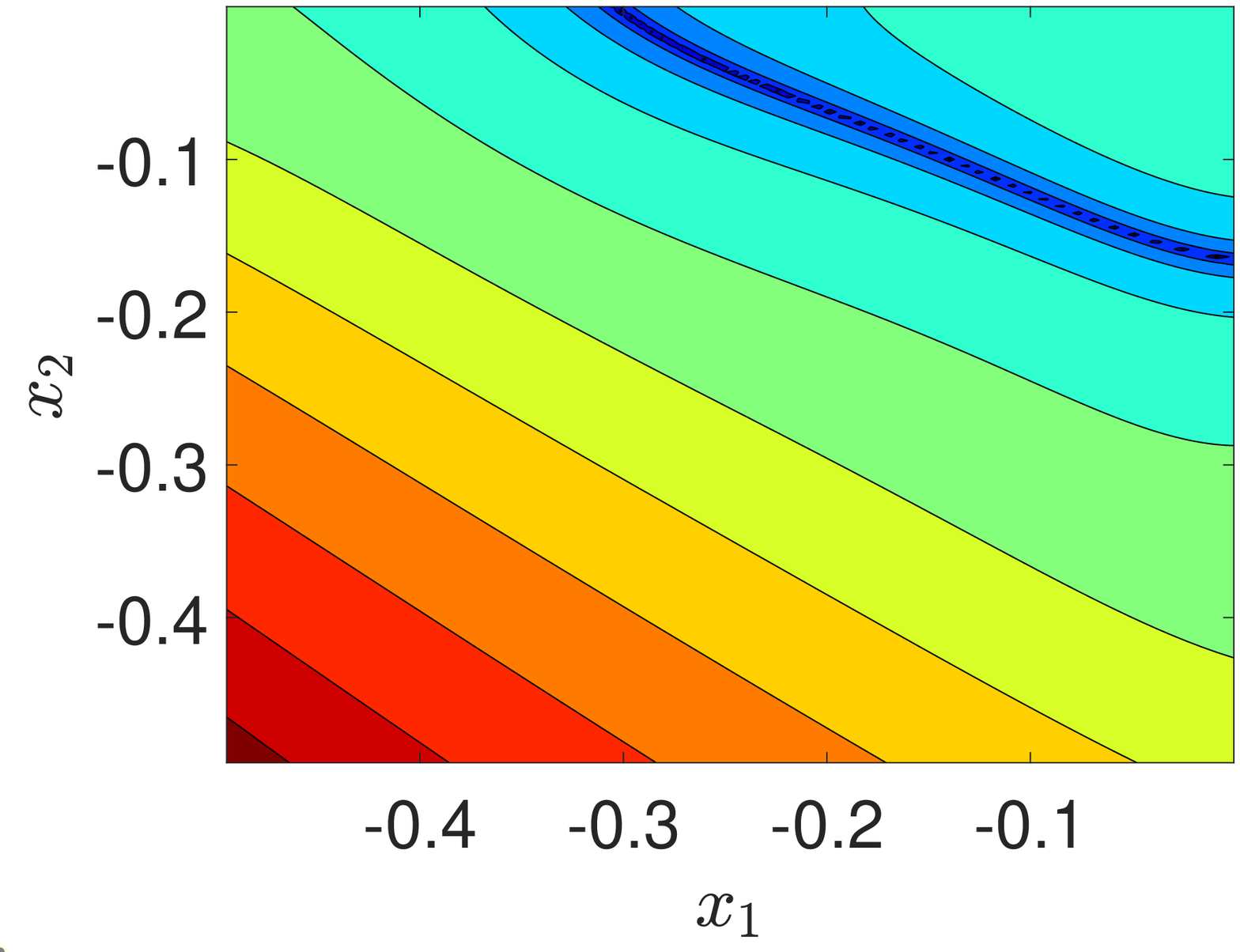}
     \caption{}
 \end{subfigure}
 \hfill
 \begin{subfigure}[h]{0.45\textwidth}
     \centering
     \includegraphics[width=5.7cm]{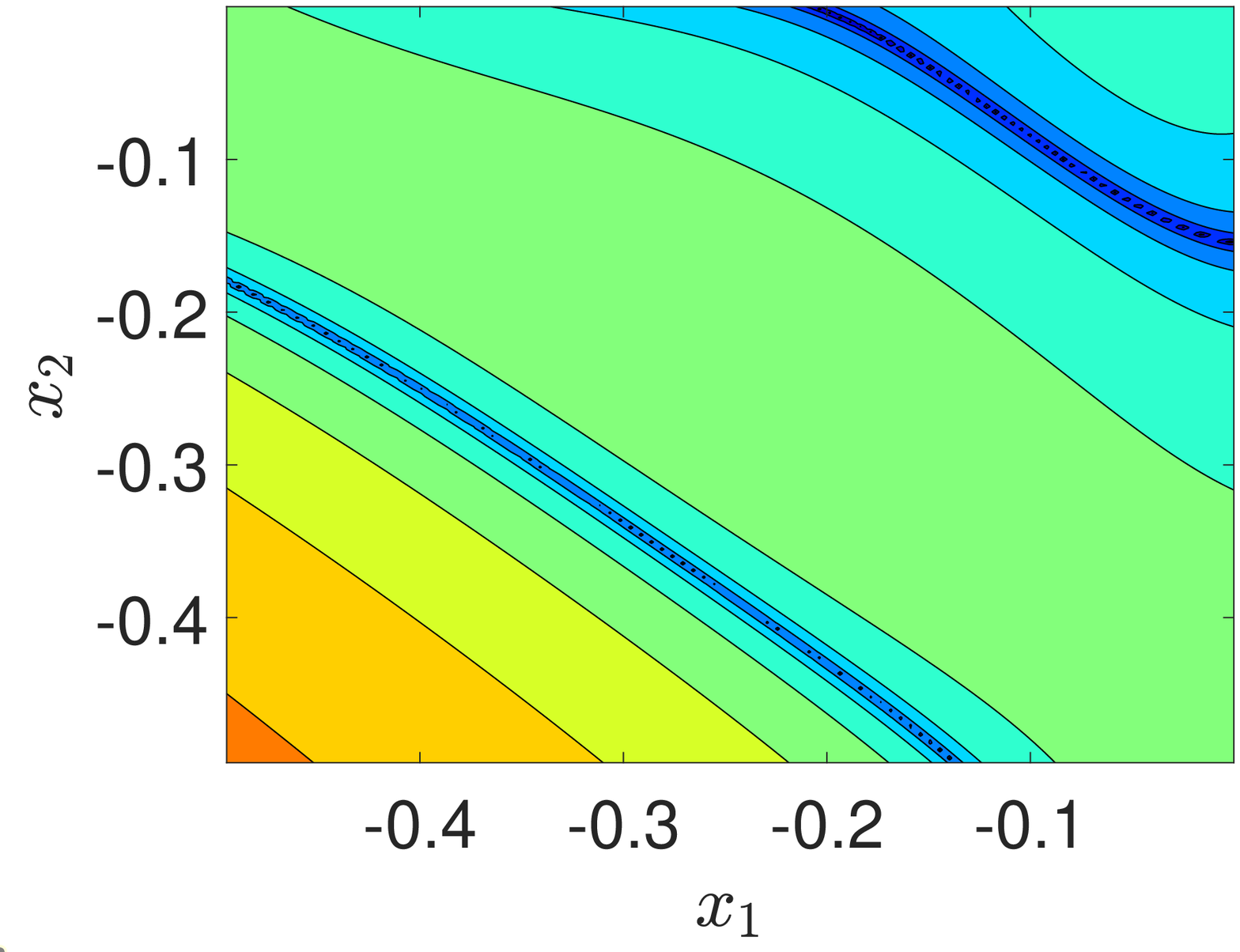}
     \caption{}
 \end{subfigure}
\caption{Model explicitly available. Test set (grid of $150 \times 150$ Chebyshev-distributed points in $[-0.495,0]\times[-0.495,0]$). Numerical approximation accuracy (difference between the computed and analytical solution) of $T_1(x_1,x_2)$ (left column) and $T_2(x_1,x_2)$ (right column) using the Keras API in TensorFlow; training was performed in the entire domain. (a),(b) PIML, two hidden layers, five neurons in each layer. (c),(d) PIML, two hidden layers, ten neurons in each layer. (e),(f) PIML, two hidden layers, fifteen neurons in each layer.}
\label{fig:100x100_Diffneurons_test}
\end{figure}

\begin{table}[ht!]
    \centering
    \caption{Model explicitly available. Training sets (grids of $20 \times 20$ equispaced distributed points). Error norms ($L_1$, $L_2$ and $L_{\infty}$) between the analytical and computed solution of $T_1(x_1,x_2)$ and $T_2(x_1,x_2)$ using the various schemes trained both greedy-wised and in the entire domain $[-0.495,0]\times[-0.495,0]$.}
    \begin{tabular}{c| c c | c c }
    \toprule
    Error norm & power-series&PIML(TF)&PIML(Matlab)&PIML(TF)\\
    & $6th$ order&Entire domain&Greedy&Greedy\\
    \midrule
    $\lVert \cdot \rVert_1$ &6.76E$+$01
        &6.28E$+$00 & 2.03E$-$03 & 3.65E$-$02   \\
     $\lVert \cdot \rVert_{2}$ & 3.62E$+$00 & 3.73E$+$00 & 1.12E$-$03 & 3.26E$-$02\\
    $\lVert \cdot \rVert_{\infty}$    & 1.21E$+$00 & 2.81E$+$00 & 1.05E$-$03 & 3.10E$-$02\\
    \midrule
       $\lVert \cdot \rVert_1$ & 0
 & 1.40E$+$00 & 6.33E$-$03 & 6.61E$-$02  \\
  $\lVert \cdot \rVert_{2}$
  & 0 & 1.00E$+$00 & 1.40E$-$03 & 3.68E$-$02\\
    $\lVert \cdot \rVert_{\infty}$ & 0 &5.94E$-$01 & 6.73E$-$04 & 1.00E$-$02\\
    \bottomrule
    \end{tabular}
    \label{tab:NormsS1}
\end{table}
\begin{table}[ht!]
    \centering
    \caption{Model explicitly available. Test sets (grids of $50 \times 50$ Chebyshev-distributed points). Error norms ($L_1$, $L_2$ and $L_{\infty}$) between the analytical and computed solution of $T_1(x_1,x_2)$ and $T_2(x_1,x_2)$ using the various schemes trained both greedy-wised and in the entire domain $[-0.495,0]\times[-0.495,0]$.}
    \begin{tabular}{c| c c | c c }
    \toprule
    Error norm & power-series&PIML(TF)&PIML(Matlab)&PIML(TF)\\
    & $6th$ order&Entire domain& Greedy & Greedy\\
    \midrule
    $\lVert \cdot \rVert_1$ & 6.89E$+$01
        & 2.17E$+$01 & 3.40E$-$02 & 1.11E$-$01    \\
    $\lVert \cdot \rVert_{2}$ & 4.55E$+$00 & 1.44E$+$01 & 2.63E$-$03 & 7.72E$-$02\\
    $\lVert \cdot \rVert_{\infty}$ & 2.88E$+$00 & 1.00E$+$01 &1.41E$-$03 & 1.11E$-$02\\
    \midrule
       $\lVert \cdot \rVert_1$ &0
 & 2.87E$+$00 & 1.45E$-$02 & 2.22E$-$01  \\
  $\lVert \cdot \rVert_{2}$
  & 0 & 1.97E$+$00 & 1.55E$-$03 & 1.45E$-$01\\
     $\lVert \cdot \rVert_{\infty}$ & 0 & 1.23E$+$00 & 1.04E$-$03 & 1.28E$-$01\\
    \bottomrule
    \end{tabular}
    \label{tab:NormsS2}
\end{table}
Taking denser grids and more neurons in each hidden layer, did not change qualitatively the numerical approximation accuracy. Indicatively, in Figure (\ref{fig:100x100_Diffneurons}), we depict the numerical approximation accuracy obtained with the PIML implemented in TensorFlow (TD) trained with BFGS in the entire domain $[-0.495,0] \times [-0.495,0]$ using $100 \times 100$ equispaced points and different number of neurons in each hidden layer. In particular, Figures (\ref{fig:100x100_Diffneurons})(a),(b) show, the numerical approximation accuracy in the training set using two hidden layers with five neurons in each layer, Figures (\ref{fig:100x100_Diffneurons})(c),(d) with ten neurons, and figures (\ref{fig:100x100_Diffneurons})(e),(f) with fifteen neurons in each layer. For the completeness of the presentation in Figure (\ref{fig:100x100_Diffneurons_test}), we provide also the corresponding numerical approximation accuracy plots for the test set.

\subsection{The black-box simulator case}
 As opposed to the ``explicitly known model'' PIML scheme, where we used analytical derivatives, in the black-box simulator scheme, the derivatives were estimated using central finite differences with a perturbation step of $eps=10^{-4}$. Here, for our illustrations, we have implemented the PIML only in Matlab in a ``fully numerical way''.\par
\begin{figure}[htbp]
 \centering
 \begin{subfigure}[h]{0.45\textwidth}
     \centering    \includegraphics[width=5.5cm]{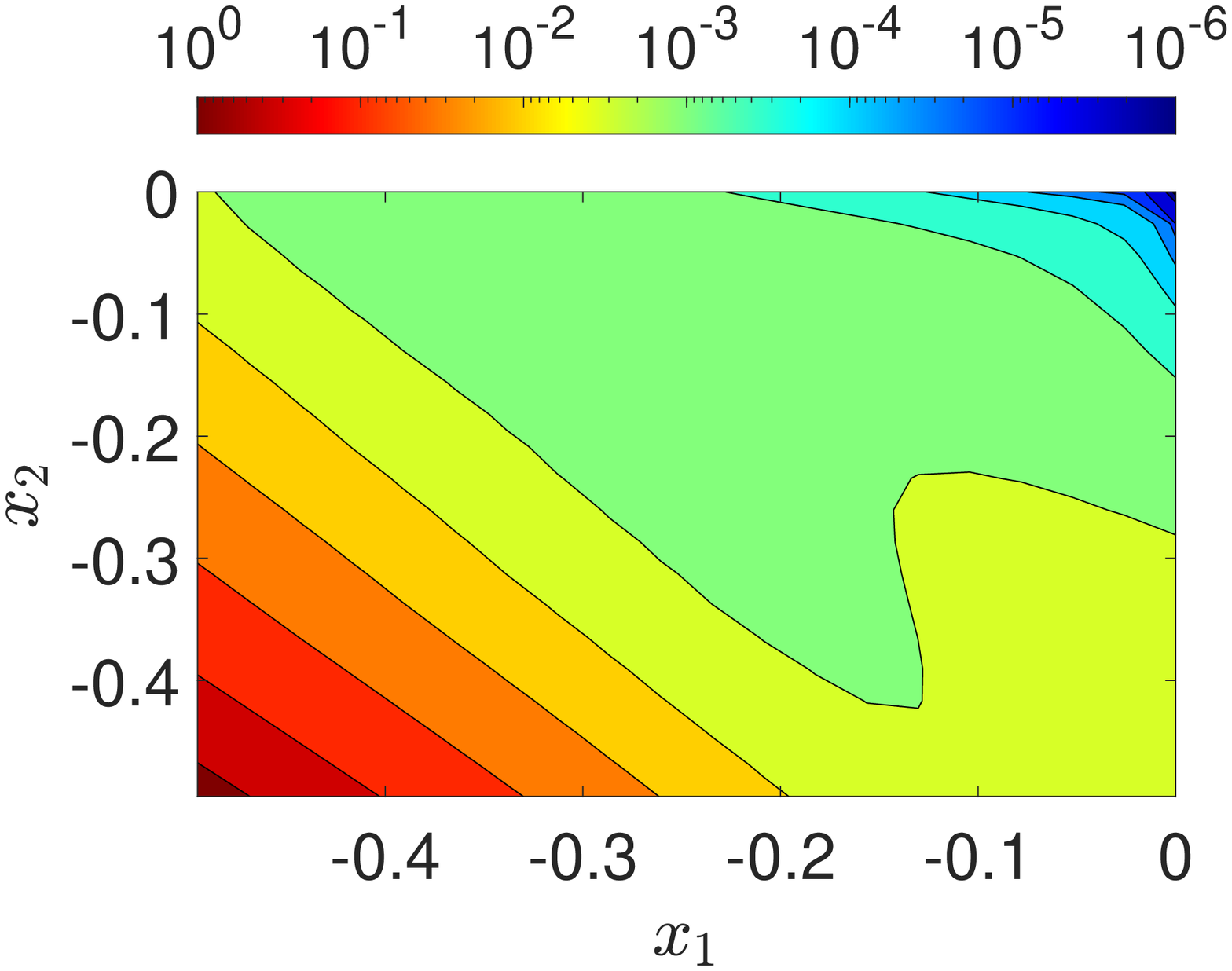}
     \caption{}
 \end{subfigure}
 \hfill
 \begin{subfigure}[h]{0.45\textwidth}
     \centering
     \includegraphics[width=5.4cm]{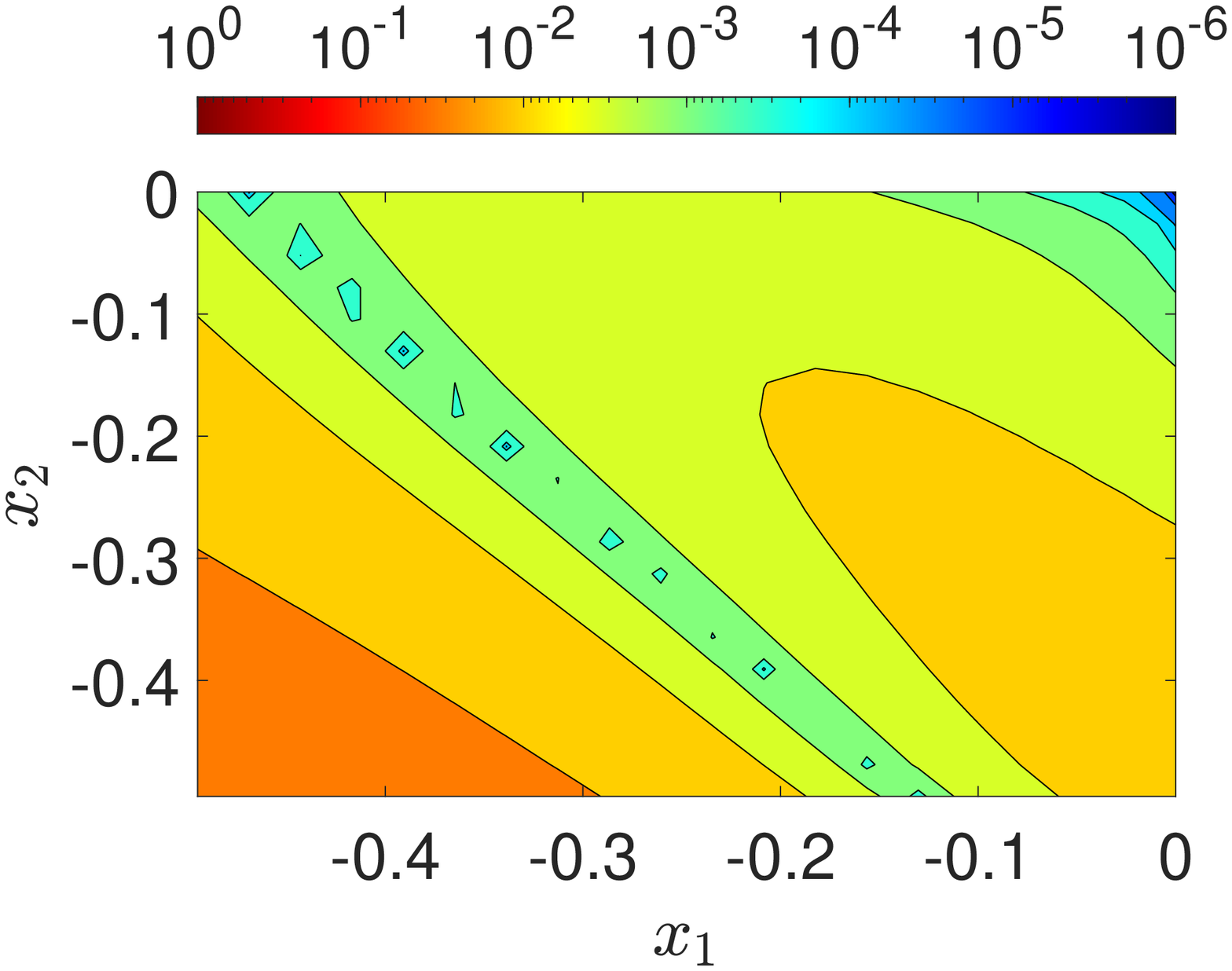}
     \caption{}
 \end{subfigure}
  \centering
 \begin{subfigure}[h]{0.45\textwidth}
     \centering \includegraphics[width=5.4cm]{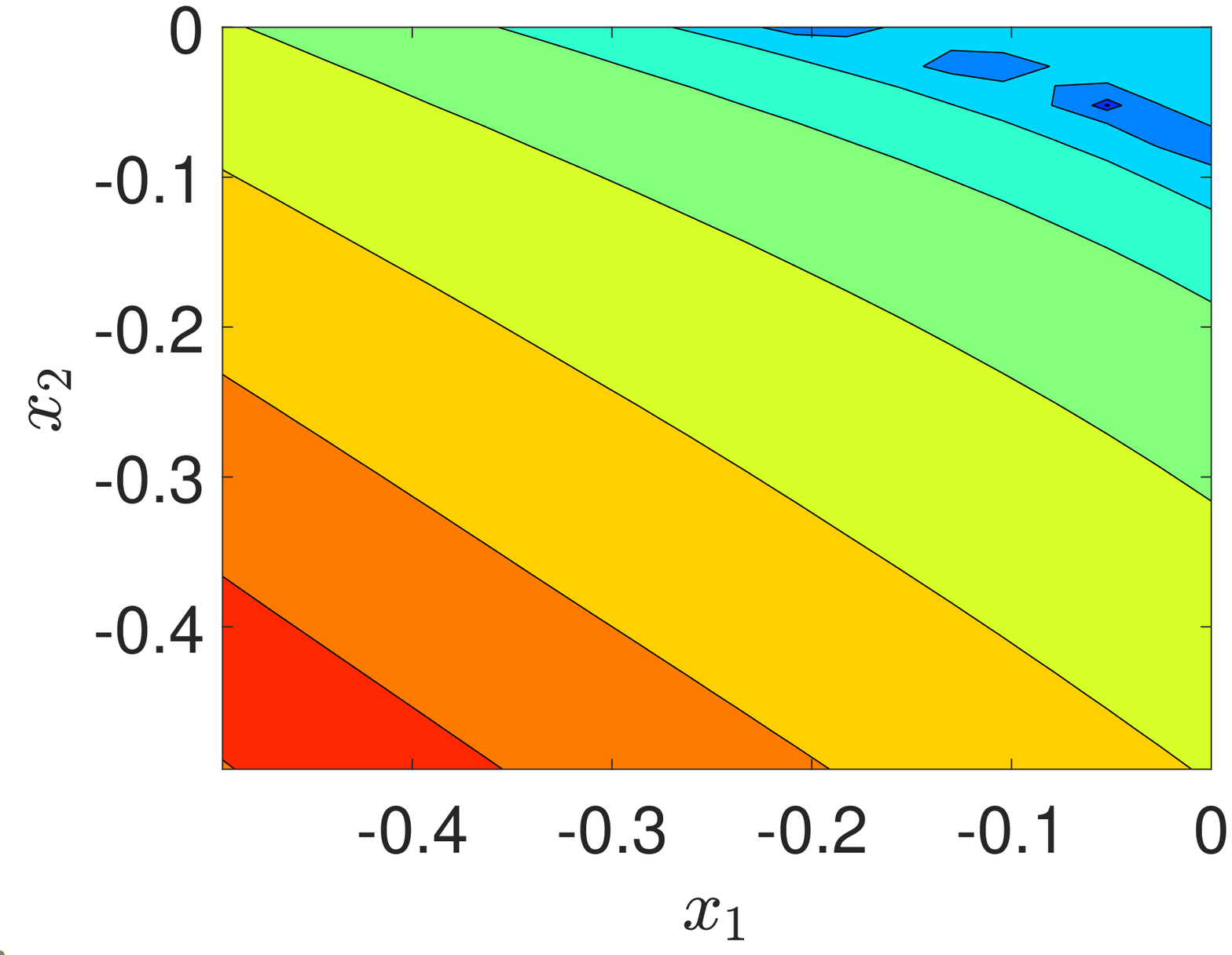}
     \caption{}
 \end{subfigure}
 \hfill
 \begin{subfigure}[h]{0.45\textwidth}
     \centering \includegraphics[width=5.5cm]{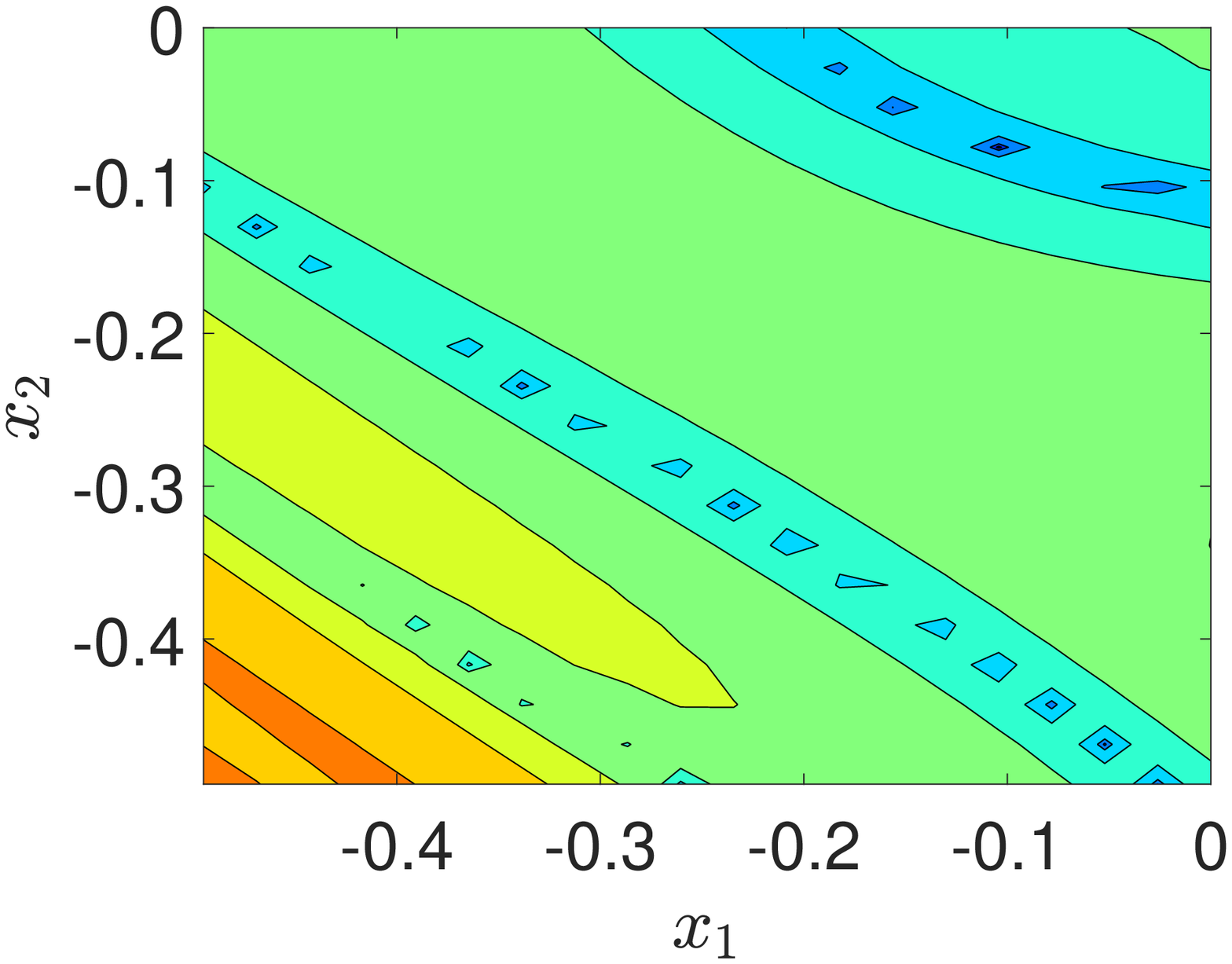}
     \caption{}
 \end{subfigure}
 \begin{subfigure}[h]{0.45\textwidth}
     \centering
     \includegraphics[width=5.4cm]{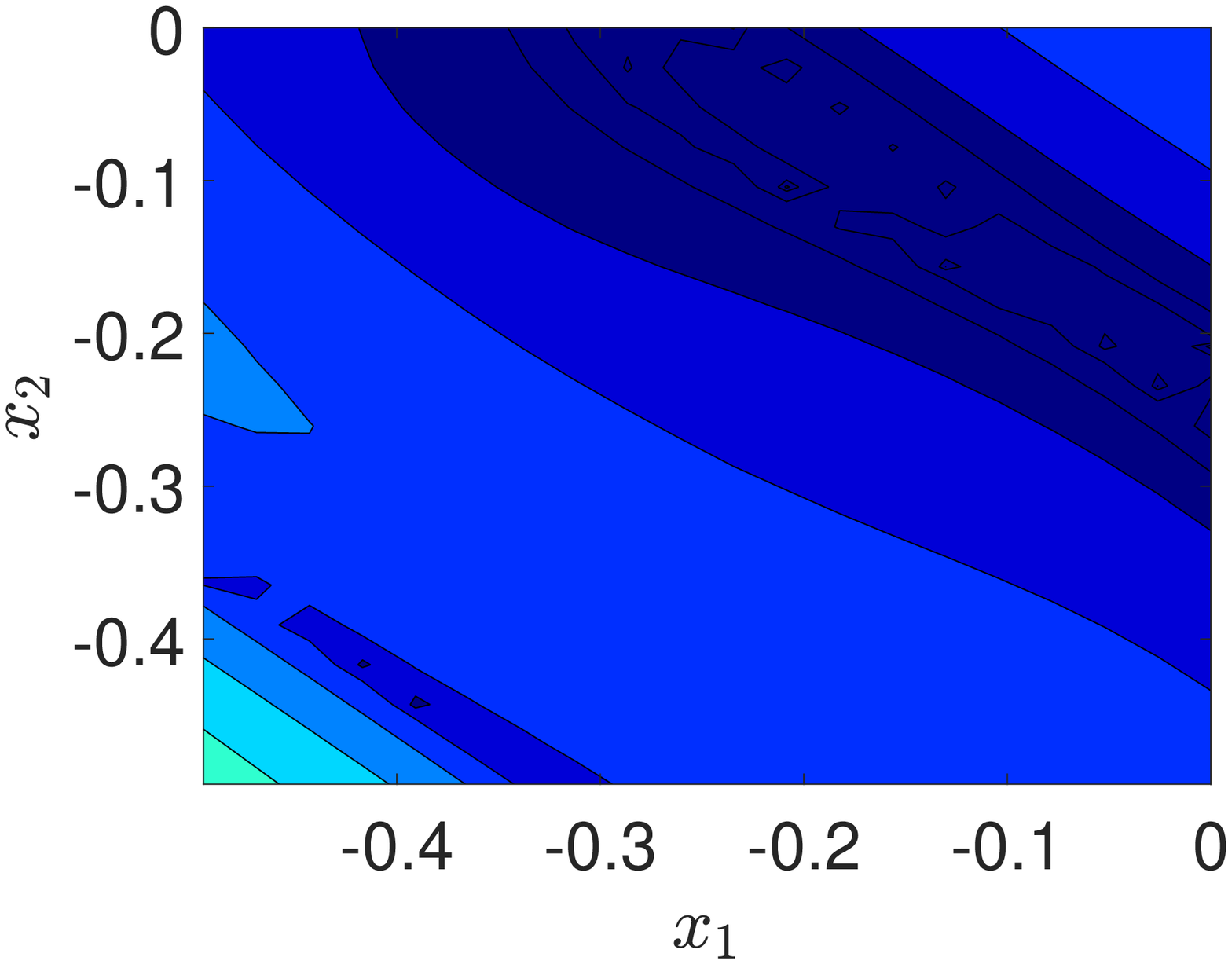}
     \caption{}
 \end{subfigure}
 \hfill
 \begin{subfigure}[h]{0.45\textwidth}
     \centering  \includegraphics[width=5.4cm]{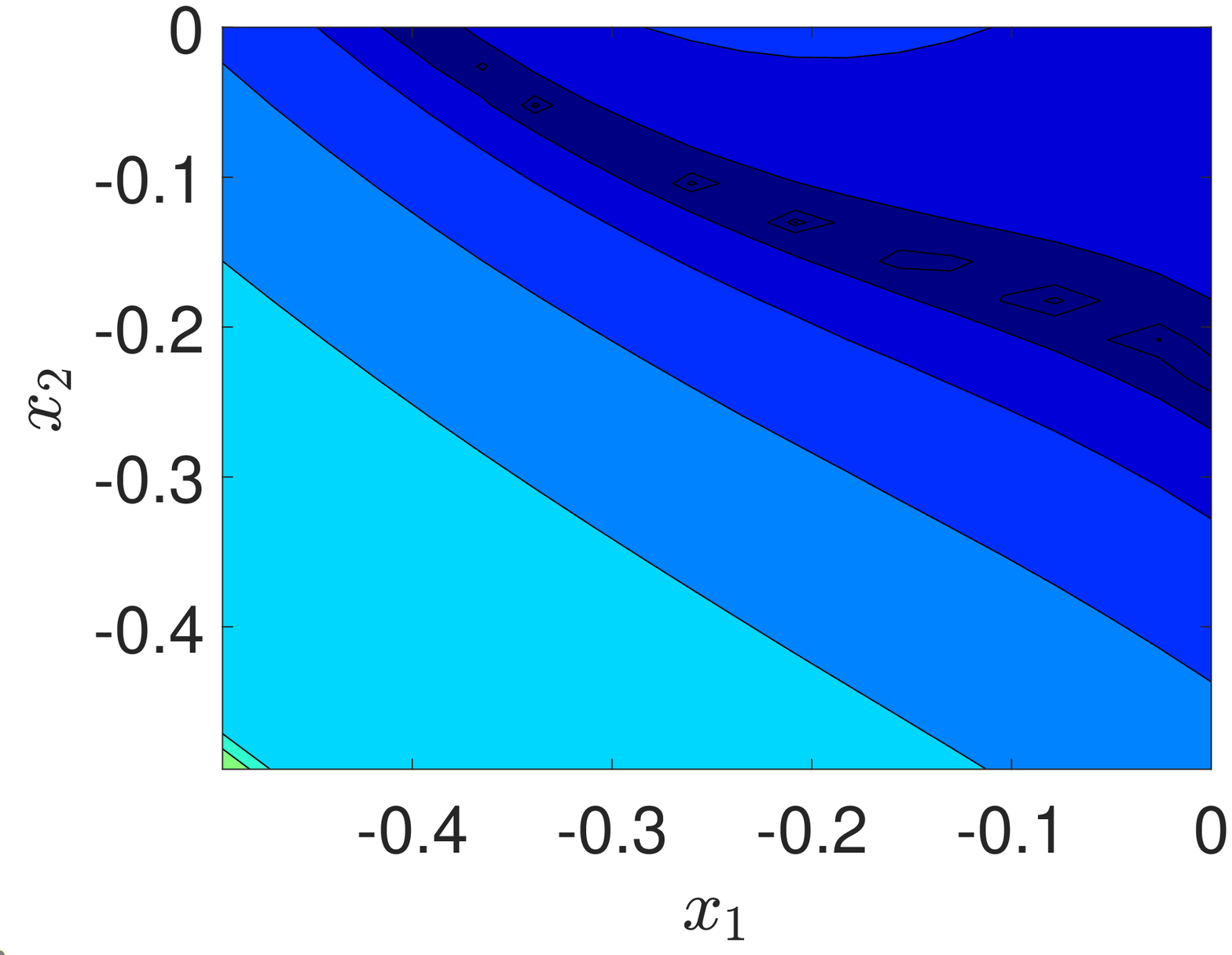}
     \caption{}
 \end{subfigure}
\caption{Black-box simulator. Training sets (grids of $20 \times 20$ equispaced distributed points). Numerical approximation accuracy (difference between the computed and analytical solution) of $T_1(x_1,x_2)$ (left column) and $T_2(x_1,x_2)$ (right column) using the various schemes. (a),(b) $6th$ order power-series expansion of $T_1(x_1,x_2)$ and $T_2(x_1,x_2)$ in $[-0.495,0]\times[-0.495,0]$. (c),(d) PIML in Matlab trained in the entire domain $[-0.495,0]\times[-0.495,0]$. (e),(f) PIML in Matlab trained via the greedy-wise approach.}
\label{fig:sXERROR}
\end{figure}
\begin{figure}[htbp]
 \centering
 \begin{subfigure}[h]{0.45\textwidth}
     \centering
\includegraphics[width=5.8cm]{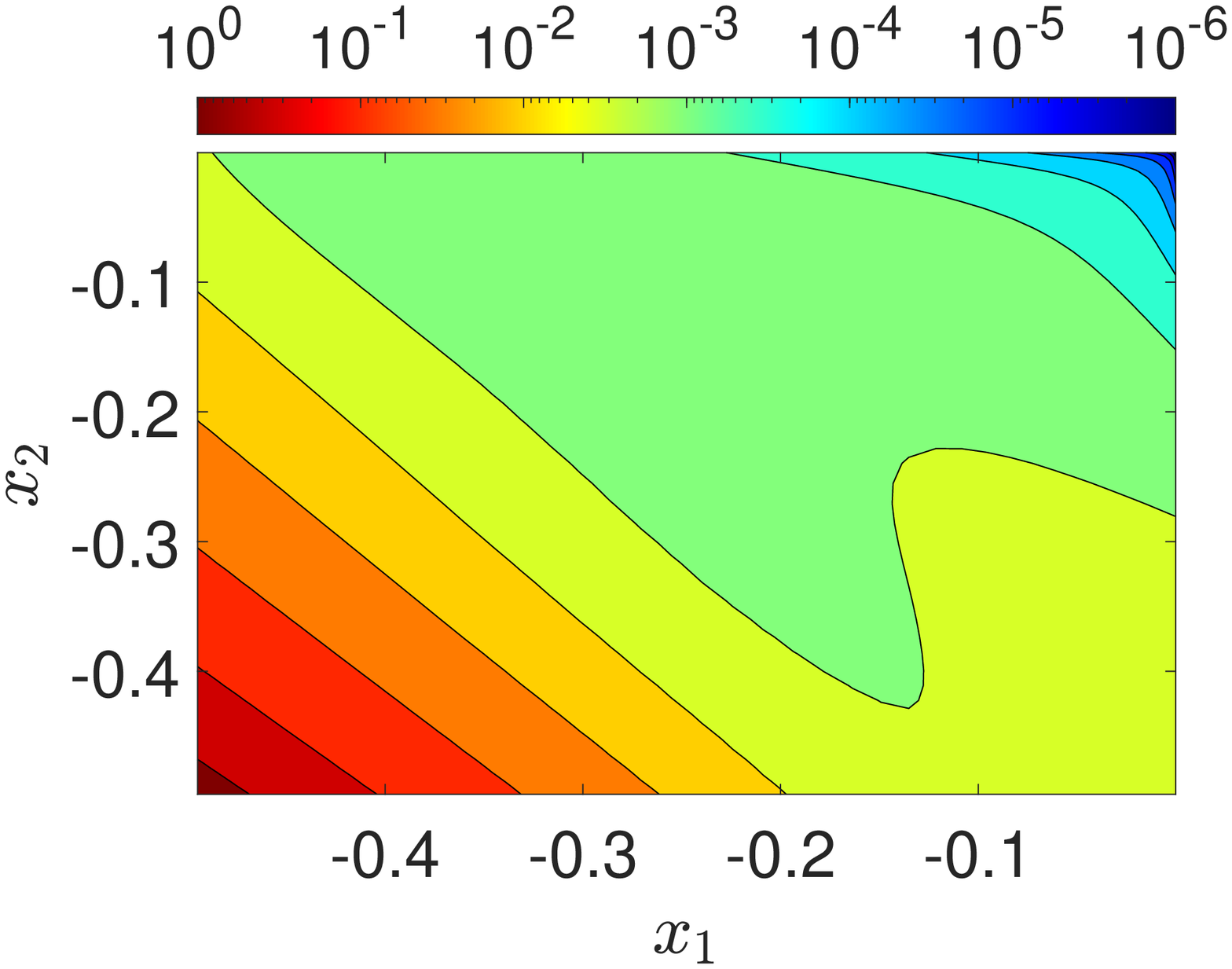}
     \caption{}
 \end{subfigure}
 \hfill
 \begin{subfigure}[h]{0.45\textwidth}
     \centering
\includegraphics[width=5.8cm]{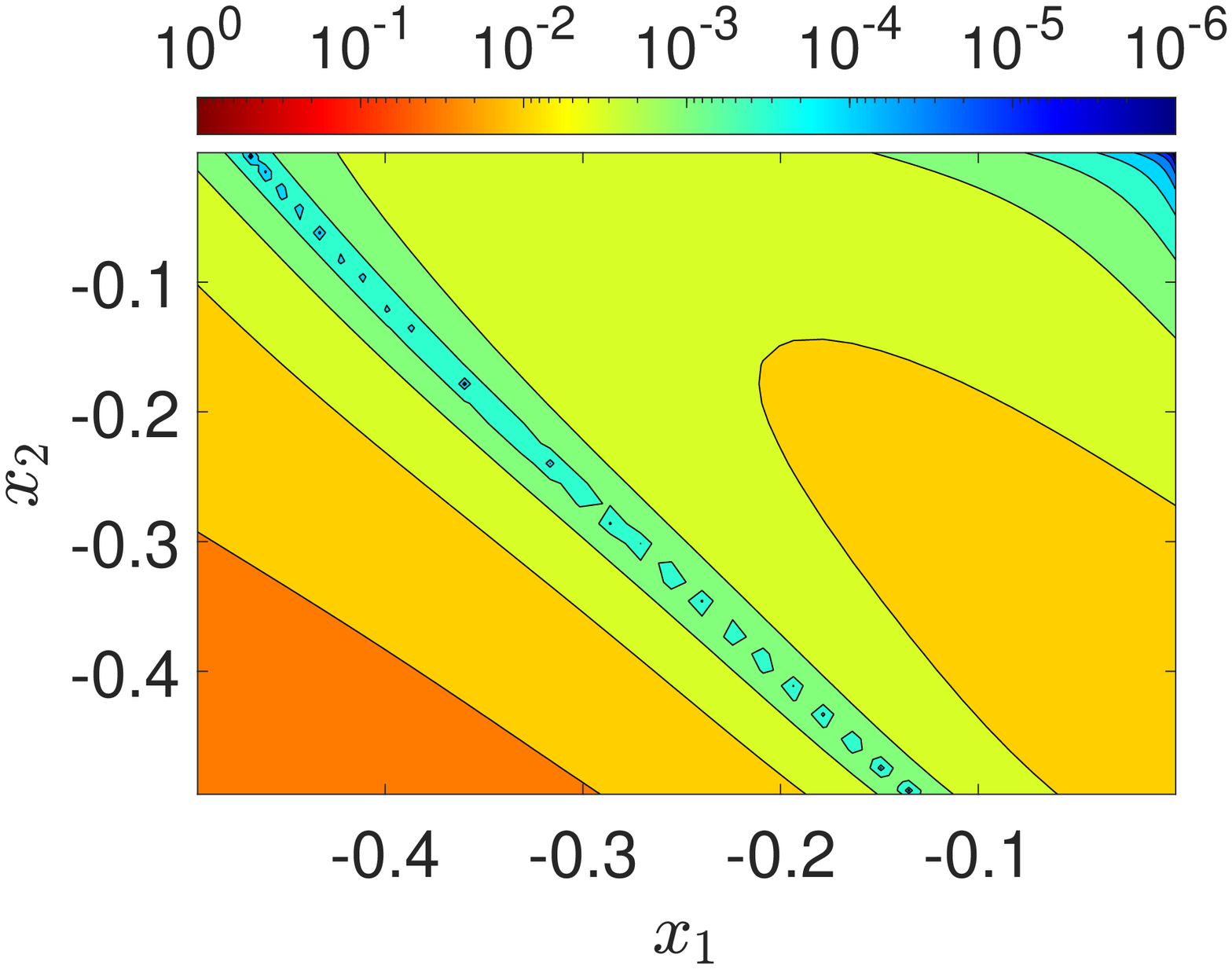}
     \caption{}
 \end{subfigure}
\centering
 \begin{subfigure}[h]{0.45\textwidth}
     \centering
\includegraphics[width=5.7cm]{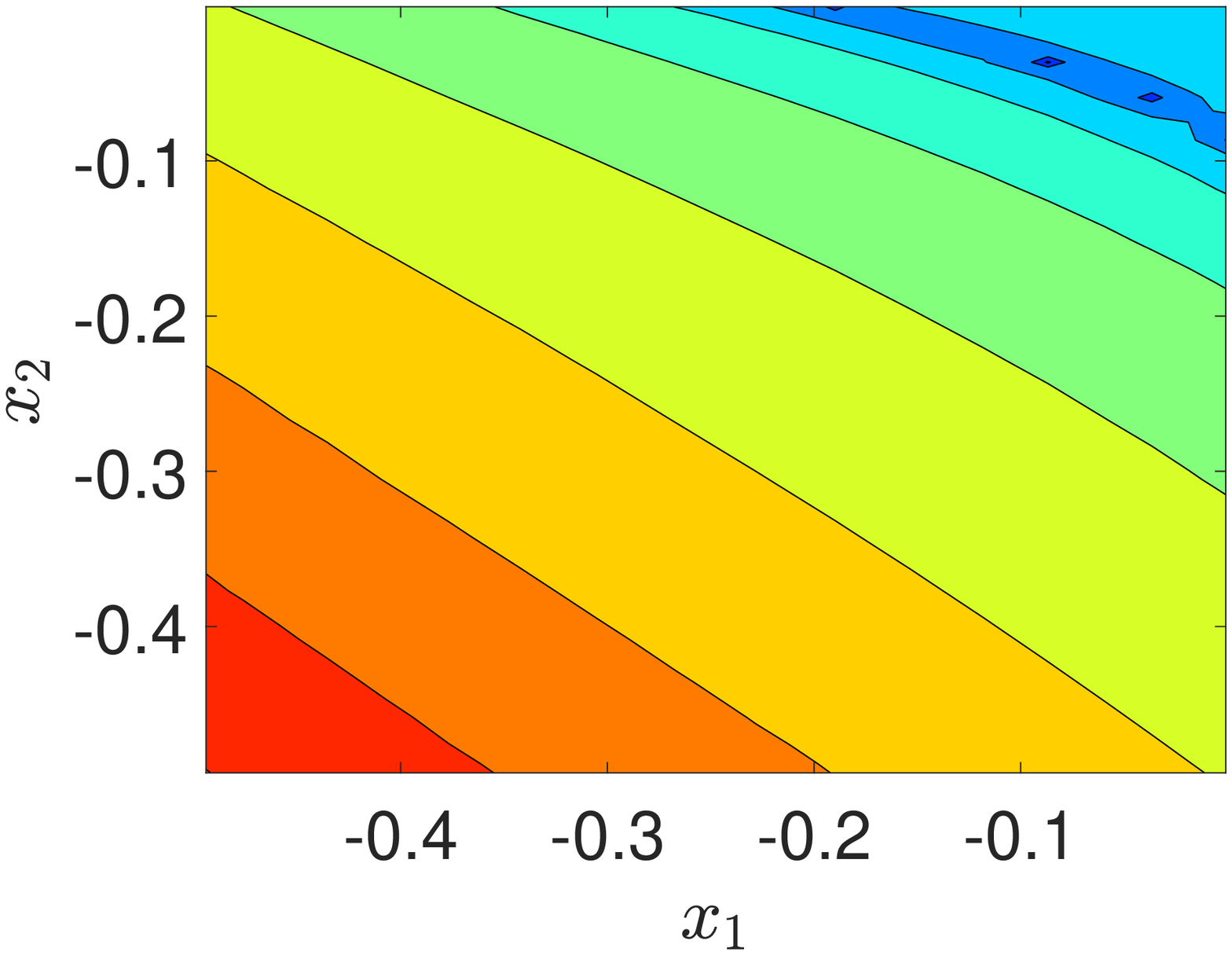}
     \caption{}
 \end{subfigure}
 \hfill
 \begin{subfigure}[h]{0.45\textwidth}
     \centering
\includegraphics[width=5.7cm]{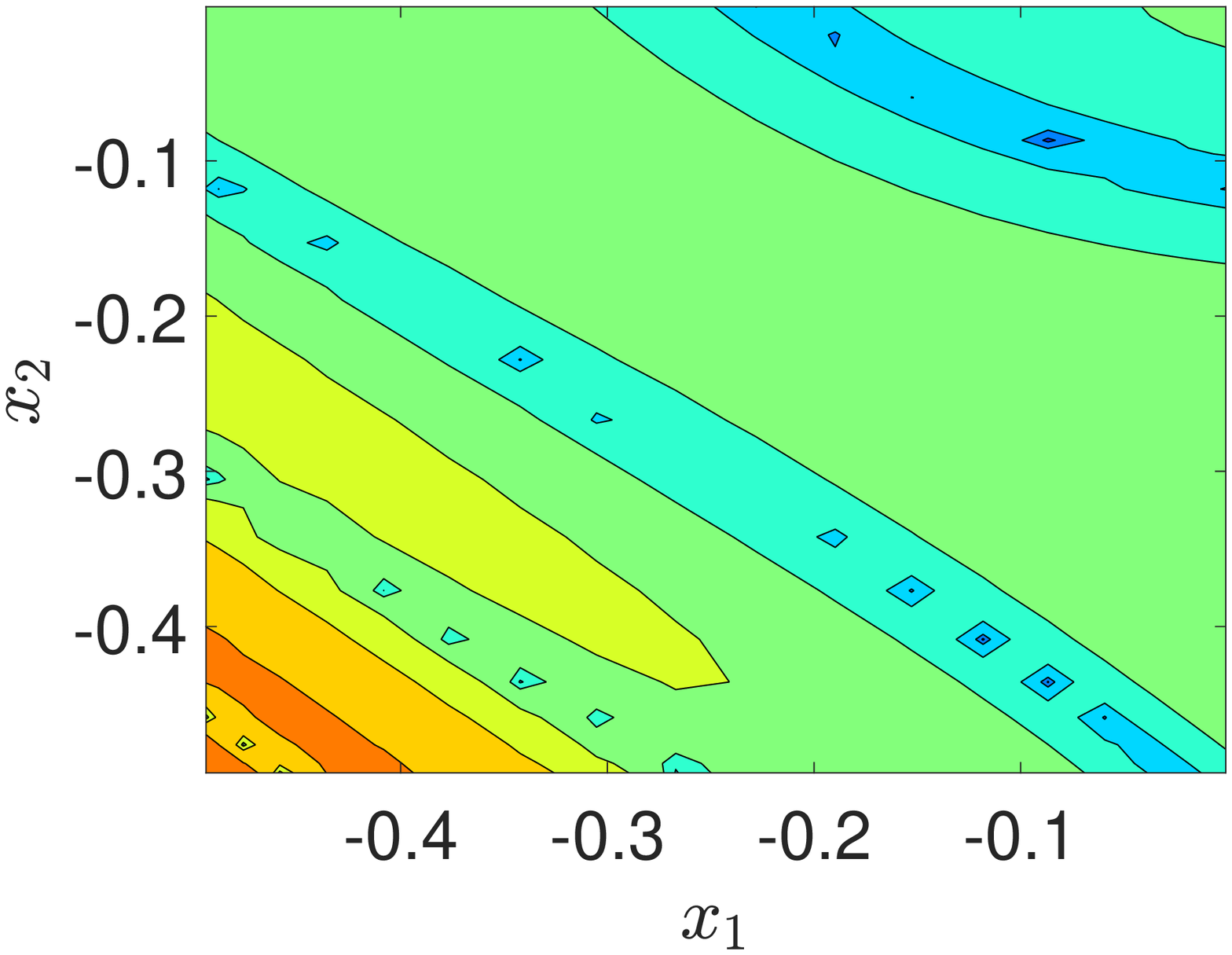}
     \caption{}
 \end{subfigure}
 \centering
 \begin{subfigure}[h]{0.45\textwidth}
     \centering
\includegraphics[width=5.7cm]{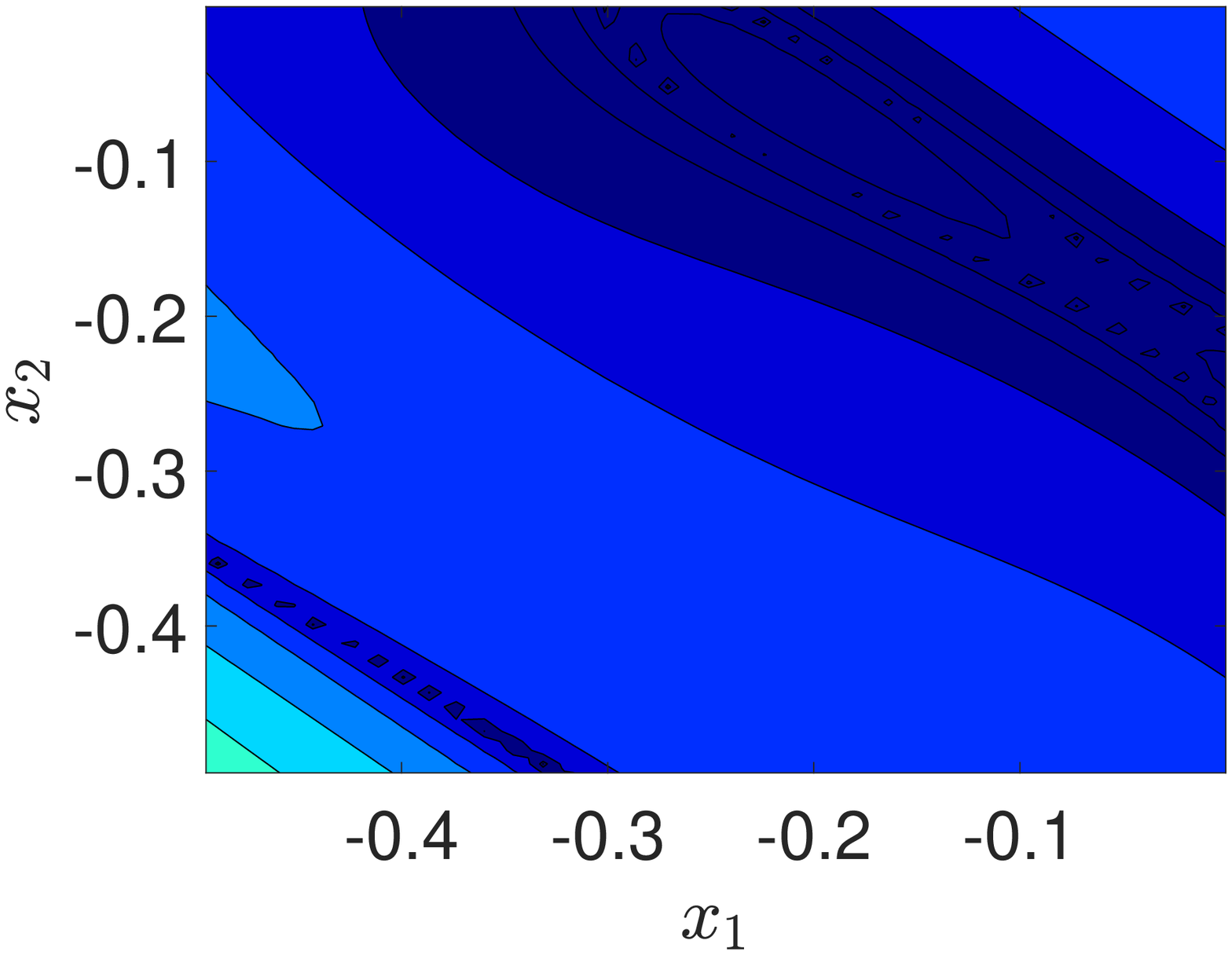}
     \caption{}
 \end{subfigure}
 \hfill
 \begin{subfigure}[h]{0.45\textwidth}
     \centering
\includegraphics[width=5.7cm]{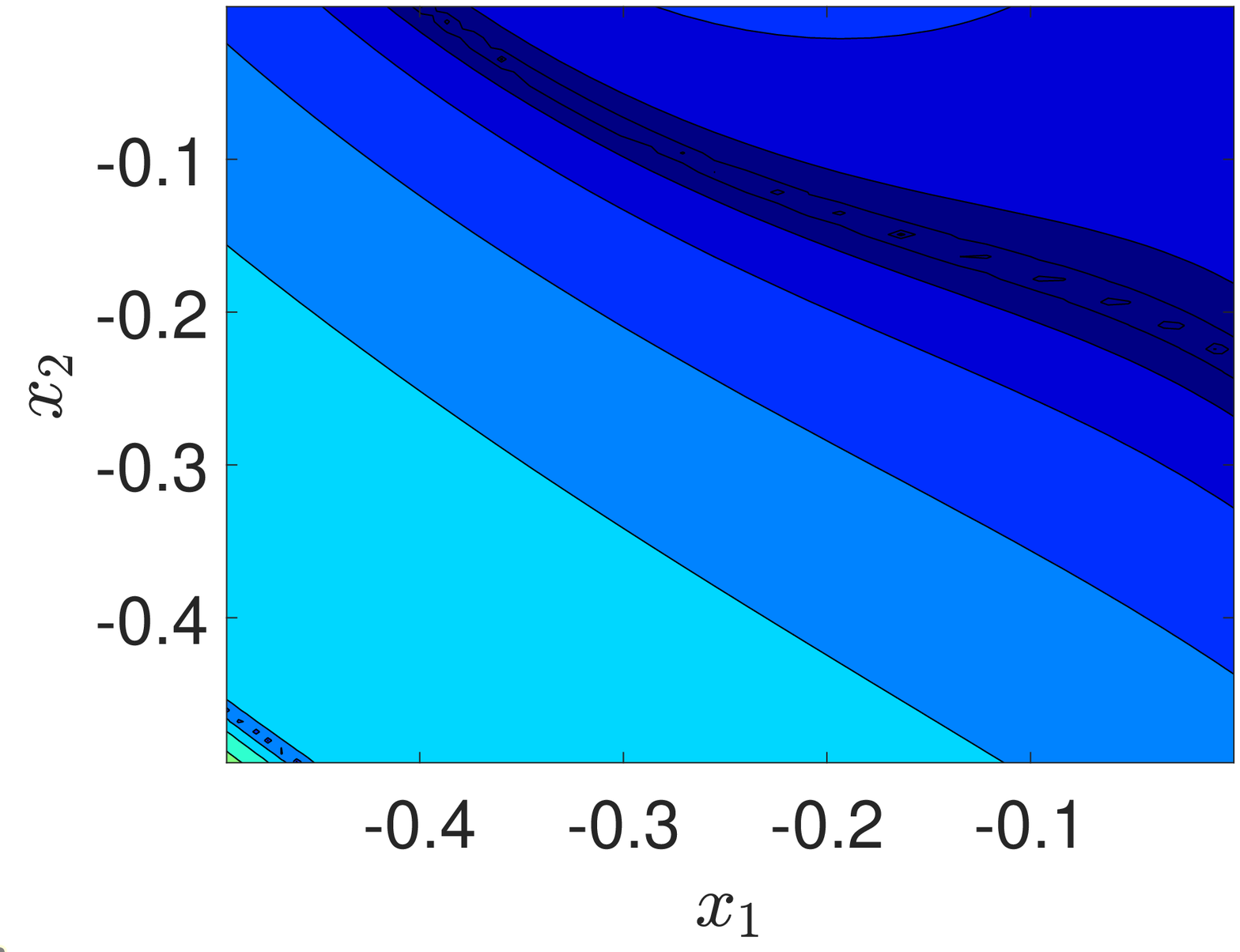}
     \caption{}
 \end{subfigure}
\caption{Black-box simulator. Test sets (grids of $50 \times 50$ Chebyshev-distributed points). Numerical approximation accuracy (difference between the computed and analytical solution) of $T_1(x_1,x_2)$ (left column) and $T_2(x_1,x_2)$ (right column) using the various schemes. (a),(b) $6th$ order power-series expansion of $T_1(x_1,x_2)$ and $T_2(x_1,x_2)$ in $[-0.495,0]\times[-0.495,0]$. (c),(d) PIML in Matlab trained in the entire domain $[-0.495,0]\times[-0.495,0]$. (e),(f) PIML in Matlab trained via the greedy-wise approach.}
\label{fig:sXERROR_test}
\end{figure}

As Figures (\ref{fig:sXERROR})(a),(b) show, in the case of the black-box simulator, the power-series expansion attained a lower/worse numerical approximation accuracy level compared to the one when the model is assumed to be explicitly known (Figure (\ref{fig:Modelavailable_TrS})(a),(b)). This is due to the fact that the right-hand side is not explicitly available. The approximation error is rather poor (of the order of $10^0$) in the region close to the singularity, i.e, the point $[-0.495,-0.495]$. Figures (\ref{fig:sXERROR})(c),(d) depict the PIML scheme as implemented in Matlab
trained in the entire domain. Here the approximation error is of the order of $10^{-1}$ in the region close to the singular point. Figures (\ref{fig:sXERROR})(e),(f) depict the PIML scheme implemented in Matlab trained with the greedy-wise procedure. The numerical approximation error in the domain where the step gradient appears is of the order of $10^{-3}$, thus outperforming the power-series expansion approximation of the transformation law, as well as the PIML trained in the entire domain at once. It should be noted that the numerical approximation error obtained in the region close to the singular point $[-0.495,-0.495]$ is of the same order as that obtained with the PIML implemented using the analytical derivatives (i.e., when the model is assumed to be explicitly known). Finally, Figure (\ref{fig:sXERROR_test}) depicts the numerical approximation errors on the test set. The results are similar both for the training and test sets.
Table \ref{tab:NormsS1_BB} and Table \ref{tab:NormsS1_BB_Test} detail the numerical approximation accuracy of the various schemes, for the training and test sets, respectively.\par
\begin{figure}[ht!]
    \centering
\includegraphics[width=12cm]{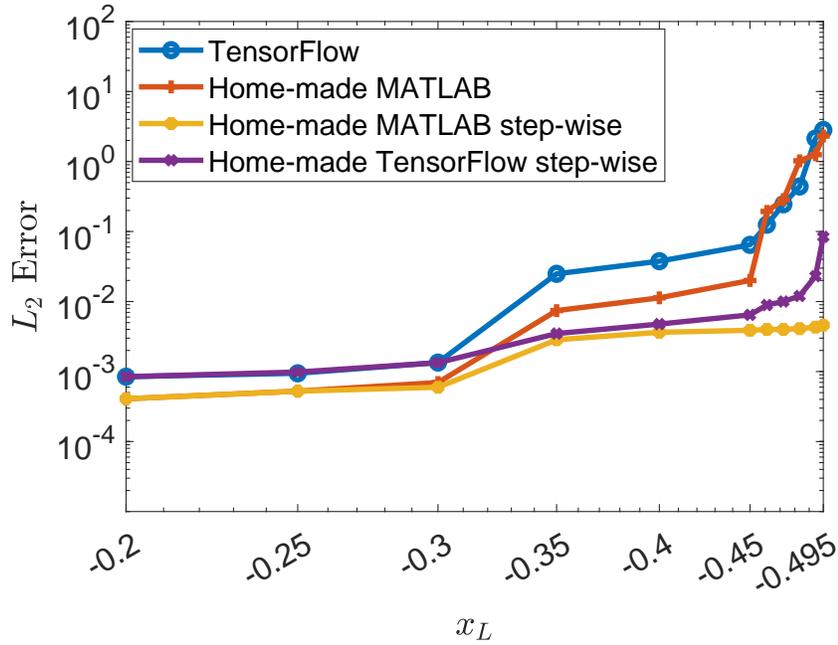}
\caption{$L_2$ norm of the numerical approximation accuracy between the analytical transformation law $T_1(x_1,x_2)$ (see Eq. \ref{NonlinTransformation}) and the various PIML implementations trained both greedy-wised and in the entire domain, for various sizes of the domain, $[x_L, 0]\times [x_L, 0]$. For our illustrations, we have used grids of $20 \times 20$ equispaced distributed collocation points.}
\label{fig:L2_error_norm}. 
\end{figure}

 
Finally, in Figure (\ref{fig:L2_error_norm}), we depict the numerical approximation error of the transformation $T_1(x_1,x_2)$ in terms of the $L_2$ norm with respect to the size of the domain for four different schemes, namely: (a) a PIML implemented in TensorFlow, trained in the entire domain (blue line), (b) a PIML implemented in Matlab trained with the Levenberg-Marquardt in the entire domain (orange line), (c) a PIML implemented in Matlab trained with the Levenberg-Marquardt using the greedy-wise procedure (yellow line), and, (d) a PIML implemented in TensorFlow using the greedy-wise procedure (purple line). For the construction of the diagram, we have used a grid of $20 \times 20$ equispaced distributed collocation points. Starting with a $-0.2\times-0.2$ grid and using a step of $-0.05$ we performed the training process each time until the interval $-0.45\times-0.45$ was reached. Then, for the interval between  $-0.45\times-0.45$ and $-0.49\times-0.49$, the step chosen for augmenting the grid were of size $-0.01$ and finally from this last interval to $-0.495\times-0.495$ the step chosen was of size $-0.001$.\par 

\begin{table}[!ht]
    \centering
    \caption{Black-box simulator. Training sets (grids of $20 \times 20$ equispaced distributed points). Error norms ($L_1$, $L_2$ and $L_{\infty}$) between the analytical and computed solution of $T_1(x_1,x_2)$ and $T_2(x_1,x_2)$ using the various schemes trained both greedy-wised and in the entire domain $[-0.495,0]\times[-0.495,0]$.}
    \begin{tabular}{c| c c | c  }
    \toprule
    Error norm & power-series&PIML(Matlab)&PIML(Matlab)\\
    & $6th$ order&Entire domain& Greedy\\
    \midrule
    $\lVert \cdot \rVert_1$ &1.50E$+$01& 1.21E$+$00& 6.64E$-$02 \\
      $\lVert \cdot \rVert_{2}$ & 9.73E$+$00&7.84E$-$01 & 2.40E$-$03 \\
    $\lVert \cdot \rVert_{\infty}$ &4.40E$+$00& 1.35E$-$01  & 1.97E$-$03\\
    \midrule
       $\lVert \cdot \rVert_1$ &1.00E$+$01& 2.36E$-$01 & 2.80E$-$03 \\
   $\lVert \cdot \rVert_{2}$
  &9.07E$+$00&2.44E$-$01& 1.77E$-$03 \\
$\lVert \cdot  \rVert_{\infty}$ &6.73E$+$00&1.10E$-$01 & 8.24E$-$04\\
    \bottomrule
    \end{tabular}
\label{tab:NormsS1_BB}
\end{table}

\begin{table}[!ht]
    \centering
    \caption{Black-box simulator. Test sets (grids of $50 \times 50$ Chebyshev-distributed points). Error norms ($L_1$, $L_2$ and $L_{\infty}$) between the analytical and computed solution of $T_1(x_1,x_2)$ and $T_2(x_1,x_2)$ using the various schemes trained both greedy-wised and in the entire domain $[-0.495,0]\times[-0.495,0]$.}
    \begin{tabular}{c| c c | c  }
    \toprule
    Error norm & power-series&PIML(Matlab)&PIML(Matlab)\\
     & $6th$ order&Entire domain& Greedy\\
    \midrule
    $\lVert \cdot \rVert_1$ &2.17E$+$01&3.54E$+$00 & 6.64E$-$02 \\
      $\lVert \cdot \rVert_{2}$ & 9.73E$+$00&2.65E$+$00& 2.40E$-$03 \\
    $\lVert \cdot \rVert_{\infty}$ &4.40E$+$00&1.81E$+$00& 1.97E$-$03\\
    \midrule
       $\lVert \cdot \rVert_1$ &1.00E$+$01&7.87E$-$01& 2.80E$-$03 \\
 $\lVert \cdot \rVert_{2}$
  &9.07E$+$00& 4.84E$-$01 & 1.77E$-$03 \\
$\lVert \cdot \rVert_{\infty}$ &6.73E$+$00&2.98E$-$01& 8.24E$-$04\\
    \bottomrule
    \end{tabular}
    \label{tab:NormsS1_BB_Test}
\end{table}
\newpage
\section{Conclusions}
We proposed and demonstrated a PIML-based scheme for the single-step feedback linearization with pole placement in one step for nonlinear discrete-time systems. 
Within the context of the present study, we considered a system for which the linearizing transformation map and state feedback control law exhibit a singular point, and thus very steep transformation gradients near it.
As the underlying optimization problem may lead to a poor solution (for example due to the effect of random initialization of the weights of the PIML), we have chosen to implement a greedy approach, thus tessellating the ``hard'' (in the entire domain) training/optimization problem into a sequence of simpler ones; this has also been suggested in other studies (see e.g. \cite{larochelle2009exploring}). Such a greedy training strategy, used to initialize weights in a region near a good local minimum, facilitates the optimization algorithm by implicitly acting as a regularization technique and thus resulting in a better generalization \cite{larochelle2009exploring}.
The existence of a singularity, on and beyond which the feedback linearization fails, is a hallmark of many problems that seek useful transformation by formulating and solving functional differential equations \cite{kevrekidis2017infinity}.
The same type of issue will, for example, arise in 
trying to compute flow-box transformations \cite{henderson2005computing}, or 
transformations to linearity (in the Koopman operator context \cite{budivsic2012applied,bollt2018matching}). 
Understanding how to test for such singularities, adaptively re-mesh in their neighborhood, estimate the associated singularity exponents, and even considering
possible analytic continuations beyond them, is an important issue \cite{kevrekidis2017infinity}. In fact, here we have implemented a simple zero-th order continuation in order to provide better initial guesses for the unknown weights of the PIML scheme to regions close to the singularity. In a future work, we aim at exploiting more advanced continuation techniques, such as the natural continuation technique proposed in Fabiani et al. \cite{fabiani2022parsimonious} for providing analytically initial guesses for the unknown weights of random projection networks for the solution of stiff ODEs and index-1 DAEs containing steep gradients in their solution profiles, or arc-length continuation for tracing branches of solutions and approximation of manifolds up to or even beyond critical/singular points (see for example \cite{fabiani2021numerical,galaris2022numerical}). Thus bridging ML programming techniques with concepts from continuation techniques, have the potential to significantly facilitate computational
experimentation and learning, and thus assist in the study of such singularities, possibly suggesting approaches to their mitigation.


\appendix
\section{Appendix: Multivariate power-series expansion approach}\label{sec:appendix_A}
In order to perform a comparative assessment of the performance of the  proposed computational approach, a practical
solution scheme for the associated system of NFE's \eqref{eq:NFESYS} is needed. Since $f(x,u)$ as well as the solution
$T(x)$ are all locally analytic around the origin, it is possible to
calculate the solution in the form of
a multivariate power-series. The proposed solution method involves the expansion of
$f(x,u)$ as well as the unknown
solution $T(x)$ in a power-series followed by equating the power coefficients of the same order of both sides of
the NFE's (3). Such a procedure leads to a hierarchy of recursion
formulas, through which one can calculate the $N$-th order power coefficients of
$T(x)$, given the power coefficients of
$T(x)$ up to the order $N-1$ (evaluated in previous recursive steps) \cite{Kazantzis2001}.

In the derivation of the associated recursion formulas, it is quite convenient to employ the
following tensorial notation:

a) The entries of a constant matrix $A$ are represented as $a_{i}^{j}$, where
the subscript $i$ refers to the corresponding
row and the superscript $j$ to the corresponding column of the matrix.

b) The partial derivatives of the $\mu$-th component $f_{\mu}(x,u)$ of the vector function
$f(x,u)$ with respect to the state variables $x$ evaluated at $(x,u)=(0,0)$ are denoted as follows:
\begin{eqnarray}
\hat{f}_{\mu}^{i}&=&\frac{\partial f_{\mu}}{\partial x_{i}}(0,0) \nonumber \\
\hat{f}_{\mu}^{ij}&=&\frac{\partial^{2} f_{\mu}}{\partial x_{i} \partial x_{j}}(0,0)
\nonumber \\
\hat{f}_{\mu}^{ijk}&=&\frac{\partial^{3} f_{\mu}}{\partial x_{i} \partial x_{j}
\partial x_{k}}(0,0)
\end{eqnarray}
etc., where $i,j,k,..$=$1,...,n$.

c) The partial derivatives of the $\mu$-th component $f_{\mu}(x,u)$ of the vector function
$f(x,u)$ with respect to the input variable $u$ evaluated at $(x,u)=(0,0)$ are denoted as follows:
\begin{equation}
g_{\mu}^{i}=\frac{\partial^{i} f_{\mu}}{\partial u^{i}}(0,0)
\end{equation}
etc.

d) The standard summation convention where repeated upper and lower tensorial indices are summed up.

Under the above notation the $l$-th component $T_{l}(x)$ of the unknown solution
$T(x)$ can be expanded in a multivariate power
series as follows:
\begin{eqnarray}
T_{l}(x)&=&\frac{1}{1 !}T_{l}^{i_{1}}x_{i_{1}}+\frac{1}{2
!}T_{l}^{i_{1}i_{2}}x_{i_{1}}x_{i_{2}}+...+ \nonumber \\
&+&\frac{1}{N
!}T_{l}^{i_{1}i_{2}...i_{N}}x_{i_{1}}x_{i_{2}}...x_{i_{N}}+...
\end{eqnarray}
As mentioned earlier, the proposed procedure is initiated by considering the expansion of the components of the vector function $f(x,u)$ in multivariate
power-series. Substituting the power-series expansions of $T(x)$, $f(x,u)$ into \eqref{eq:NFESYS} and matching the power coefficients of the same order,
the following recursive relations are obtained:

\underline{First order terms; N=1}

\begin{equation}
T_{l}^{\mu}(\hat{f}_{\mu}^{i_{1}}-g_{\mu}^{i}c^{k}T_{k}^{i_{1}})=T_{l}^{\mu}\hat{f}_{\mu}^{i_{1}}-T_{l}^{\mu}g_{\mu}^{i}c^{k}T_{k}^{i_{1}}=a_{l}^{\mu}T_{\mu}^{i_{1}}
\end{equation}
with: $i_{1}=1,...n$ and $l=1,..n$. Note that under the matrix notation and the summation convention introduced above, the set of algebraic equations (48) can be recast into the
following matrix equation:
\begin{equation}
    \bar{T}J-A\bar{T}=\bar{T}Gc\bar{T}
\end{equation}
where the unknown matrix $\bar{T}$ in (49) is the Jacobian of the map $T(x)$ evaluated at the origin.
Under the assumptions of Theorem 2.1, the unique invertible solution of the above quadratic matrix equation is given by : $\bar{T}=W^{-1}$, where $W$ is the unique and invertible solution of the Lyapunov matrix equation shown below \cite{Kazantzis2001}:
\begin{equation}
    JW-WA=Gc
\end{equation}
Since Lyapunov equations of the above type can be solved using a software package such as Matlab/Maple, the calculation of the solution of the first-order algebraic equations (48) does not pose any challenges. 

\underline{$N$-th order terms; $N \geq 2$}

\begin{equation}
\sum_{L=1}^{N}\sum_{ \genfrac{}{}{0pt}{}{0 \leq m_{1} \leq m_{2} \leq \dots\leq m_{L}}{m_{1}+m_{2}+...+m_{L}=N}}T_{l}^{
j_{1}...j_{L}}(\hat{f}^{m_{1}}_{j_{1}}..\hat{f}^{m_{L}}_{j_{L}}-\pi^{m_{1}}_{j_{1}}...\pi^{m_{L}}_{j_{L}})=a_{l}^{\mu}T_{\mu}^{i_{1}...i_{N}}
\end{equation}
where:
\begin{equation}
\pi_{j_{l}}^{m_{L}}=\sum_{P=1}^{L} \sum_{\genfrac{}{}{0pt}{}{0 \leq n_{1} \leq n_{2} \leq ...\leq n_{P}}{ n_{1}+n_{2}+...+n_{P}=m_{L}}} g_{j_{l}}^{n_{1}}
c^{k}T_{k}^{n_{2}...n_{P}}
\end{equation}
with  $i_{1},...,i_{N}=1,...,n$ and  $l=1,...,n$. Notice
that the second summation symbol in (51) indicates
summing
up the relevant quantities over the $\displaystyle{\frac{N!}{m_{1}!...m_{L}!}}$ possible combinations to assign the $N$ indices $(i_{1},...,i_{N})$
as upper indices to the $L$ positions: $\{\hat{f}_{j_{1}},...\hat{f}_{j_{L}}\}$ and $\{\pi_{j_{1}},...\pi_{j_{L}}\}$, with $m_{1}$ of them being put in the first position, $m_{2}$ of them in the
second position , etc. ($\displaystyle{\sum_{i=1}^{L}}m_{i}=N)$. Similar rules apply to equation (52).
Please notice that equations
(51,52) represent a set of linear algebraic equations in the unknown
coefficients $T_{\mu}^{i_{1},...,i_{N}}$ for $N \geq 2$. Furthermore, it should be pointed out, that the above series solution method for the system of NFEs \eqref{eq:NFESYS} may be accomplished in an automatic fashion by exploiting the computational capabilities of a symbolic software package such as MAPLE.

\section{Appendix: Learning the Feedback Linearization Operator from Black-Box simulators}
\label{sec:appendix_B}
One of the main differences between this method and the previous one, is that in this method we are not expanding both sides of the NFEs (\ref{NFEsys}), but just $T(x)$ in order to calculate the unknown coefficients of its series expansion through let's say nonlinear least squares. The information regarding the system is derived from or given by the output of the black-box simulator. Therefore, the problem under consideration can be stated as one whose target is finding the values of the vector $h$ such that the sum of squared errors on the discretization mesh is minimized in some norm for instance the 2-norm, i.e.
\begin{equation}
\min_{h} {\sum_{i=1}^{N} \parallel R_{i}(h)
\parallel ^{2} _{2}}
\end{equation}
where the vector function $R_{i}(h)$ is defined as follows:
\begin{gather}
 R_{i}(h)=\hat{T}(f(x_{i},-c\hat{T}(x_{i});h)-A\hat{T}(x_{i};h),\quad \forall
x_{i}   \label{equalityR}
\end{gather}
As mentioned earlier, this nonlinear optimization problem can be solved using a Gauss-Newton method or the Levenberg Marquard method in an iterative fashion, by enforcing equality (\ref{equalityR}) at every point of the discretized mesh. For example, for the power-series expansion the algorithm for computing the transformation law using a black-box simulator reads as follows:
\begin{itemize}
\item Choose a subdomain of the solution state space of the nonlinear system, i.e. $D\subseteq \mathbb{R}^{n}$ in a mesh of $N\times N$ points. In such a domain the solution of the NFEs system (on the basis of which the the feedback controller itself is synthesized)  will be learnt as well.

\item Expand the transformation map $T(x)$ in a power-series up to order $p$ around the equilibrium $x_o$, meaning that $T(x)$ must be expressed as a function of the vector $x$ and also the power-series coefficients $h\in \mathbb{R}^{m}$, ie $\hat{T}(x,h)$, i.e.
\begin{eqnarray}
\begin{matrix}
\hat{T}_1(x_{i=1,\cdots,n};h_{j,k=1,\cdots,p})=h_{1,1}x_1+h_{1,2}x_2+\frac{1}{2!}h_{1,3}x_1^2+\frac{1}{2!}h_{1,4}x_2^2+h_{1,5}x_1x_2+\cdots+O_1(p+1)\label{eqpower}\\ 
\hat{T}_2(x_{i=1,\cdots,n};h_{j,k=1,\cdots,p})=h_{2,1}x_1+h_{2,2}x_2+\frac{1}{2!}h_{2,3}x_1^2+\frac{1}{2!}h_{2,4}x_2^2+h_{2,5}x_1x_2+\cdots+O_2(p+1)\\ 
\vdots
\\ 
\hat{T}_n(x_{i=1,\cdots,n};h_{j,k=1,\cdots,p})=h_{n,1}x_1+h_{n,2}x_2+\frac{1}{2!}h_{n,3}x_1^2+\frac{1}{2!}h_{n,4}x_2^2+h_{n,5}x_1x_2+\cdots+O_n(p+1)
\end{matrix}
\end{eqnarray}

Then, write the feedback control law as $u= -c\hat{T}(x_i)$.
\item Obtain the information of the system by calling the output of the black-box simulator using the series expansion up to order $p$ of $T(x)$ as $u$.

\item Construct both sides of the NFEs system and get a residual as indicated in (\ref{equalityR}).
\item Add to the residual  \begin{gather}
  R_{i}(h)=\hat{T}(f(x_{i},-c\hat{T}(x_{i});h)-A\hat{T}(x_{i};h)=0
\end{gather} the following conditions:
\begin{itemize}
\item Initial condition 
\begin{gather}
 \hat{T}_{j}(0)=0,
\quad j,k=1,2,\dots,n
\label{pinningcondition1}
\end{gather}
\item Derivative of $\hat{T}(x_i)$ evaluated the equilibrium $x_o=0$
\begin{gather}
\frac{\partial \hat{T}_j}{\partial x_k}(0)-\frac{\partial {T}_j}{\partial x_k}(0)=0, \quad j,k=1,2,\dots,n
\label{pinningcondition}
\end{gather}
where, as mentioned before, $\frac{\partial T_j}{\partial x_k}(0)$ is the $(j,k)$-th element of the Jacobian matrix of $T(x)$ computed at the equilibrium and obtained by solving equation (\ref{eq:pinning}). The latter serves as a pinning condition for the optimization problem to find the best coefficients for the series expansion of the transformation map $T(x)$ that satisfy the residual $R_{i}(h)=0$ equality. Indeed, without this pinning condition, it is also probable that the optimization process finds the trivial solution that also satisfies these properties, but of course does not represent the feedback controller since the trivial solution maps all the states to the kernel of the linearized space. Finally, the derivative of $T(x)$ can be computed, for instance using finite differences.
\end{itemize}
\item Compute the unknown coefficients of $T(x_i; h)$ using a nonlinear optimization algorithm, such as the Levenberg–Marquardt, Gauss-Newton or perhaps using an unconstrained optimization algorithm, such as the Broyden, Fletcher, Goldfarb, Shanno (BFGS) method. 
\end{itemize}
A similar procedure can be used for the Physics Informed Machine-Learning (PIML) scheme.

\end{document}